\documentclass[final,preprint]{elsarticle}
\usepackage{lineno}

\modulolinenumbers[5]

\journal{Swarm and Evolutionary Computation}

\usepackage{amsmath}
\usepackage{amssymb}
\usepackage{amsfonts}

\usepackage{booktabs}
\usepackage{multirow}
\usepackage{subfig}
\usepackage{floatflt}
\usepackage{wrapfig}
\usepackage[linesnumbered, ruled,vlined]{algorithm2e}
\usepackage{url}
\usepackage{xcolor}
\usepackage{adjustbox}
\usepackage{tablefootnote}
\usepackage{lscape}
\usepackage{pdflscape}

\def\vector#1{\mbox{\boldmath $#1$}}

\begin{document}

\begin{frontmatter}

\title{A Niching Indicator-Based Multi-modal Many-objective Optimizer}


\author{Ryoji Tanabe}
\ead{rt.ryoji.tanabe@gmail.com}

\author{Hisao Ishibuchi\corref{cor1}}
\ead{hisao@sustc.edu.cn}
\cortext[cor1]{Corresponding author}

\address{Shenzhen Key Laboratory of Computational Intelligence, University Key Laboratory of Evolving Intelligent Systems of Guangdong Province, Department of Computer Science and Engineering, Southern University of Science and Technology, Shenzhen 518055, China}





\begin{abstract}

Multi-modal multi-objective optimization is to locate (almost) equivalent Pareto optimal solutions as many as possible.
Some evolutionary algorithms for multi-modal multi-objective optimization have been proposed in the literature.
However, there is no efficient method for multi-modal many-objective optimization, where the number of objectives is more than three.
To address this issue, this paper proposes a niching indicator-based multi-modal multi- and many-objective optimization algorithm.
In the proposed method, the fitness calculation is performed among a child and its closest individuals in the solution space to maintain the diversity.
The performance of the proposed method is evaluated on multi-modal multi-objective test problems with up to 15 objectives.
Results show that the proposed method can handle a large number of objectives and find a good approximation of multiple equivalent Pareto optimal solutions.
The results also show that the proposed method performs significantly better than eight multi-objective evolutionary algorithms.


\end{abstract}

\begin{keyword}
Multi-modal multi-objective optimization, many-objective optimization, indicator-based evolutionary algorithms, niching methods
\end{keyword}

\end{frontmatter}

\nolinenumbers

\section{Introduction}
\label{sec:introduction}


A bound-constrained multi-objective optimization problem (MOP) can be formulated as follows: 
%
\begin{align}
\label{eqn:mops}
&\text{Minimize  } \:\, \vector{f}(\vector{x}) = \bigl(f_1 (\vector{x}), ..., f_M(\vector{x}) \bigr)^{\rm T}\\
&\text{subject to  } \vector{x} \in \mathbb{S} \subseteq \mathbb{R}^D , \notag
\end{align}
where $\vector{f}: \mathbb{S} \rightarrow \mathbb{R}^M$ is an objective function vector that consists of $M$ potentially conflicting objective functions, and $\mathbb{R}^M$ is the objective space.
Here, $\vector{x} = (x_1, ..., x_D)^{\rm T}$ is a $D$-dimensional solution vector, and $\mathbb{S} = \Pi^D_{j=1} [x^{\rm min}_j, x^{\rm max}_j]$ is the bound-constrained solution space where $x^{\rm min}_j \leq x_j \leq x^{\rm max}_j$ for each index $j \in \{1, ..., D\}$.



We say that $\vector{x}^1$ dominates $\vector{x}^2$  iff $f_i (\vector{x}^1) \leq f_i (\vector{x}^2)$ for all $i \in \{1, ..., M\}$ and $f_i (\vector{x}^1) < f_i (\vector{x}^2)$ for at least one index $i$.
Here, $\vector{x}^*$ is a Pareto-optimal solution if there exists no $\vector{x} \in \mathbb{S}$ such that $\vector{x}$ dominates $ \vector{x}^*$.
 In this case, $\vector{f} (\vector{x}^*)$ is a Pareto-optimal objective vector.
The set of all $\vector{x}^*$ in $\mathbb{S}$ is the Pareto-optimal solution set (PS), and the set of all $\vector{f}(\vector{x}^*)$ is the Pareto front (PF).
In general, the goal of MOPs is to find a set of non-dominated solutions that are well-distributed and close to the PF in the objective space.



A multi-objective evolutionary algorithm (MOEA) is an efficient approach for solving MOPs \cite{Deb01}.
However, representative Pareto dominance-based MOEAs (e.g., NSGA-II \cite{DebAPM02}) do not work well on MOPs with a large number of objectives \cite{WagnerBN07}, where MOPs with four or more objectives are referred to as many-objective optimization problems (MaOPs).
This is because most individuals in the population are non-dominated with each other in MaOPs, resulting a weak selection pressure to the PF \cite{WagnerBN07}.
Since MaOPs appear in real-world problems \cite{LiLTY15}, the poor performance of MOEAs on MaOPs is critical.
For this reason, a number of MOEAs for MaOPs have been proposed in the literature \cite{LiLTY15}. 

Some recent studies (e.g., \cite{WagnerBN07,TanabeIO17}) show that indicator-based MOEAs perform well on MaOPs.
In this framework, a so-called fitness value is assigned to each individual in the population using quality indicators \cite{ZitzlerTLFF03}.
Then, the mating and/or environmental selections are performed based on the fitness values.
Representative indicator-based MOEAs include IBEA \cite{ZitzlerK04}, SMS-EMOA \cite{BeumeNE07}, and HypE \cite{BaderZ11}.

\begin{figure}[t]
\newcommand{\widthvar}{0.7}
  \begin{center} 
\includegraphics[width=\widthvar\textwidth]{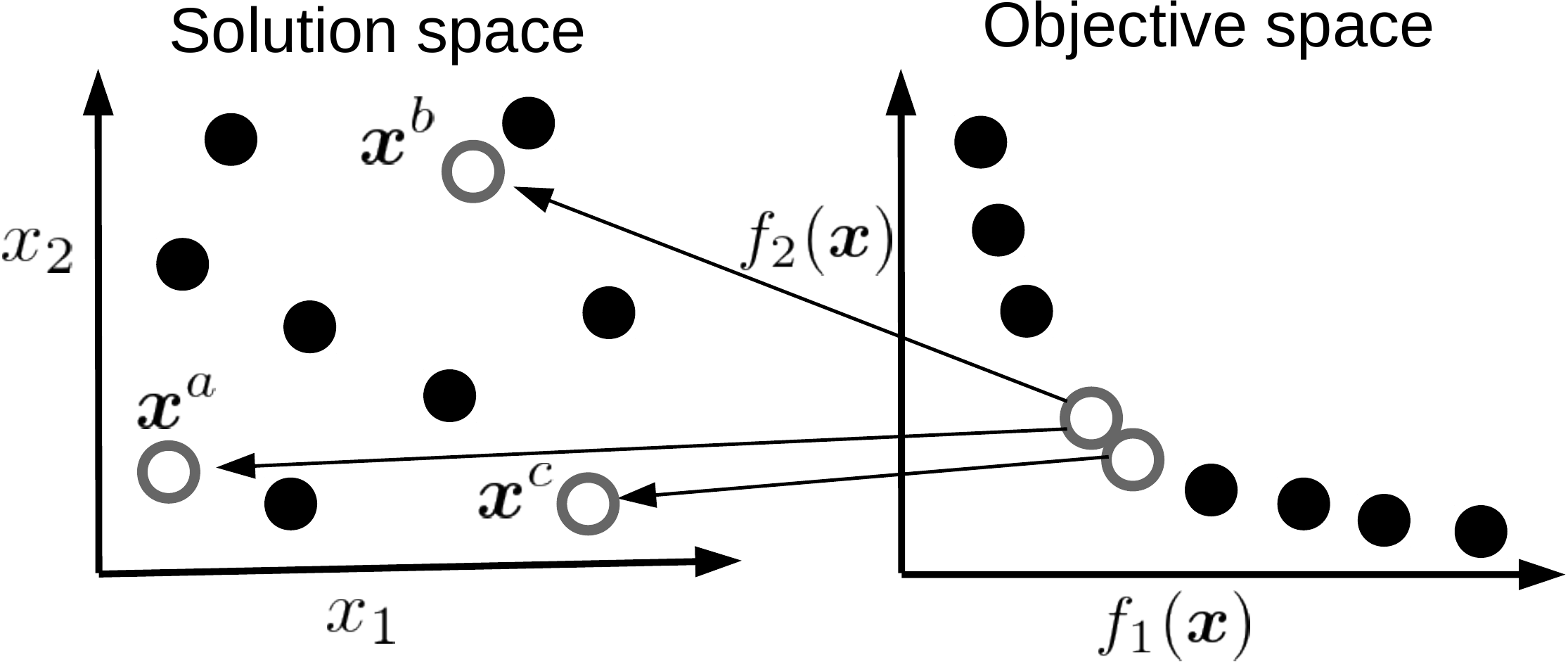}
\caption{
\small
Illustration of a situation where three solutions ($\vector{x}^a$, $\vector{x}^b$, and $\vector{x}^c$) are identical or close to each other in the objective space (right) but far from each other in the solution space (left).
This figure was made using \cite{ShirPNE09,LiEDE17} as reference.
}
\label{fig:mmop_example}
  \end{center}
\end{figure}

There are multiple Pareto optimal solutions that have (almost) the same objective vector in some real-world problems, such as diet design problems \cite{RudolphNP07}, space mission design problems \cite{SchutzeVC11}, rocket engine design problems \cite{KudoYF11}, and functional brain imaging problems \cite{SebagTTLB05}.
Fig. \ref{fig:mmop_example} explains a situation where three solutions are identical or close to each other in the objective space but far from each other in the solution space.
On the one hand, the performance of MOEAs is generally evaluated based only on the distribution of non-dominated solutions obtained in the objective space \cite{ZitzlerTLFF03}.
Thus, the distribution of non-dominated solutions in the solution space is usually not taken into account to evaluate the performance of MOEAs \cite{ZhouZJ09}.
On the other hand, diverse non-dominated solutions in the solution space are helpful for decision-making in practice \cite{SchutzeVC11,RudolphNP07,DebT08,ShirPNE09}.
If multiple non-dominated solutions with (almost) the same objective vector are obtained, users can make a final decision according to their preference which cannot be represented by the objective functions.
For example, in Fig. \ref{fig:mmop_example}, if $\vector{x}^a$ becomes unavailable for some reasons (e.g., materials shortages and traffic accidents), $\vector{x}^b$ and $\vector{x}^c$ can be considered as candidates for the final solution instead of $\vector{x}^a$.
Diverse solutions are also helpful to figure out properties of problems \cite{SebagTTLB05,HiroyasuNM05}.





The above-discussed issue has been addressed as a multi-modal MOP (MMOP) \cite{SebagTTLB05,DebT08,TanabeI19emmo}.
While the goal of MOPs is to achieve a good approximation of the PF, the goal of MMOPs is to obtain as many as possible equivalent Pareto optimal solutions.
For example, in Fig. \ref{fig:mmop_example}, it is sufficient to find one of $\vector{x}^a$, $\vector{x}^b$, and $\vector{x}^c$ for multi-objective optimization.
This is because their objective vectors are almost the same.
In contrast, it is necessary to find all of $\vector{x}^a$, $\vector{x}^b$, and $\vector{x}^c$ for multi-modal multi-objective optimization.


Since most MOEAs (e.g., NSGA-II \cite{DebAPM02} and SPEA2 \cite{ZitzlerLT01}) do not have mechanisms to find diverse solutions in the solution space, they are unlikely to perform well for MMOPs.
Thus, multi-modal multi-objective evolutionary algorithms (MMEAs) that handle the solution space diversity are necessary for MMOPs.
MMEAs need the following three abilities: (i) the ability to find solutions with high quality, (ii) the ability to find diverse solutions in the objective space, and (iii) the ability to find diverse solutions in the solution space.
MOEAs need the abilities (i) and (ii) to find a solution set that approximates the Pareto front in the objective space.
Multi-modal single-objective evolutionary algorithms \cite{DasMQS11,LiEDE17} need the abilities (i) and (iii) to find a set of global optimal solutions.
In contrast, an efficient MMEA should be able to handle all three abilities (i)--(iii) in order to find all equivalent Pareto optimal solutions.




Wheres multi-modal single-objective optimization is a hot research topic in the evolutionary computation community, MMOPs have not been well studied \cite{LiEDE17}.
As reviewed in \cite{TanabeI19emmo}, some MMEAs have been proposed, such as Omni-optimizer \cite{DebT08}, Niching-CMA \cite{ShirPNE09}, and MO\_Ring\_PSO\_SCD \cite{YueQL17}.
%
In addition, most existing MMEAs use the Pareto dominance relation for the mating and environmental selections.
Since dominance-based MOEAs do not work well on MaOPs as mentioned above, it is expected that existing MMEAs are not capable of handling many objectives.
In this paper, an MMOP with four or more objectives is denoted as a multi-modal many-objective optimization problem (MMaOP).
While MMaOPs appear in real-world problems (e.g., five-objective rocket engine design problems \cite{KudoYF11}), there is no method that can efficiently locate multiple equivalent Pareto optimal solutions in problems with more than three objectives.



To address this issue, this paper proposes a niching indicator-based multi-modal many-objective optimizer (NIMMO).
%
%
Niching is a loosely defined term.
Usually, in evolutionary multi-modal single-objective optimization, the term ``niching'' indicates a mechanism to find multiple local optimal solutions and/or maintain the diversity of the population \cite{LiEDE17}.
An evolutionary algorithm with the niching method is usually capable of finding multiple local optima.
Also, as mentioned above, some indicator-based MOEAs can handle a large number of objectives.
Therefore, it is expected that an efficient multi-modal many-objective optimization method can be realized by incorporating a niching mechanism in the solution space into an indicator-based MOEA.

Our contributions in this paper are as follows:

\begin{enumerate}
\item We address MMaOPs. Although most previous studies address only two- and three-objective MMOPs, MMaOPs can be found in real-world applications and important problems.
\item We propose NIMMO. While most existing MMEAs were designed for two- and three-objective MMOPs, NIMMO is designed for MMaOPs. NIMMO handles the diversity in both the objective and solution spaces in a novel manner.
\item We investigate the performance of NIMMO on various test problems. We demonstrate that NIMMO can find multiple equivalent solutions on MMaOPs with up to 15 objectives. We also compare NIMMO with state-of-the-art MMEAs (TriMOEA-TA\&R \cite{LiuYG18}, MO\_Ring\_PSO\_SCD \cite{YueQL17}, and Omni-optimizer \cite{DebT08}).  
\item We analyze the performance of NIMMO. We examine the influence of a control parameter and the population size on the performance of NIMMO. In addition, we investigate how the fitness assignment scheme affects the performance of NIMMO.
\end{enumerate}

The rest of this paper is organized as follows.
Section \ref{sec:related_work} describes related work.
Section \ref{sec:proposed_method} introduces NIMMO.
Section \ref{sec:experimental_settings} describes experimental settings. 
Section \ref{sec:experimental_results} reports experimental results of NIMMO on MMaOPs.
Section \ref{sec:conclusion} concludes this paper with discussions on directions for future work.

\section{Related work}
\label{sec:related_work}

Since the proposed NIMMO is an indicator-based MMEA, this section describes indicator-based MOEAs and MMEAs as related work.
Subsection \ref{sec:indicator_based_emoas} explains indicator-based MOEAs.
Subsection \ref{sec:emoas_for_mmops} describes MMEAs.







\subsection{Indicator-based MOEAs for multi-objective optimization}
\label{sec:indicator_based_emoas}

Unary quality indicators map a solution set $\vector{A}$ to a real value that quantitatively represents one or more properties of $\vector{A}$ in the objective space \cite{ZitzlerTLFF03,LiY19}.
For example, the generational distance (GD) \cite{VeldhuizenL98} evaluates the convergence of $\vector{A}$ to the PF in the objective space.
The hypervolume \cite{ZitzlerT98} evaluates both convergence and diversity of $\vector{A}$ in the objective space.
In general, quality indicators play a central role in evaluating the performance of MOEAs.




Indicator-based MOEAs directly use a quality indicator in the search \cite{ZitzlerK04}.
For each iteration, so-called fitness values are assigned to individuals in the population using a quality indicator.
Then, the fitness values are used in the mating and/or environmental selection.
The first indicator-based MOEA is IBEA \cite{ZitzlerK04}.
IBEA can use any binary Pareto-compliant indicator.
The additive $\epsilon$ indicator ($I_{\epsilon +}$) and the hypervolume difference indicator ($I_{HD}$) are proposed in \cite{ZitzlerK04} for IBEA.
Subsection \ref{sec:nibea} explains $I_{\epsilon +}$ in detail.
The $I_{HD}(\vector{y}, \vector{x})$ value is the volume of the space dominated by $\vector{x}$ but not by $\vector{y}$ with respect to a reference point \cite{ZitzlerK04}.
A binary version of the R2 indicator \cite{HansenJ98} ($I_{R2}$) is also proposed in \cite{PhanS13} for IBEA.
The R2 indicator uses a utility function with a weight vector set that maps a solution set to a scalar value.
$I_{R2}(\vector{y}, \vector{x})$ is calculated from the R2 values of $\{\vector{y}\}$ and $\{\vector{y}, \vector{x}\}$.

One of the main issues of IBEA with $I_{\epsilon +}$ is that the distribution of individuals in the objective space is biased \cite{WagnerBN07,GomezC15,LiOJDNW17}.
To address this issue, some variants of IBEA with $I_{\epsilon +}$ have been proposed in the literature.
Two\_Arch2 \cite{WangJY15} uses two archives: the convergence archive and the diversity archive.
While individuals in the convergence archive are updated based on the selection in IBEA with $I_{\epsilon +}$, individuals in the diversity archive are updated based on their L$_p$ norm-based crowding distance values.
SRA \cite{LiTLY16} simultaneously uses $I_{\epsilon +}$ and $I_{SDE}$, where $I_{SDE}$ is the shift-based density estimation in SPEA2$+$SDE \cite{LiYL14}.
SRA uses the stochastic ranking \cite{RunarssonY00} to sort individuals based on their $I_{\epsilon +}$ and $I_{SDE}$ values.
The BCE framework \cite{LiYL16} aims to address the biased distribution of individuals in the population in IBEA with $I_{\epsilon +}$ and MOEA/D  \cite{ZhangL07}.
BCE uses a bounded external archive that maintains well-distributed non-dominated individuals in the objective space.
The selection is performed from individuals in the population and the external archive.
The above-mentioned Two\_Arch2, SRA, and IBEA with BCE have shown good performance for many-objective optimization.

Another representative indicator-based MOEA is SMS-EMOA \cite{BeumeNE07}, which is a steady-state algorithm.
In the environmental selection in SMS-EMOA, the primary and secondary criteria are based on the non-domination levels and the hypervolume contribution, respectively.
The worst individual regarding the hypervolume contribution is removed from the last front.
Since the hypervolume calculation is expensive on problems with many objectives, it is difficult to apply SMS-EMOA to MaOPs.
HypE \cite{BaderZ11} addresses this issue by using Monte Carlo simulation for the hypervolume approximation.
HypE uses an approximated hypervolume value instead of its exact value.
The results presented in \cite{BaderZ11} show that  HypE has good performance on problems with up to 50 objectives.
Some MOEAs use a similar framework to SMS-EMOA, such as $R2$-EMOA \cite{BrockhoffWT15}, MyO-DEMR \cite{DenysiukCE13}, and AR-MOEA \cite{TianCZCJ18}.
Some other hypervolume-based MOEAs have also been proposed, such as FV-MOEA \cite{JiangZOZT15} and LIBEA \cite{MartinezJG19}.

In addition, a number of indicator-based MOEAs have been proposed in the literature.
GD-MOEA \cite{Menchaca-Mendez15} and IGD$^+$-EMOA \cite{LopezC16} use the GD indicator \cite{VeldhuizenL98} and the IGD$^+$ indicator \cite{IshibuchiMTN15}, respectively.
DDE \cite{VillalobosC12} uses the averaged Hausdorff distance indicator ($\Delta_p$) \cite{SchutzeELC12} for the search, where $\Delta_p$ is calculated from GD and IGD \cite{CoelloS04}.
Although MOMBI-II \cite{GomezC15} uses the R2 indicator similar to IBEA with $I_{R2}$, the ranking method in MOMBI-II differs from that in IBEA.
AGE \cite{BringmannFNW11} uses an unbounded external archive that maintains all non-dominated solutions found so far.
In AGE, the fitness value of an individual is based on the contribution to the additive $\epsilon$ indicator \cite{ZitzlerTLFF03} (not $I_{\epsilon+}$ in IBEA \cite{ZitzlerK04}) using the unbounded external archive.
Unlike most indicator-based MOEAs, MIHPS \cite{Falcon-CardonaC18} adaptively uses multiple indicators.
DIVA \cite{UlrichBZ10} uses an indicator that integrates the solution space diversity into the hypervolume indicator.

\subsection{MMEAs}
\label{sec:emoas_for_mmops}


As reviewed in \cite{TanabeI19emmo}, multi-modal multi-objective optimization has been studied since around 2005 in the evolutionary computation community.
Real-world applications of MMEAs include functional brain imaging problems \cite{SebagTTLB05}, molecular drug design problems \cite{KruisselbrinkAEIBBH09}, and military operational planning problems \cite{ZengDLZC12}.
Here, we explain some MMEAs.
For an exhaustive survey of MMEAs, refer to \cite{TanabeI19emmo}.

Omni-optimizer \cite{DebT08} is the most representative MMEA.
Omni-optimizer is an NSGA-II-based algorithm which can be applied to various problem domains, including MOPs and MMOPs.
The population in Omni-optimizer is initialized using the Latin hypercube sampling method so that individuals are distributed as uniformly as possible.
In the environmental selection, Omni-optimizer uses the $\epsilon$-dominance-based non-dominated sorting and the alternative crowding distance.
The crowding distance in Omni-optimizer is based on both the objective and solution spaces.
Roughly speaking, the alternative crowding distance value of an individual is the maximum value between its crowding distance values in the objective and solution spaces.
In contrast to NSGA-II, Omni-optimizer uses the restricted tournament mating selection.
First, an individual $\vector{x}_a$ is randomly selected from the population.
The nearest neighborhood individual $\vector{x}_b$ of $\vector{x}_a$ in the solution space is also selected from the population.
Then, $\vector{x}_a$ and $\vector{x}_b$ are compared based on their non-domination levels.
Ties are broken by the alternative crowding distance values of $\vector{x}_a$ and $\vector{x}_b$.
The winner can be a parent.
This procedure is iterated to select another parent.

Niching-CMA \cite{ShirPNE09} and MO\_Ring\_PSO\_SCD \cite{YueQL17} are CMA-ES and PSO algorithms for multi-modal multi-objective optimization, respectively.
Niching-CMA uses a niching strategy with the dynamic peak identification method \cite{ShirB09} in the environmental selection.
The number of niches and the niche radius are adaptively adjusted.
For each niche, better individuals regarding non-domination levels survive to the next iteration.
Niching-CMA uses the aggregated distance in both the objective and solution spaces.
Although the crowding distance metric in MO\_Ring\_PSO\_SCD is similar to that of Omni-optimizer, MO\_Ring\_PSO\_SCD handles extreme individuals in the objective space in a different manner.
MO\_Ring\_PSO\_SCD also uses the index-based ring topology \cite{Li10} to maintain high diversity in the solution space.
SMPSO-MM \cite{LiangGYQY18} is based on MO\_Ring\_PSO\_SCD.
SMPSO-MM uses the self-organizing map to select neighborhood best particles.
The new positions of the particles are perturbed by the mutation operation with a pre-defined probability.
In addition, some dominance-based MMEAs have been proposed, such as SPEA2$+$ \cite{KimHMW04}, 4D-Miner \cite{SebagTTLB05}, $P_{Q, \epsilon}$-MOEA \cite{SchutzeVC11}, and DN-NSGA-II \cite{LiangYQ16}.

Some decomposition-based MMEAs have been proposed, such as a multi-start decomposition-based approach \cite{RudolphNP07}.
MOEA/D-AD \cite{TanabeI18} can assign one or more individuals to each decomposed subproblem.
This mechanism is to handle equivalent solutions in the framework of MOEA/D \cite{ZhangL07}.
First, a child is assigned to the subproblem whose weight vector is closest to the objective vector of the child, with respect to the perpendicular distance.
If the child is close to one of the individuals assigned to the same subproblem in the solution space, they are compared based on their scalarizing function values.
Otherwise, the child is assigned to the subproblem with no comparison.
Unlike other MMEAs, the population size in MOEA/D-AD is adaptively adjusted.

TriMOEA-TA\&R \cite{LiuYG18} consists of various components, including the two archives-based method as in Two\_Arch2 \cite{WangJY15} and the decision variable-decomposition method in MOEA/DVA \cite{MaLQWLJYG16}.
At the beginning of the search, TriMOEA-TA\&R decomposes design variables of a given problem into position-related variables and distance-related variables.
While position-related variables affect only the position of the objective vector on the PF, distance-related variables affect only the distance between the objective vector and the PF.
Then, TriMOEA-TA\&R exploits the properties of position-related and distance-related variables in an explicit manner.
In TriMOEA-TA\&R, a child is assigned to the subproblem whose weight vector is closest to the objective vector of the child, with respect to their angle as in RVEA \cite{ChenJOS16}.
Although the methods of assigning the child to the subproblem in TriMOEA-TA\&R and MOEA/D-AD are similar, the population size in TriMOEA-TA\&R is constant.
While MOEA/D-AD uses the relative distance-based neighborhood criterion, TriMOEA-TA\&R uses the absolute distance-based neighborhood criterion.
In TriMOEA-TA\&R, if the distance between two individuals is less than a pre-defined value, they are said to be in the same niche.


\section{Proposed method}
\label{sec:proposed_method}

This section introduces the proposed NIMMO.
NIMMO can use any indicator-based fitness assignment scheme.
In this paper, we use the fitness assignment scheme with the additive epsilon indicator $I_{\epsilon+}$ in IBEA \cite{ZitzlerK04,BasseurB07}.
As reviewed in Subsection \ref{sec:indicator_based_emoas}, a number of indicator-based MOEAs have been proposed in the literature.
Since most indicator-based MOEAs ignore the diversity of the population in the solution space, it is likely that they cannot locate multiple equivalent Pareto optimal solutions.
For example, in Fig. \ref{fig:mmop_example}, even if all of $\vector{x}^a$, $\vector{x}^b$, and $\vector{x}^c$ are included in the current population, two of them are very likely to be removed from the population after the environmental selection.
NIMMO addresses this issue by using a niching mechanism in the solution space.
Although DIVA \cite{UlrichBZ10} uses the indicator that handles the solution space diversity, the indicator calculation in DIVA is expensive.
For this reason, DIVA has been applied only to two-objective problems.
In contrast, NIMMO handles the solution space diversity in a computationally cheap manner.

MMEAs need to handle the diversity in both the objective and solution spaces.
While most MMEAs  aggregate the objective and solution space diversity metrics (e.g., Omni-optimizer and SMPSO-MM), NIMMO handles the objective and solution space diversity in a two-phase manner similar to MOEA/D-AD and TriMOEA-TA\&R.
NIMMO also uses the relative distance-based neighborhood criterion similar to MOEA/D-AD.
As described in Subsection \ref{sec:emoas_for_mmops}, first, MOEA/D-AD (TriMOEA-TA\&R) assigns the child to the subproblem based on the perpendicular distance (angle) between the objective vector of the child and the weight vector.
Then, the child is compared to individuals in the same niche in the solution space.
In NIMMO, first, individuals to be compared with the child are selected based on the distribution in the solution space.
Then, the fitness values are assigned to the child and the selected individuals based on the distribution in the objective space.


Multi-modal single-objective optimization has been well studied in the evolutionary computation community.
As reviewed in \cite{LiEDE17,DasMQS11}, a number of niching techniques have been proposed for multi-modal single-objective optimization.
Traditional approaches include fitness sharing \cite{GoldbergR96}, crowding \cite{DeJong75}, and clustering \cite{YinG93}.
Recent approaches include the adaptive radius-based method \cite{BirdL06,EpitropakisLB13}, the nearest-better clustering \cite{Preuss12}, and the taboo region-based method \cite{AhrariDP17}.
The niching method used in NIMMO is similar to the deterministic crowding method \cite{Mahfoud92}.
While the deterministic crowding method compares the child to its closest parent in the solution space, NIMMO compares the child to its $T$ closest individuals in the solution space.
In addition, the environmental selection in NIMMO is based on the indicator values of the child and those $T$ closest individuals.

\subsection{The procedure of NIMMO}
\label{sec:nibea}



Algorithm \ref{alg:nibea} shows the procedure of NIMMO.
Similar to SMS-EMOA \cite{BeumeNE07}, NIMMO is a steady state MMEA and generates a child $\vector{u}$ for each iteration $t$.
After the initialization of the population $\vector{P}$ (line 1), the following operations are iteratively applied until a termination condition is satisfied.


At the beginning of each iteration $t$, the mating and the reproduction are performed (lines 3--5).
First, parent individuals $\vector{x}^a$ and $\vector{x}^b$ are randomly selected from $\vector{P}$ (line 3).
Then, a child $\vector{u}$ is generated by recombining $\vector{x}^a$ and $\vector{x}^b$ (line 4).
A mutation operator is also applied to $\vector{u}$ (line 5).

After the generation of $\vector{u}$, the environmental selection is performed (lines 6--13).
Since NIMMO is based on the $(\mu + 1)$-selection, only one individual is discarded from the population $\vector{P}$.
First, for each individual $\vector{x}^i$ in $\vector{P}$, the distance $d_i$ between $\vector{x}^i$ and $\vector{u}$ is calculated (lines 6--7).
The function ${\rm normalizedEuclideanDistance}(\vector{x}, \vector{y})$ in line 7 returns the normalized Euclidean distance between $\vector{x}$ and $\vector{y}$ using the upper and lower bounds for each variable. 
%
Next, $T$ closest individuals to $\vector{u}$ in the solution space are selected from $\vector{P}$ based on their distances and stored into $\vector{R}$ (line 8), where $T$ is the neighborhood size.
The $T$ individuals stored in $\vector{R}$ are temporarily removed from $\vector{P}$ (line 9), and $\vector{u}$ is inserted into $\vector{R}$ (line 10).
%
Thus, $\vector{R}$ is a set of $T$ neighbors of $\vector{u}$ in the solution space and $\vector{u}$.
The environmental selection in NIMMO is performed only among $T+1$ individuals in $\vector{R}$.
The worst individual in $\vector{R}$ cannot survive to the next iteration.

In line 11, a fitness value $F(\vector{x})$ is assigned to each individual $\vector{x}$ in $\vector{R}$ (not $\vector{P}$).
Although any fitness assignment method of indicator-based MOEAs can be incorporated into NIMMO, we use that of IBEA \cite{ZitzlerK04}, which is the most basic one.
The fitness assignment in IBEA is performed as follows:
%
\begin{align}
\label{eqn:fitness_ibea}
F(\vector{x}) = \sum_{\vector{y} \in \vector{R}\backslash \{\vector{x}\}} {\rm exp}\left(-\frac{I(\vector{y}, \vector{x})}{\kappa \, I^{\rm max}}\right),
\end{align}
where $I$ is a Pareto dominance preserving binary indicator.
If $\vector{x}$ dominates $\vector{y}$, $I(\vector{x}, \vector{y}) < I(\vector{y}, \vector{x})$.
The maximum absolute indicator value $I^{\rm max}$ is defined as follows: $I^{\rm max} = \max_{\vector{x}, \vector{y} \in \vector{R}} |I(\vector{x}, \vector{y})|$.
The scale factor $\kappa$ in  \eqref{eqn:fitness_ibea} is usually set to $0.05$.
In  \eqref{eqn:fitness_ibea}, $\vector{x}$ and every $\vector{y}$ in $\vector{R}\backslash \{\vector{x}\}$ are compared with each other according to their objective vectors.
A smaller fitness value represents a better individual.


\IncMargin{0.5em}
\begin{algorithm}[t]
\small
\SetSideCommentRight
$t \leftarrow 1$, initialize the population $\vector{P} =\{ \vector{x}^{1}, ..., \vector{x}^{\mu}\}$\;
%
\While{$\textsf{\upshape{The termination criteria are not met}}$}{
  Randomly select the parent individuals $\vector{x}^a$ and $\vector{x}^b$ from $\vector{P}$ such that $a \neq b$\;
  Generate the child $\vector{u}$ by recombining $\vector{x}^a$ and $\vector{x}^b$\;
  Apply the mutation operator to $\vector{u}$\;
  \For{$i  \in \{1, ..., \mu\}$}{
    $d_i \leftarrow {\rm normalizedEuclideanDistance}(\vector{x}^i, \vector{u})$\;
  }
  Select $T$ closest individuals to $\vector{u}$ based on the distance values $d_1, ..., d_\mu$ and store them into $\vector{R}$\;
  $\vector{P} \leftarrow \vector{P} \backslash \vector{R}$\;
  $\vector{R} \leftarrow \vector{R} \cup \{\vector{u}\}$\;
  Calculate the fitness value $F(\vector{x})$ of $\vector{x} \in \vector{R}$ using  \eqref{eqn:fitness_ibea}\;
  Remove the worst individual $\vector{x}^{\rm worst}$ from $\vector{R}$\;
  $\vector{P} \leftarrow \vector{P} \cup \vector{R}$\;
  $t \leftarrow t + 1$\;
}
\caption{The proposed NIMMO}
\label{alg:nibea}
\end{algorithm}\DecMargin{0.5em}

We use the additive epsilon indicator $I_{\epsilon+}$ \cite{ZitzlerK04,BasseurB07} for $I$ in \eqref{eqn:fitness_ibea}.
As described in Subsection \ref{sec:indicator_based_emoas}, some variants of IBEA with $I_{\epsilon+}$ (e.g., Two{\_}Arch2 and SRA) show good performance on MaOPs.
Therefore, it is expected that NIMMO with $I_{\epsilon+}$ is capable of handling many objectives.
%
The indicator value $I_{\epsilon+}(\vector{y}, \vector{x})$  is calculated as follows:
%
\begin{align}
\label{eqn:additive_epsilon_indicator}
I_{\epsilon+} (\vector{y}, \vector{x}) = \max_{i \in \{1, ..., M\}}  \bigl\{f'_i (\vector{y}) - f'_i (\vector{x})\bigr\},
%
\end{align}
%
where $f'_i (\vector{x})$ is the normalized objective value of $\vector{x}$ with $f_i^{\rm min} = \min_{\vector{x} \in \vector{R}} \{f_i (\vector{x})\}$ and $f_i^{\rm max} = \max_{\vector{x} \in \vector{R}} \{f_i(\vector{x})\}$ as follows: $f'_i (\vector{x}) = (f_i (\vector{x}) - f_i^{\rm min}) / (f_i^{\rm max} - f_i^{\rm min})$.
In \eqref{eqn:additive_epsilon_indicator}, $I_{\epsilon+} (\vector{y}, \vector{x})$ represents the minimal translation such that $\vector{y}$ dominates $\vector{x}$ \cite{BasseurB07}.
Although the original $I_{\epsilon+}$ is for comparing two sets of solutions,  \eqref{eqn:additive_epsilon_indicator} describes $I_{\epsilon+}$ as an indicator for comparing two solutions for the sake of simplicity.




The worst individual $\vector{x}^{\rm worst}$ regarding the fitness value is discarded from $\vector{R}$ (line 12).
After the removal operation, the remaining individuals in $\vector{R}$ are returned into $\vector{P}$ (line 13).


\subsection{The effect of the niching mechanism in NIMMO on MMOPs}
\label{sec:ibea_vs_nibea}



\begin{figure*}[t]
\newcommand{\widthvar}{0.327}
  \begin{center} 
\includegraphics[width=0.332\textwidth]{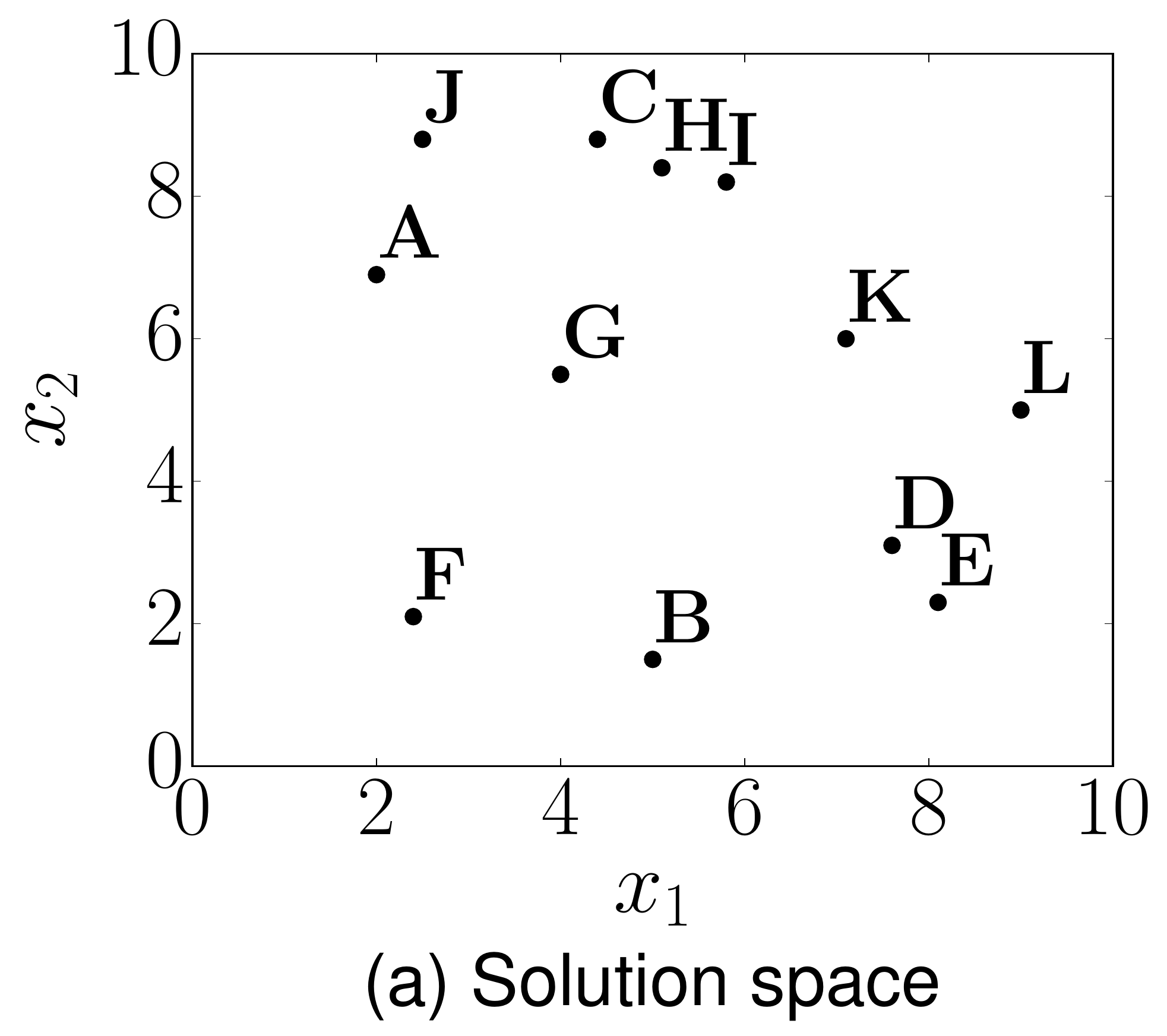}
\includegraphics[width=0.321\textwidth]{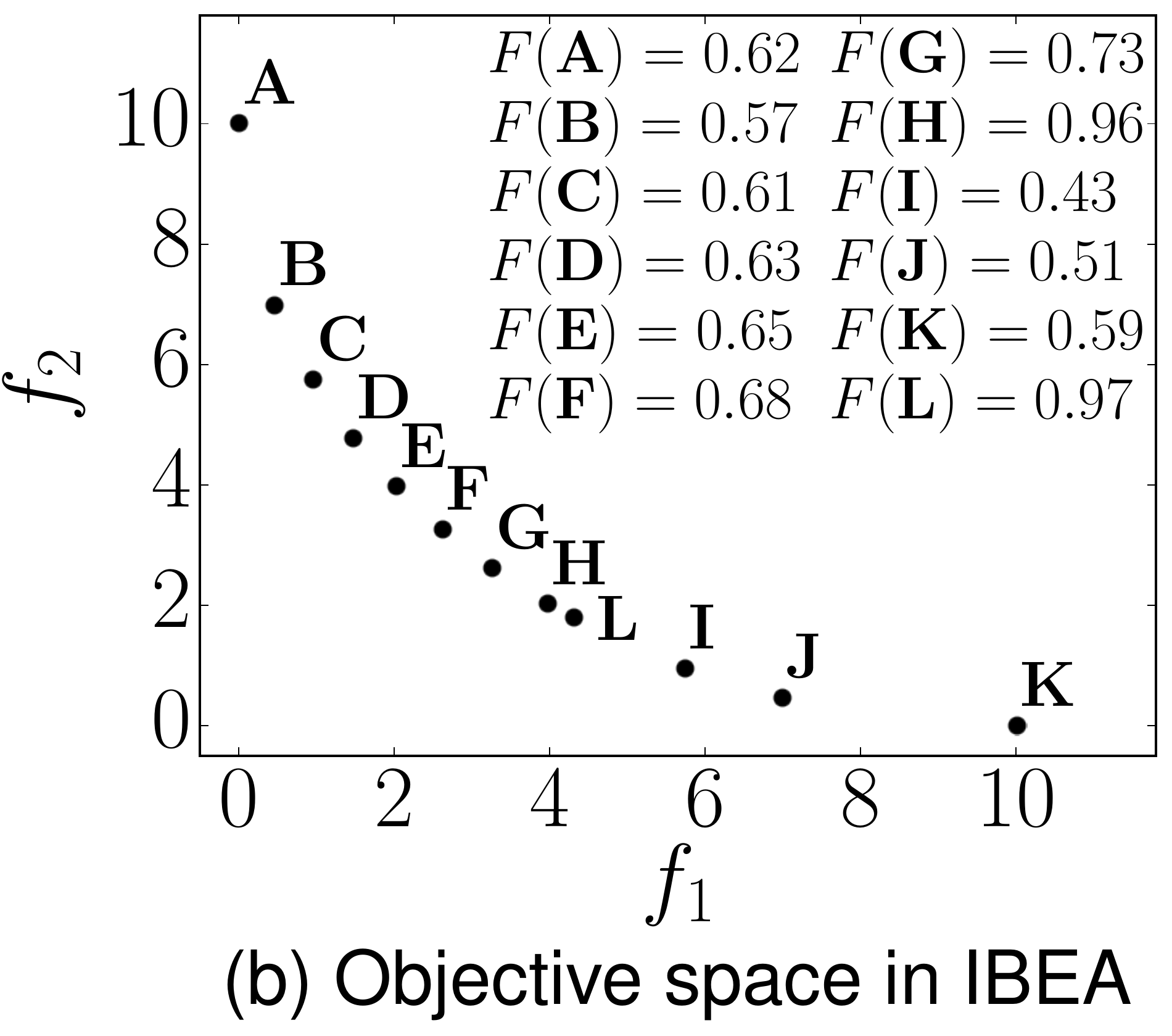}
\includegraphics[width=0.327\textwidth]{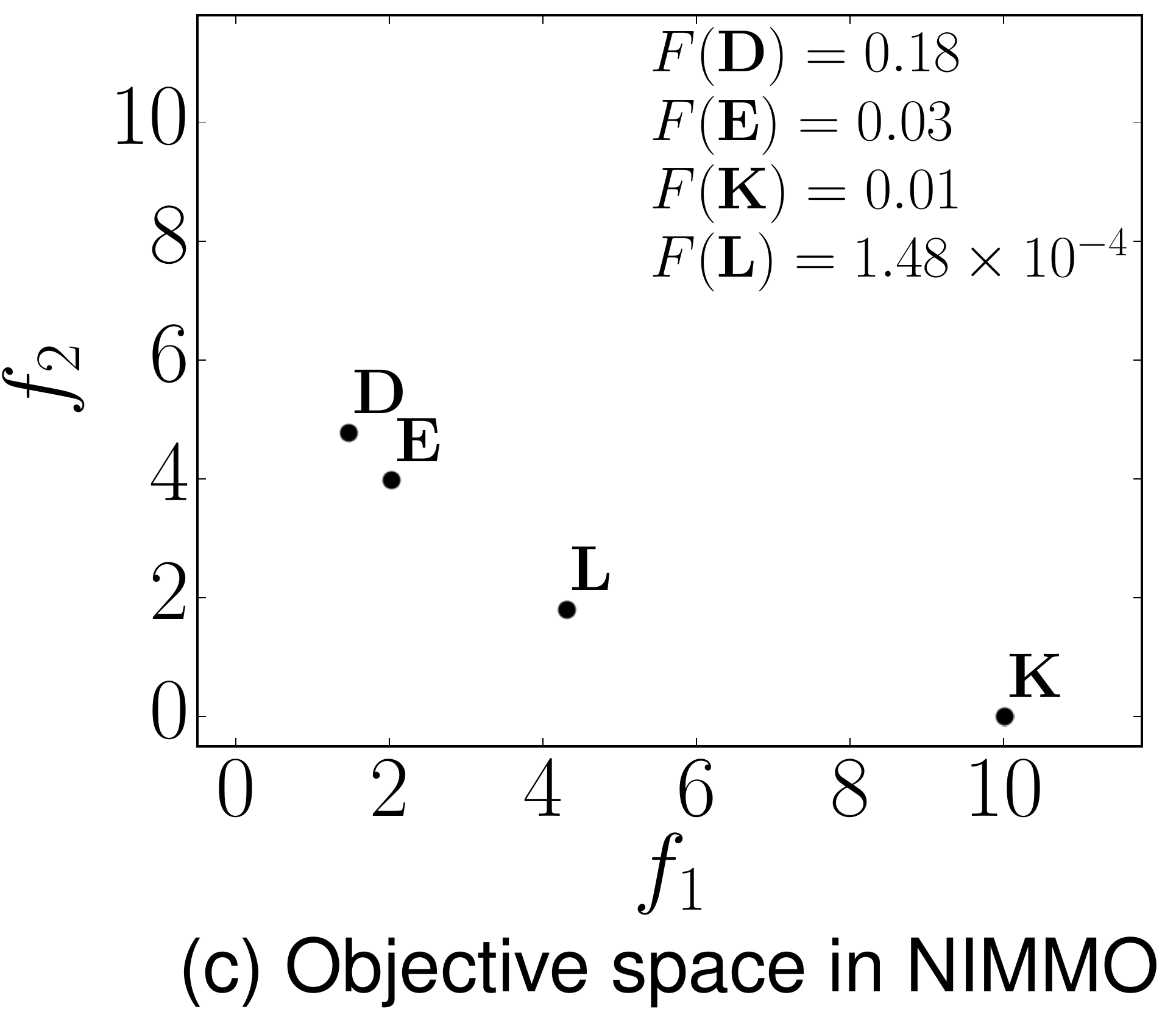}
\caption{
\small
Illustration of the fitness assignment in IBEA and NIMMO with $T=3$ on a two-objective two-variable problem.
}
\label{fig:exapmle_nibea}
  \end{center}
\end{figure*}

We explain the effect of the niching mechanism in NIMMO. 
Fig. \ref{fig:exapmle_nibea} shows examples of fitness assignment results in IBEA and NIMMO on a two-objective two-variable problem.

Figs. \ref{fig:exapmle_nibea}(a) and (b) show the distribution of $12$ individuals $\{\vector{A}, ..., \vector{L}\}$ in the solution and objective spaces, respectively.
The fitness value of each individual in IBEA is also shown, where it can be calculated by replacing $\vector{R}$ with $\vector{P}$ in \eqref{eqn:fitness_ibea}.




In Fig. \ref{fig:exapmle_nibea}, let us assume that $\vector{L}$ is a newly generated child. 
Our next task is to remove a single solution from the 12 individuals.
As seen from Figs. \ref{fig:exapmle_nibea}(a) and (b), $\vector{L}$ and $\vector{H}$ are close to each other in the objective space but far from each other in the solution space.
Therefore, it is desirable that both $\vector{L}$ and $\vector{H}$ remain in the population in the next iteration for multi-modal multi-objective optimization.


Fig. \ref{fig:exapmle_nibea}(b) shows that $\vector{L}$ has the worst fitness value ($0.97$) in IBEA due to its poor diversity in the objective space.
As a result, $\vector{L}$ is discarded from the population after the environmental selection.
Since $\vector{L}$ contributes to the diversity of the population in the solution space more than some individuals, such a removal operation is not appropriate.

NIMMO addresses this issue by using the niching mechanism in the solution space.
If $T=3$, neighborhood individuals of the child $\vector{L}$ in the solution space are $\vector{D}$, $\vector{E}$, and $\vector{K}$.
In NIMMO, the environmental selection is performed only among the four individuals ($\vector{L}$, $\vector{D}$, $\vector{E}$, and $\vector{K}$).
Fig. \ref{fig:exapmle_nibea}(c) shows the distribution of the four individuals in the objective space and their fitness values obtained by \eqref{eqn:fitness_ibea}.
Since the fitness calculation in \eqref{eqn:fitness_ibea} is relative, the fitness value of each individual in NIMMO is different from that in IBEA.


Fig. \ref{fig:exapmle_nibea}(c) indicates that $\vector{D}$ is the worst individual and thus removed from the population in NIMMO.
As a result, the individual $\vector{L}$, which does not survive to the next iteration in IBEA, can remain in the population. 
As shown here, NIMMO can maintain the solution space diversity in the population by using the niching mechanism.



\section{Experimental settings}
\label{sec:experimental_settings}






\subsection{Test problems}
\label{sec:benchmark_problems}

We use the following 13 two-objective MMOPs: the Two-On-One problem \cite{PreussNR06}, the Omni-test problem \cite{DebT08}, the 3 SYM-PART problems \cite{RudolphNP07}, and the 8 MMF problems \cite{LiangYQ16}.
Also, we evaluate the performance of MMEAs on two MMaOPs: the Polygon problem \cite{IshibuchiAN11} and its rotated version named ``RPolygon problem''.
%
The number of Pareto optimal solutions with equivalent quality $N^{\rm same}$ is $2$ for Two-On-One, $9$ for the three SYM-PART problems, and $360$ for the Omni-test problem.
For the eight MMF problems, $N^{\rm same}$ is $2$ for MMF1, MMF2, MMF3, and MMF7, and $N^{\rm same}$ is $4$ for the other problems.
The number of objectives $M$ can be set to an arbitrary number in the Polygon problem.
Thus, the Polygon problem is suitable for evaluating the scalability of MMEAs with respect to $M$.
According to \cite{DebJ14}, we set $M$ of the Polygon problem as follows: $M \in \{3, 5, 8, 10, 15\}$.
In total, we use 23 problem instances.
In our study, $N^{\rm same}$ in the Polygon problem is set to nine.
The number of design variables $D$ is five for the Omni-test problem and two for the other problems.

\subsection{Performance indicators} 
\label{sec:performance_indicators}


We use the inverted generational distance (IGD) \cite{ZitzlerTLFF03}, IGDX \cite{ZhouZJ09}, Pareto sets proximity (PSP) \cite{YueQL17} for performance assessment of MMEAs.
All indicators require a set of reference solutions $\vector{A}^*$.
For all problems (except for the eight MMF problems), we use two reference solution sets $\vector{A}^*_{\rm sol}$ and $\vector{A}^*_{\rm obj}$ of about $5\,000$ Pareto optimal solutions.
While the solutions in $\vector{A}^*_{\rm sol}$ are uniformly distributed in the solution space, those in $\vector{A}^*_{\rm obj}$ are uniformly distributed in the objective space.
We use $\vector{A}^*_{\rm sol}$ for the IGDX and PSP calculations and $\vector{A}^*_{\rm obj}$ for the IGD calculation.
The reference point sets are available at the supplementary website (\url{https://sites.google.com/view/nimmopt/}).
For the eight MMF problems, we use reference solution sets provided by the authors of \cite{YueQL17}.

Below, $\vector{A}$ denotes a set of non-dominated solutions of the final population of an MMEA.
The IGD value \cite{ZitzlerTLFF03} is the average distance from each solution in $\vector{A}^*$ to its nearest non-dominated solution in $\vector{A}$ in the objective space as follows:
%
  \begin{align}
\label{eqn:igd}
{\rm IGD} (\vector{A}) &= \frac{1}{|\vector{A}^*|} \left(\sum_{\vector{z} \in \vector{A}^*} \min_{\vector{x} \in \vector{A}} \Bigl\{ {\rm ED} \bigl(\vector{f}(\vector{x}), \vector{f}(\vector{z}) \bigr) \Bigr\} \right),
 \end{align}
where ${\rm ED}(\vector{x}, \vector{y})$ is the Euclidean distance between $\vector{x}$ and $\vector{y}$.
The IGD metric evaluates the quality of $\vector{A}$ in terms of both convergence to the PF and diversity in the objective space.

While IGD measures the quality of $\vector{A}$ in the objective space, IGDX evaluates how good $\vector{A}$ approximates $\vector{A}^*$ in the solution space.
Similar to the IGD indicator described in \eqref{eqn:igd}, the IGDX value of $\vector{A}$ is given as follows:
%
\begin{align}
\label{eqn:igdx}
{\rm IGDX} (\vector{A}) &= \frac{1}{|\vector{A}^*|} \left(\sum_{\vector{z} \in \vector{A}^*} \min_{\vector{x} \in \vector{A}} \Bigl\{ {\rm ED} \bigl(\vector{x}, \vector{z} \bigr) \Bigr\} \right).
 \end{align}
%

Similar to IGDX, the following PSP evaluates the quality of $\vector{A}$ in the solution space:
%
\begin{align}
\label{eqn:psp}
{\rm PSP} (\vector{A}) &= \frac{{\rm CR} (\vector{A})}{{\rm IGDX} (\vector{A})},
 \end{align}
%

The cover rate (CR) in \eqref{eqn:psp} is given as follows:
%
\begin{align}
\label{eqn:cr}
{\rm CR} (\vector{A}) &= \left(\prod^D_{i=1} \delta_i \right)^{\frac{1}{2D}},
\end{align}

\begin{align}
\label{eqn:cr_delta}
\delta_{i} = \left(\frac{{\rm min}(x^{*,{\rm max}}_i, x^{{\rm max}}_i) - {\rm max}(x^{*,{\rm min}}_i, x^{{\rm min}}_i)}{x^{*,{\rm max}}_i - x^{*,{\rm min}}_i} \right)^2,
\end{align}
where $x^{*,{\rm min}}_i$ and $x^{*,{\rm max}}_i$ in \eqref{eqn:cr_delta} are the minimum and maximum values of the $i$-th design variable in the Pareto solution set, respectively.
If $x^{*,{\rm max}}_i = x^{*,{\rm min}}_i$, $\delta_{i}$ is specified as $\delta_{i} = 1$.
If $x^{{\rm min}}_i \geq x^{*,{\rm max}}_i$ or $x^{{\rm max}}_i \leq x^{*,{\rm min}}_i$, $\delta_{i}$ is specified as $\delta_{i} = 0$.

On the one hand, since a small IGD value indicates that $\vector{A}$ is a good approximation of the PF, the corresponding MMEA is an efficient multi-objective optimizer.
On the other hand, $\vector{A}$ with a small IGDX value and a large PSP value is a good approximation of the Pareto optimal solutions.
Thus, the MMEA finding such an $\vector{A}$ is a high performance multi-modal multi-objective optimization method.

\begin{figure}[t]
\newcommand{\widthvar}{0.325}
  \begin{center} 
\includegraphics[width=\widthvar\textwidth]{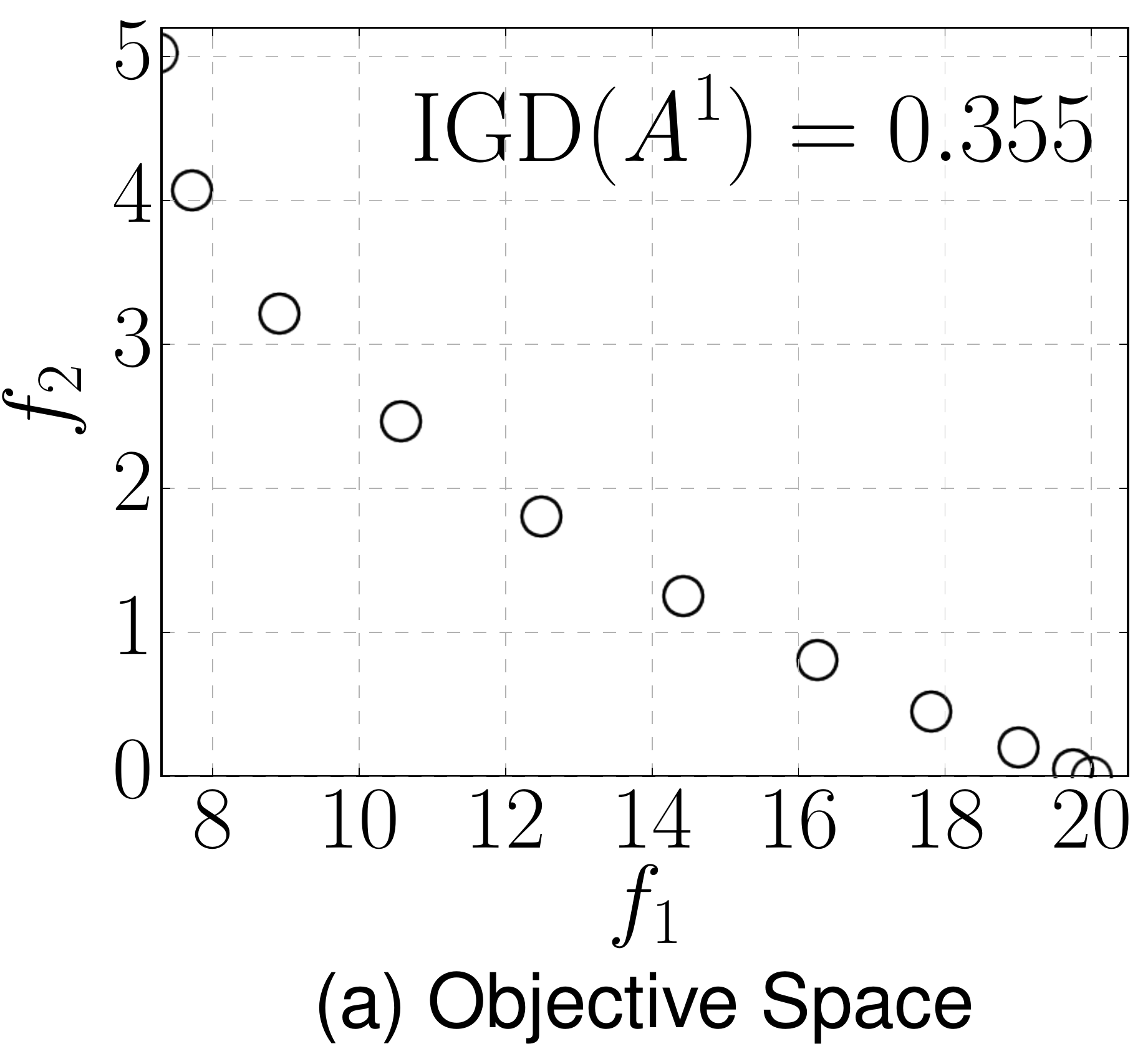}
\includegraphics[width=0.35\textwidth]{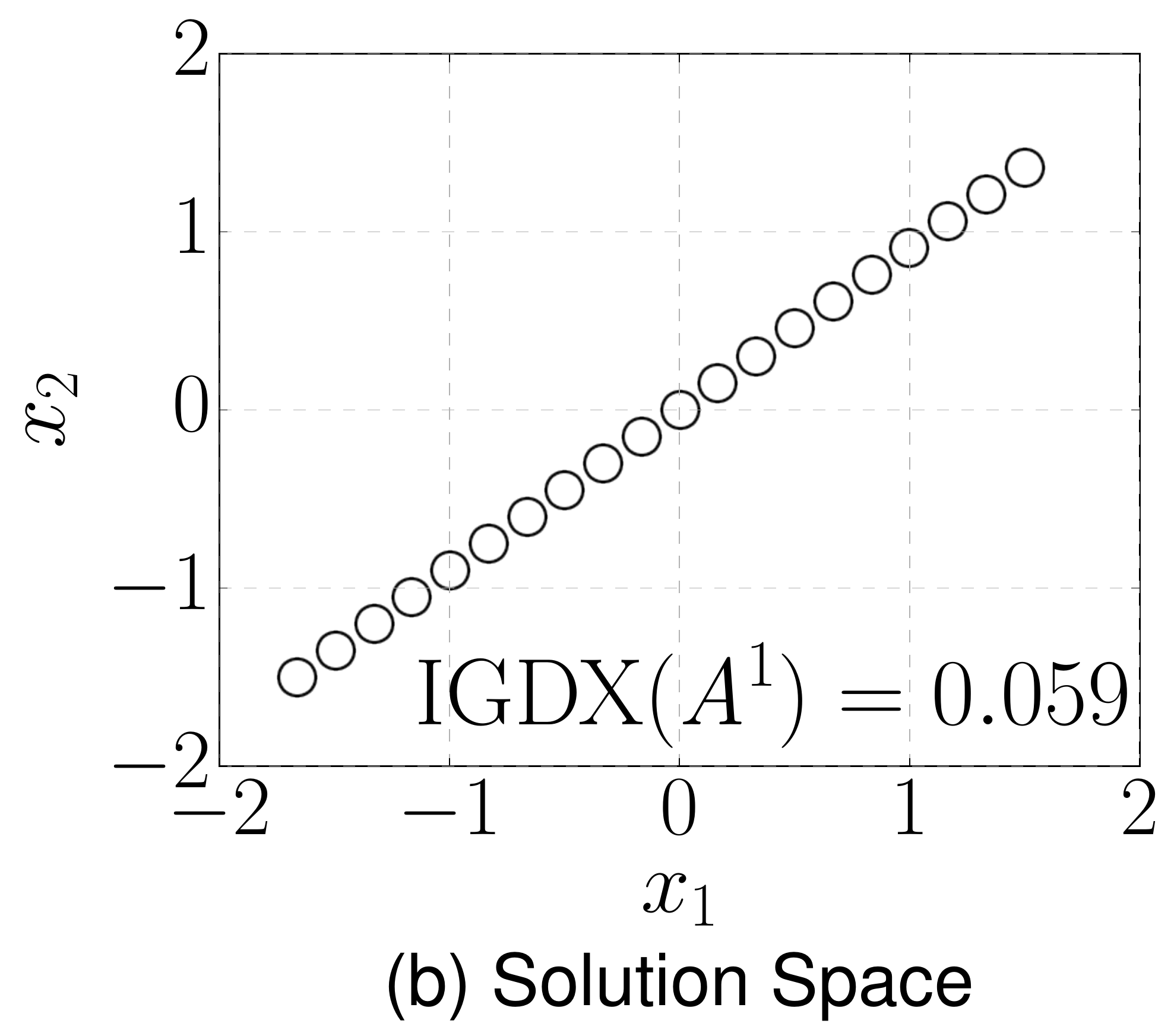}
\\
\includegraphics[width=\widthvar\textwidth]{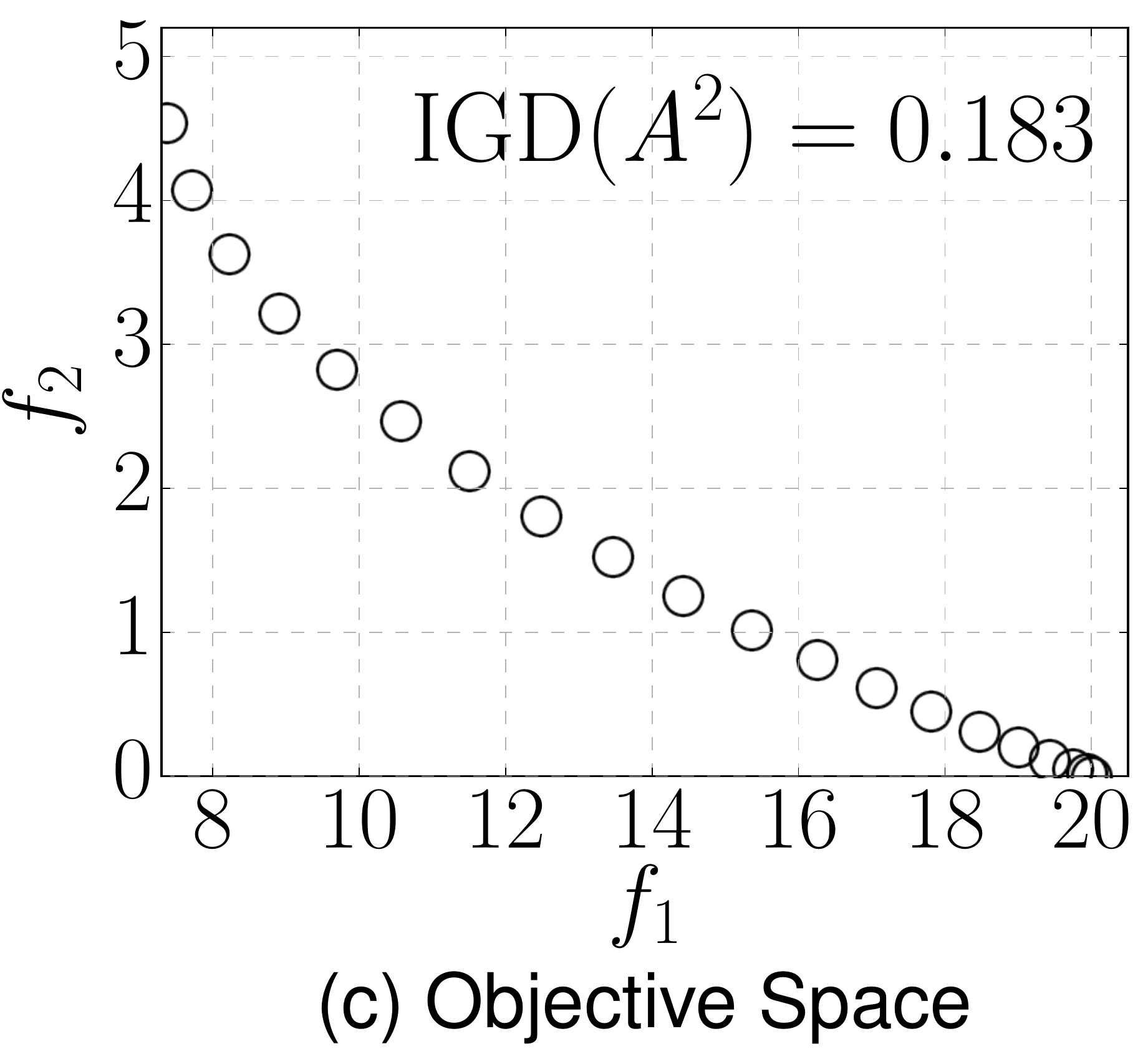}
\includegraphics[width=0.35\textwidth]{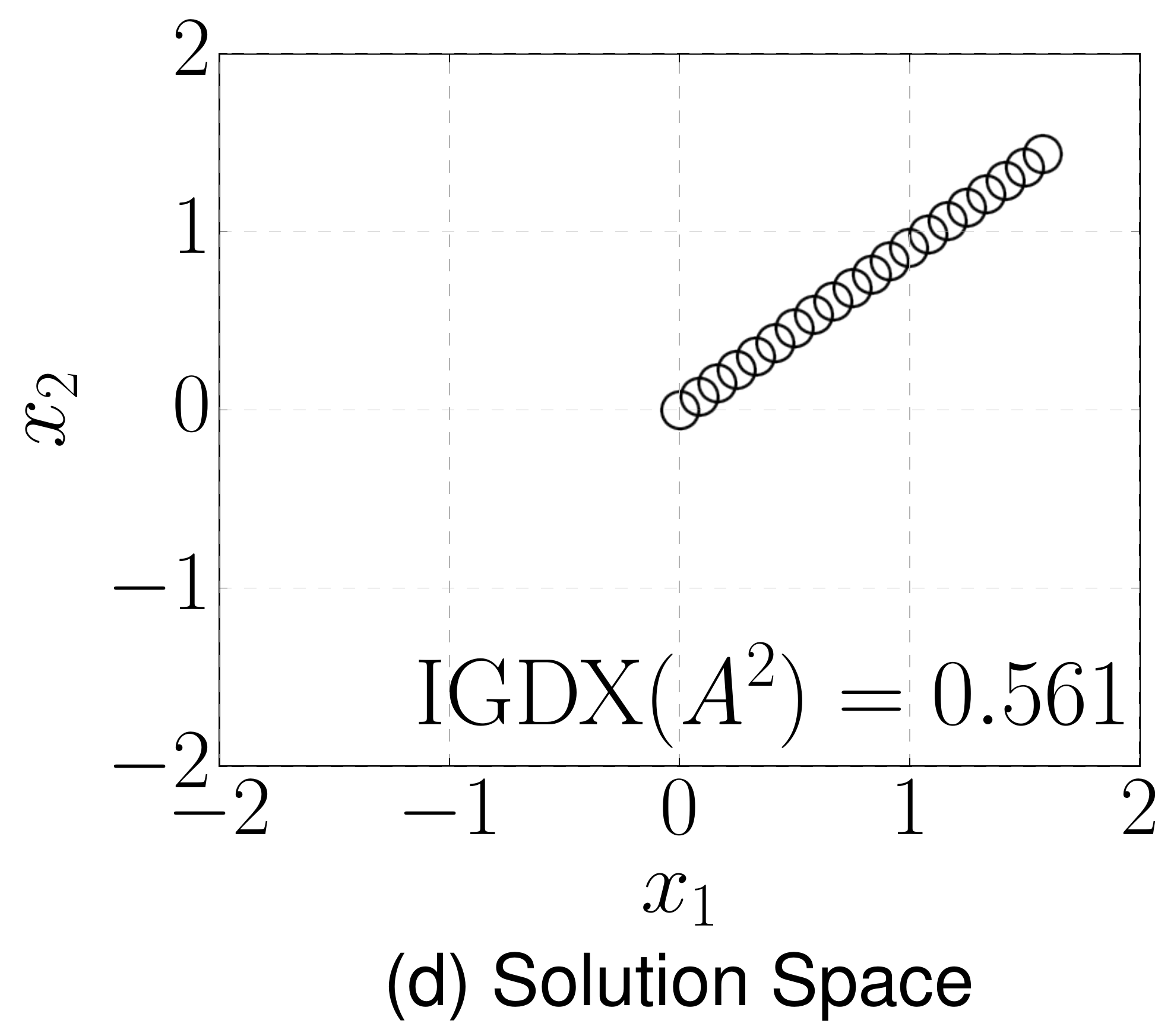}
\caption{
\small
Comparison of two sets of uniformly generated non-dominated solutions $\vector{A}^1$ (upper) and $\vector{A}^2$ (lower) on the Two-On-One problem. 
}
\label{fig:example_two-on-one}
  \end{center}
\end{figure}

Fig. \ref{fig:example_two-on-one} shows a comparison of two sets of non-dominated solutions $\vector{A}^1$ and $\vector{A}^2$ in the objective (Figs. \ref{fig:example_two-on-one}(a) and (c)) and solution (Figs. \ref{fig:example_two-on-one}(b) and (d)) spaces on the Two-On-One problem, where $\vector{A}^1$ and $\vector{A}^2$ are evenly generated on the PS, and $|\vector{A}^1| = |\vector{A}^2| = 20$.
The IGD and IGDX values of $\vector{A}^1$ and $\vector{A}^2$ are also shown in Fig. \ref{fig:example_two-on-one}.
In the Two-On-One problem, there are two equivalent Pareto optimal solution sets that are symmetrical with respect to the origin.

Figs. \ref{fig:example_two-on-one}(a) and (c) show that $\vector{A}^2$ covers the PF more densely than $\vector{A}^1$.
However, $\vector{A}^2$ does not cover the whole PS while $\vector{A}^1$ covers the whole PS as shown in Figs. \ref{fig:example_two-on-one}(b) and (d).
Thus, $\vector{A}^1$ and $\vector{A}^2$ are good approximations of the PS and the PF, respectively.
While the IGD value of $\vector{A}^2$ is approximately two times better than that of $\vector{A}^1$, 
the IGDX value of $\vector{A}^2$ is approximately $10$ times worse than that of $\vector{A}^1$.
It is likely that if a set of non-dominated solutions has a very good IGDX value (e.g., $\vector{A}^1$), its IGD value is good.
However, a set of non-dominated solutions with a very good IGD value (e.g., $\vector{A}^2$) can have a bad IGDX value.





\subsection{MMEAs}
\label{sec:settings_EMOAs}

We compare NIMMO with the following MMEAs and MOEAs: Omni-optimizer \cite{DebT08}, MO\_Ring\_PSO\_SCD \cite{YueQL17}, TriMOEA-TA\&R \cite{LiuYG18}, IBEA \cite{ZitzlerK04}, NSGA-II \cite{DebAPM02}, MOEA/D \cite{ZhangL07}, NSGA-III \cite{DebJ14}, and MOEA/DD \cite{LiDZK15}.
NSGA-II, IBEA, and MOEA/D are representative dominance-, indicator-, and decomposition-based MOEAs, respectively.
NSGA-III and MOEA/DD are well-known MOEAs for MaOPs. 
Omni-optimizer is a representative MMEA for MMOPs.
MO\_Ring\_PSO\_SCD is a recently proposed multi-modal multi-objective PSO algorithm.
TriMOEA-TA\&R is the latest MMEA.







We used available source code through the Internet for algorithm implementation.
%
 A set of weight/reference vectors of MOEA/D, MOEA/DD, and NSGA-III were generated using the simplex-lattice design for $M \leq 5$ and its two-layered version \cite{DebJ14} for $M \geq 8$.
 For all algorithms (except for NSGA-II and Omni-optimizer), the population size $\mu$ was set to $200$, $210$, $210$, $156$, $210$, $230$, and $135$ for $M=2$, $3$, $5$, $8$, $9$, $10$, and $15$, respectively.
The number of weight vectors cannot be specified arbitrarily due to the combinatorial nature of the simplex-lattice design.
For this reason, we selected the population size $\mu$ such that $\mu$ is close to $200$ as possible.
 Since $\mu$ values in NSGA-II and Omni-optimizer must be multiples of two and four respectively, they were set to the closest multiples of two and four  which are larger than the above $\mu$ values respectively.
The number of maximum function evaluations was set to $10\,000$, and $31$ runs were performed.



We set $T$ of NIMMO to $\lfloor 0.1 \mu \rfloor$.
To investigate the contribution of the niching mechanism of NIMMO, we evaluated the performance of a steady-state IBEA with $I_{\epsilon+}$ and the randomized mating selection, which is equivalent to NIMMO with $T=\mu$.
Since our preliminary results show that this version of IBEA outperforms the original IBEA regarding the IGD, IGDX, and PSP indicators, we show only the results of the former.
%
According to \cite{LiDZK15}, the neighborhood size of MOEA/D and MOEA/DD was set to $20$.
We used the Tchebycheff and PBI functions for MOEA/D and MOEA/DD, respectively.
%
The SBX crossover and the polynomial mutation were used in all MMEAs and MOEAs, except for MO\_Ring\_PSO\_SCD. 
Their control parameters were set as follows: $p_c = 1$, $\eta_c = 20$, $p_m = 1/D$, and $\eta_m = 20$.
All the control parameters of MO\_Ring\_PSO\_SCD and TriMOEA-TA\&R were set according to \cite{YueQL17,LiuYG18}.




\begin{landscape}
\begin{table*}[t]
\renewcommand{\arraystretch}{1}
\centering
\caption{\small 
  Results of the 9 algorithms on the 23 test problem instances. The mean IGD values for 31 runs are shown. The numbers in parenthesis indicate the ranks of the nine algorithms.
  }
  \label{tab:comparison_8methods_igd}
{\scriptsize
  \scalebox{1}[1]{
\begin{tabular}{ccccccccccc}
  \midrule
   & \raisebox{0.5em}{NIMMO} & \shortstack{TriMOEA-\\TA\&R} & \shortstack{MO\_Ring\_\\PSO\_SCD} & \shortstack{Omni-\\optimizer} & \raisebox{0.5em}{IBEA} & \raisebox{0.5em}{NSGA-II} & \raisebox{0.5em}{MOEA/D} & \raisebox{0.5em}{NSGA-III} & \raisebox{0.5em}{MOEA/DD}\\
  \toprule
Two-On-One & 0.0250 (3) & 0.0627$-$ (9) & 0.0348$-$ (7) & 0.0254$-$ (5) & {\scriptsize \textbf{0.0233}}$+$ (1) & {\em 0.0245}$\approx$ (2) & 0.0264$\approx$ (6) & 0.0251$\approx$ (4) & 0.0429$-$ (8) \\
Omni-test & 0.0769 (7) & 0.1365$-$ (8) & 0.2957$-$ (9) & {\em 0.0268}$+$ (2) & 0.0315$+$ (3) & {\scriptsize \textbf{0.0266}}$+$ (1) & 0.0622$+$ (5) & 0.0525$+$ (4) & 0.0700$\approx$ (6) \\
SYM-PART1 & 0.0246 (4) & 0.0379$-$ (7) & 0.0386$-$ (9) & {\em 0.0137}$+$ (2) & 0.0169$+$ (3) & {\scriptsize \textbf{0.0104}}$+$ (1) & 0.0341$-$ (6) & 0.0262$-$ (5) & 0.0383$-$ (8) \\
SYM-PART2 & 0.0365 (6) & 0.0396$-$ (7) & 0.0446$-$ (9) & {\em 0.0183}$+$ (2) & 0.0199$+$ (3) & {\scriptsize \textbf{0.0120}}$+$ (1) & 0.0361$\approx$ (5) & 0.0284$+$ (4) & 0.0413$-$ (8) \\
SYM-PART3 & 0.0298 (4) & 0.0798$-$ (9) & 0.0493$-$ (8) & 0.0413$\approx$ (7) & {\em 0.0196}$+$ (2) & {\scriptsize \textbf{0.0118}}$+$ (1) & 0.0369$-$ (5) & 0.0279$\approx$ (3) & 0.0389$-$ (6) \\
\midrule
MMF1 & 0.0050 (7) & 0.0092$\approx$ (8) & 0.0037$+$ (5) & 0.0033$+$ (3) & 0.0043$+$ (6) & {\scriptsize \textbf{0.0028}}$+$ (1) & 0.0152$-$ (9) & 0.0034$+$ (4) & {\em 0.0030}$+$ (2) \\
MMF2 & 0.0216 (5) & 0.0294$-$ (8) & 0.0216$\approx$ (6) & {\em 0.0178}$+$ (2) & 0.0286$-$ (7) & 0.0179$+$ (3) & 0.1246$-$ (9) & 0.0181$+$ (4) & {\scriptsize \textbf{0.0118}}$+$ (1) \\
MMF3 & 0.0190 (6) & 0.0230$-$ (8) & 0.0154$\approx$ (5) & 0.0134$+$ (3) & 0.0215$\approx$ (7) & {\em 0.0114}$+$ (2) & 0.0966$-$ (9) & 0.0144$+$ (4) & {\scriptsize \textbf{0.0104}}$+$ (1) \\
MMF4 & 0.0048 (7) & 0.0496$-$ (9) & 0.0037$+$ (5) & 0.0029$+$ (4) & 0.0057$-$ (8) & 0.0026$+$ (3) & 0.0047$+$ (6) & {\em 0.0023}$+$ (2) & {\scriptsize \textbf{0.0020}}$+$ (1) \\
MMF5 & 0.0039 (6) & 0.0087$-$ (8) & 0.0038$+$ (5) & 0.0032$+$ (3) & 0.0042$-$ (7) & {\scriptsize \textbf{0.0028}}$+$ (1) & 0.0104$-$ (9) & 0.0034$+$ (4) & {\em 0.0029}$+$ (2) \\
MMF6 & 0.0044 (7) & 0.0087$-$ (8) & 0.0035$+$ (5) & 0.0031$+$ (4) & 0.0043$+$ (6) & {\em 0.0027}$+$ (2) & 0.0108$-$ (9) & 0.0031$+$ (3) & {\scriptsize \textbf{0.0026}}$+$ (1) \\
MMF7 & 0.0030 (7) & 0.0064$-$ (9) & 0.0037$-$ (8) & 0.0029$\approx$ (6) & 0.0027$+$ (4) & 0.0026$+$ (3) & 0.0027$+$ (5) & {\em 0.0025}$+$ (2) & {\scriptsize \textbf{0.0023}}$+$ (1) \\
MMF8 & 0.0092 (8) & 0.0113$-$ (9) & 0.0047$+$ (6) & 0.0031$+$ (4) & 0.0079$+$ (7) & {\em 0.0026}$+$ (2) & 0.0044$+$ (5) & 0.0026$+$ (3) & {\scriptsize \textbf{0.0022}}$+$ (1) \\
\midrule
3-Polygon & {\em 0.0025} (2) & 0.0040$-$ (8) & 0.0034$-$ (6) & 0.0028$-$ (4) & {\scriptsize \textbf{0.0022}}$+$ (1) & 0.0028$-$ (3) & 0.0034$-$ (7) & 0.0028$-$ (5) & 0.0042$-$ (9) \\
5-Polygon & {\em 0.0044} (2) & 0.0149$-$ (9) & 0.0057$-$ (6) & 0.0051$-$ (4) & {\scriptsize \textbf{0.0036}}$+$ (1) & 0.0052$-$ (5) & 0.0051$-$ (3) & 0.0057$-$ (7) & 0.0136$-$ (8) \\
8-Polygon & {\em 0.0069} (2) & 0.0180$-$ (8) & 0.0092$-$ (7) & 0.0082$-$ (4) & {\scriptsize \textbf{0.0057}}$+$ (1) & 0.0083$-$ (5) & 0.0084$-$ (6) & 0.0081$-$ (3) & 0.0279$-$ (9) \\
10-Polygon & {\em 0.0064} (2) & 0.0204$-$ (8) & 0.0087$-$ (7) & 0.0074$-$ (3) & {\scriptsize \textbf{0.0053}}$+$ (1) & 0.0075$-$ (4) & 0.0080$-$ (6) & 0.0077$-$ (5) & 0.0348$-$ (9) \\
15-Polygon & {\em 0.0106} (2) & 0.0211$-$ (8) & 0.0136$-$ (7) & 0.0122$-$ (4) & {\scriptsize \textbf{0.0085}}$+$ (1) & 0.0123$-$ (5) & 0.0106$\approx$ (3) & 0.0128$-$ (6) & 0.0406$-$ (9) \\
3-RPolygon & {\em 0.0025} (2) & 0.0046$-$ (9) & 0.0034$-$ (7) & 0.0028$-$ (4) & {\scriptsize \textbf{0.0022}}$+$ (1) & 0.0028$-$ (3) & 0.0034$-$ (6) & 0.0029$-$ (5) & 0.0043$-$ (8) \\
5-RPolygon & {\em 0.0044} (2) & 0.0149$-$ (9) & 0.0058$-$ (7) & 0.0052$-$ (4) & {\scriptsize \textbf{0.0036}}$+$ (1) & 0.0052$-$ (5) & 0.0051$-$ (3) & 0.0058$-$ (6) & 0.0133$-$ (8) \\
8-RPolygon & {\em 0.0069} (2) & 0.0190$-$ (8) & 0.0093$-$ (7) & 0.0083$-$ (3) & {\scriptsize \textbf{0.0057}}$+$ (1) & 0.0083$-$ (4) & 0.0086$-$ (6) & 0.0083$-$ (5) & 0.0279$-$ (9) \\
10-RPolygon & {\em 0.0064} (2) & 0.0185$-$ (8) & 0.0086$-$ (7) & 0.0075$-$ (3) & {\scriptsize \textbf{0.0053}}$+$ (1) & 0.0075$-$ (4) & 0.0082$-$ (6) & 0.0078$-$ (5) & 0.0321$-$ (9) \\
15-RPolygon & {\em 0.0105} (2) & 0.0226$-$ (8) & 0.0140$-$ (7) & 0.0125$-$ (5) & {\scriptsize \textbf{0.0085}}$+$ (1) & 0.0123$-$ (4) & 0.0107$-$ (3) & 0.0131$-$ (6) & 0.0413$-$ (9) \\
\midrule
\end{tabular}
}
}
\end{table*}
\end{landscape}

\begin{landscape}
\begin{table*}[t]
\renewcommand{\arraystretch}{1}
\centering
\caption{\small
  Results of the 9 algorithms on the 23 test problem instances. The mean IGDX values for 31 runs are shown.   
  }
  \label{tab:comparison_8methods_igdx}
{\scriptsize
  \scalebox{1}[1]{
\begin{tabular}{ccccccccccc}
  \midrule
   & \raisebox{0.5em}{NIMMO} & \shortstack{TriMOEA-\\TA\&R} & \shortstack{MO\_Ring\_\\PSO\_SCD} & \shortstack{Omni-\\optimizer} & \raisebox{0.5em}{IBEA} & \raisebox{0.5em}{NSGA-II} & \raisebox{0.5em}{MOEA/D} & \raisebox{0.5em}{NSGA-III} & \raisebox{0.5em}{MOEA/DD}\\
  \toprule
Two-On-One & {\scriptsize \textbf{0.0226}} (1) & 0.5699$-$ (9) & 0.0296$-$ (3) & {\em 0.0261}$-$ (2) & 0.0548$-$ (6) & 0.0405$-$ (4) & 0.2209$-$ (8) & 0.0461$-$ (5) & 0.0727$-$ (7) \\
Omni-test & {\scriptsize \textbf{1.6052}} (1) & 5.2291$-$ (9) & 2.4142$-$ (6) & {\em 1.7443}$-$ (2) & 2.4403$-$ (7) & 1.9110$-$ (4) & 3.5174$-$ (8) & 1.8711$-$ (3) & 2.1113$-$ (5) \\
SYM-PART1 & {\scriptsize \textbf{0.0530}} (1) & 0.7040$-$ (6) & {\em 0.1673}$-$ (2) & 0.3452$-$ (3) & 3.8899$-$ (8) & 1.0533$-$ (7) & 6.6705$-$ (9) & 0.6525$-$ (5) & 0.5588$-$ (4) \\
SYM-PART2 & {\scriptsize \textbf{0.0838}} (1) & 9.8161$-$ (9) & {\em 0.2303}$-$ (2) & 0.4325$-$ (3) & 3.7481$-$ (7) & 1.8697$-$ (6) & 7.5890$-$ (8) & 1.5404$-$ (5) & 1.5147$-$ (4) \\
SYM-PART3 & {\scriptsize \textbf{0.1418}} (1) & 5.5682$-$ (8) & 0.8697$-$ (3) & {\em 0.8538}$-$ (2) & 3.8662$-$ (7) & 2.3685$-$ (6) & 6.6383$-$ (9) & 2.3069$-$ (5) & 2.2809$-$ (4) \\
\midrule
MMF1 & 0.0611 (3) & 0.1369$-$ (8) & {\scriptsize \textbf{0.0490}}$+$ (1) & {\em 0.0529}$+$ (2) & 0.0930$-$ (7) & 0.0665$-$ (5) & 0.2606$-$ (9) & 0.0765$-$ (6) & 0.0657$\approx$ (4) \\
MMF2 & 0.0440 (3) & 0.0697$-$ (7) & {\em 0.0430}$\approx$ (2) & 0.0614$-$ (6) & 0.1219$-$ (8) & 0.0531$\approx$ (4) & 0.3290$-$ (9) & 0.0580$\approx$ (5) & {\scriptsize \textbf{0.0373}}$+$ (1) \\
MMF3 & 0.0440 (4) & 0.0747$-$ (7) & {\scriptsize \textbf{0.0282}}$+$ (1) & 0.0423$\approx$ (3) & 0.0841$-$ (8) & 0.0448$\approx$ (5) & 0.2359$-$ (9) & 0.0488$\approx$ (6) & {\em 0.0315}$+$ (2) \\
MMF4 & 0.0372 (3) & 0.1628$\approx$ (8) & {\scriptsize \textbf{0.0275}}$+$ (1) & {\em 0.0332}$+$ (2) & 0.0820$-$ (7) & 0.0523$-$ (6) & 0.2741$-$ (9) & 0.0487$-$ (4) & 0.0519$-$ (5) \\
MMF5 & {\scriptsize \textbf{0.0846}} (1) & 0.1993$-$ (8) & {\em 0.0869}$\approx$ (2) & 0.0980$-$ (3) & 0.1618$-$ (7) & 0.1187$-$ (4) & 0.4480$-$ (9) & 0.1263$-$ (6) & 0.1191$-$ (5) \\
MMF6 & 0.1037 (3) & 0.1641$-$ (8) & {\scriptsize \textbf{0.0740}}$+$ (1) & {\em 0.0835}$+$ (2) & 0.1471$-$ (7) & 0.1119$-$ (6) & 0.3753$-$ (9) & 0.1113$\approx$ (5) & 0.1054$\approx$ (4) \\
MMF7 & {\scriptsize \textbf{0.0216}} (1) & 0.0591$-$ (8) & {\em 0.0263}$-$ (2) & 0.0267$-$ (3) & 0.0435$-$ (7) & 0.0383$-$ (5) & 0.1231$-$ (9) & 0.0414$-$ (6) & 0.0365$-$ (4) \\
MMF8 & 0.2064 (3) & 0.4460$-$ (5) & {\scriptsize \textbf{0.0680}}$+$ (1) & {\em 0.1088}$+$ (2) & 0.7538$-$ (8) & 0.4556$-$ (6) & 2.2762$-$ (9) & 0.5011$-$ (7) & 0.4089$-$ (4) \\
\midrule
3-Polygon & {\scriptsize \textbf{0.0056}} (1) & {\em 0.0063}$-$ (2) & 0.0091$-$ (4) & 0.0083$-$ (3) & 0.0133$-$ (7) & 0.0093$-$ (5) & 0.0496$-$ (9) & 0.0097$-$ (6) & 0.0232$-$ (8) \\
5-Polygon & {\scriptsize \textbf{0.0070}} (1) & 0.0162$-$ (6) & 0.0113$-$ (4) & 0.0110$-$ (3) & {\em 0.0109}$-$ (2) & 0.0119$-$ (5) & 0.0584$-$ (8) & 0.0169$-$ (7) & 0.0883$-$ (9) \\
8-Polygon & {\scriptsize \textbf{0.0089}} (1) & {\em 0.0136}$-$ (2) & 0.0143$-$ (4) & 0.0140$-$ (3) & 0.0151$-$ (5) & 0.0189$-$ (7) & 0.1059$-$ (8) & 0.0187$-$ (6) & 0.1349$-$ (9) \\
10-Polygon & {\scriptsize \textbf{0.0072}} (1) & 0.0123$-$ (5) & 0.0120$-$ (4) & 0.0112$-$ (3) & {\em 0.0108}$-$ (2) & 0.0123$-$ (6) & 0.0799$-$ (8) & 0.0139$-$ (7) & 0.0858$-$ (9) \\
15-Polygon & {\scriptsize \textbf{0.0100}} (1) & {\em 0.0122}$-$ (2) & 0.0156$-$ (4) & 0.0155$-$ (3) & 0.0212$-$ (6) & 0.0201$-$ (5) & 0.0949$-$ (8) & 0.0286$-$ (7) & 0.1364$-$ (9) \\
3-RPolygon & {\scriptsize \textbf{0.0059}} (1) & 0.0295$-$ (6) & 0.0090$-$ (3) & {\em 0.0085}$-$ (2) & 0.0339$-$ (7) & 0.0194$-$ (4) & 0.1652$-$ (9) & 0.0260$-$ (5) & 0.0463$-$ (8) \\
5-RPolygon & {\scriptsize \textbf{0.0074}} (1) & 0.0400$-$ (7) & 0.0113$-$ (3) & {\em 0.0110}$-$ (2) & 0.0289$-$ (5) & 0.0196$-$ (4) & 0.1798$-$ (9) & 0.0348$-$ (6) & 0.1021$-$ (8) \\
8-RPolygon & {\scriptsize \textbf{0.0093}} (1) & 0.0747$-$ (6) & 0.0144$-$ (3) & {\em 0.0138}$-$ (2) & 0.0426$-$ (5) & 0.0402$-$ (4) & 0.1662$-$ (9) & 0.0788$-$ (7) & 0.1454$-$ (8) \\
10-RPolygon & {\scriptsize \textbf{0.0076}} (1) & 0.0404$-$ (7) & 0.0118$-$ (3) & {\em 0.0112}$-$ (2) & 0.0257$-$ (5) & 0.0228$-$ (4) & 0.1418$-$ (9) & 0.0289$-$ (6) & 0.1182$-$ (8) \\
15-RPolygon & {\scriptsize \textbf{0.0104}} (1) & 0.0667$-$ (6) & {\em 0.0154}$-$ (2) & 0.0166$-$ (3) & 0.0408$-$ (5) & 0.0373$-$ (4) & 0.1542$-$ (9) & 0.0952$-$ (7) & 0.1406$-$ (8) \\
\midrule
\end{tabular}
}
}
\end{table*}
\end{landscape}

\begin{landscape}
\begin{table*}[t]
\renewcommand{\arraystretch}{1}
\centering
\caption{\small
  Results of the 9 algorithms on the 23 test problem instances. The mean PSP values for 31 runs are shown.   
  }
  \label{tab:comparison_8methods_psp}
{\scriptsize
  \scalebox{1}[1]{
\begin{tabular}{ccccccccccc}
  \midrule
   & \raisebox{0.5em}{NIMMO} & \shortstack{TriMOEA-\\TA\&R} & \shortstack{MO\_Ring\_\\PSO\_SCD} & \shortstack{Omni-\\optimizer} & \raisebox{0.5em}{IBEA} & \raisebox{0.5em}{NSGA-II} & \raisebox{0.5em}{MOEA/D} & \raisebox{0.5em}{NSGA-III} & \raisebox{0.5em}{MOEA/DD}\\
  \toprule
Two-On-One & {\scriptsize \textbf{40.09}} (1) & 0.13$-$ (9) & 32.72$-$ (3) & {\em 36.84}$-$ (2) & 15.58$-$ (6) & 24.19$-$ (4) & 3.56$-$ (8) & 20.89$-$ (5) & 13.83$-$ (7) \\
Omni-test & {\scriptsize \textbf{0.49}} (1) & 0.00$-$ (9) & 0.33$-$ (5) & {\em 0.42}$-$ (2) & 0.12$-$ (7) & 0.33$-$ (4) & 0.00$-$ (8) & 0.35$-$ (3) & 0.22$-$ (6) \\
SYM-PART1 & {\scriptsize \textbf{18.73}} (1) & {\em 10.35}$-$ (2) & 5.98$-$ (4) & 7.63$-$ (3) & 1.31$-$ (8) & 4.73$-$ (7) & 0.08$-$ (9) & 5.32$-$ (5) & 5.14$-$ (6) \\
SYM-PART2 & {\scriptsize \textbf{11.47}} (1) & 0.00$-$ (9) & 4.81$-$ (3) & {\em 4.83}$-$ (2) & 0.13$-$ (7) & 0.84$-$ (6) & 0.01$-$ (8) & 0.93$-$ (5) & 1.21$-$ (4) \\
SYM-PART3 & {\scriptsize \textbf{9.48}} (1) & 0.02$-$ (8) & 1.28$-$ (3) & {\em 2.15}$-$ (2) & 0.12$-$ (7) & 0.23$-$ (6) & 0.01$-$ (9) & 0.25$-$ (5) & 0.75$-$ (4) \\
\midrule
MMF1 & 16.22 (3) & 11.64$-$ (7) & {\scriptsize \textbf{19.89}}$+$ (1) & {\em 18.34}$+$ (2) & 10.21$-$ (8) & 14.67$-$ (5) & 2.02$-$ (9) & 12.59$-$ (6) & 14.77$-$ (4) \\
MMF2 & {\em 20.61} (2) & 13.73$-$ (7) & 19.36$\approx$ (4) & 16.87$\approx$ (6) & 7.48$-$ (8) & 20.07$\approx$ (3) & 0.85$-$ (9) & 18.17$\approx$ (5) & {\scriptsize \textbf{25.75}}$\approx$ (1) \\
MMF3 & 19.90 (5) & 12.22$-$ (7) & {\scriptsize \textbf{29.54}}$+$ (1) & 21.72$\approx$ (3) & 9.84$-$ (8) & 20.62$\approx$ (4) & 0.96$-$ (9) & 18.38$\approx$ (6) & {\em 25.28}$+$ (2) \\
MMF4 & 27.06 (3) & 22.98$\approx$ (4) & {\scriptsize \textbf{35.25}}$+$ (1) & {\em 29.99}$+$ (2) & 12.00$-$ (8) & 19.30$-$ (7) & 2.42$-$ (9) & 20.12$-$ (5) & 19.60$-$ (6) \\
MMF5 & {\scriptsize \textbf{11.68}} (1) & 7.72$-$ (6) & {\em 11.20}$-$ (2) & 10.01$-$ (3) & 5.97$-$ (8) & 8.11$-$ (4) & 0.78$-$ (9) & 7.35$-$ (7) & 7.99$-$ (5) \\
MMF6 & 9.09 (5) & 9.49$+$ (3) & {\scriptsize \textbf{13.16}}$+$ (1) & {\em 11.58}$+$ (2) & 6.09$-$ (8) & 8.80$\approx$ (6) & 0.77$-$ (9) & 8.67$\approx$ (7) & 9.14$\approx$ (4) \\
MMF7 & {\scriptsize \textbf{43.02}} (1) & 18.86$-$ (8) & 36.57$-$ (3) & {\em 36.68}$-$ (2) & 19.59$-$ (7) & 25.25$-$ (5) & 5.24$-$ (9) & 21.52$-$ (6) & 25.88$-$ (4) \\
MMF8 & 4.97 (3) & 2.22$-$ (4) & {\scriptsize \textbf{14.29}}$+$ (1) & {\em 8.84}$+$ (2) & 0.73$-$ (8) & 1.48$-$ (6) & 0.11$-$ (9) & 1.16$-$ (7) & 1.93$-$ (5) \\
\midrule
3-Polygon & {\scriptsize \textbf{178.71}} (1) & {\em 158.39}$-$ (2) & 109.85$-$ (4) & 120.17$-$ (3) & 104.94$-$ (6) & 108.30$-$ (5) & 31.29$-$ (9) & 102.75$-$ (7) & 50.72$-$ (8) \\
5-Polygon & {\scriptsize \textbf{141.30}} (1) & 61.55$-$ (7) & 87.99$-$ (4) & {\em 90.96}$-$ (2) & 90.83$-$ (3) & 84.30$-$ (5) & 32.56$-$ (8) & 68.18$-$ (6) & 10.60$-$ (9) \\
8-Polygon & {\scriptsize \textbf{111.35}} (1) & {\em 72.69}$-$ (2) & 69.98$-$ (4) & 71.09$-$ (3) & 68.43$-$ (5) & 58.88$-$ (7) & 10.85$-$ (8) & 60.18$-$ (6) & 5.86$-$ (9) \\
10-Polygon & {\scriptsize \textbf{138.23}} (1) & 77.00$-$ (6) & 83.24$-$ (4) & 89.26$-$ (3) & {\em 92.51}$-$ (2) & 81.59$-$ (5) & 16.26$-$ (8) & 76.29$-$ (7) & 10.85$-$ (9) \\
15-Polygon & {\scriptsize \textbf{97.52}} (1) & {\em 79.39}$-$ (2) & 63.86$-$ (4) & 63.97$-$ (3) & 54.63$-$ (5) & 52.91$-$ (6) & 13.80$-$ (8) & 42.42$-$ (7) & 5.72$-$ (9) \\
3-RPolygon & {\scriptsize \textbf{166.54}} (1) & 49.63$-$ (6) & 107.81$-$ (3) & {\em 113.59}$-$ (2) & 41.75$-$ (7) & 65.24$-$ (4) & 2.74$-$ (9) & 53.17$-$ (5) & 24.66$-$ (8) \\
5-RPolygon & {\scriptsize \textbf{132.04}} (1) & 28.83$-$ (7) & 86.99$-$ (3) & {\em 88.22}$-$ (2) & 45.59$-$ (5) & 54.22$-$ (4) & 2.47$-$ (9) & 31.70$-$ (6) & 6.34$-$ (8) \\
8-RPolygon & {\scriptsize \textbf{104.28}} (1) & 13.01$-$ (7) & 67.94$-$ (3) & {\em 70.04}$-$ (2) & 26.01$-$ (5) & 26.76$-$ (4) & 2.07$-$ (9) & 14.12$-$ (6) & 3.02$-$ (8) \\
10-RPolygon & {\scriptsize \textbf{126.33}} (1) & 25.64$-$ (7) & 82.54$-$ (3) & {\em 86.48}$-$ (2) & 45.27$-$ (5) & 49.44$-$ (4) & 3.66$-$ (9) & 41.00$-$ (6) & 5.06$-$ (8) \\
15-RPolygon & {\scriptsize \textbf{90.97}} (1) & 15.45$-$ (6) & {\em 62.65}$-$ (2) & 58.17$-$ (3) & 26.24$-$ (4) & 24.72$-$ (5) & 2.86$-$ (9) & 7.56$-$ (7) & 2.94$-$ (8) \\
\midrule
\end{tabular}
}
}
\end{table*}
\end{landscape}

\section{Experimental results}
\label{sec:experimental_results}


\begin{figure}[htp]
\newcommand{\widthvar}{0.3}
  \begin{center} 
\includegraphics[width=\widthvar\textwidth]{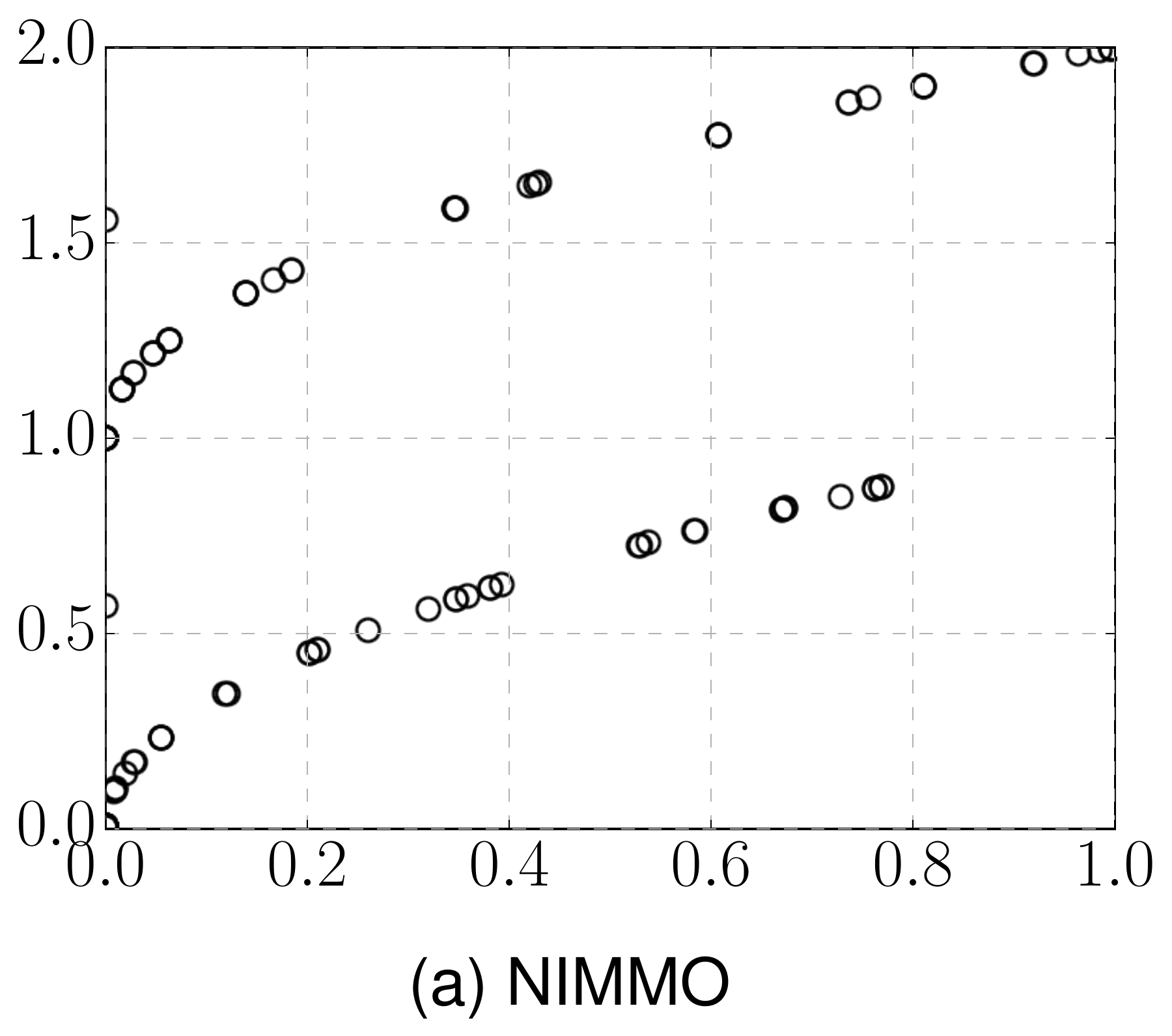}
\includegraphics[width=\widthvar\textwidth]{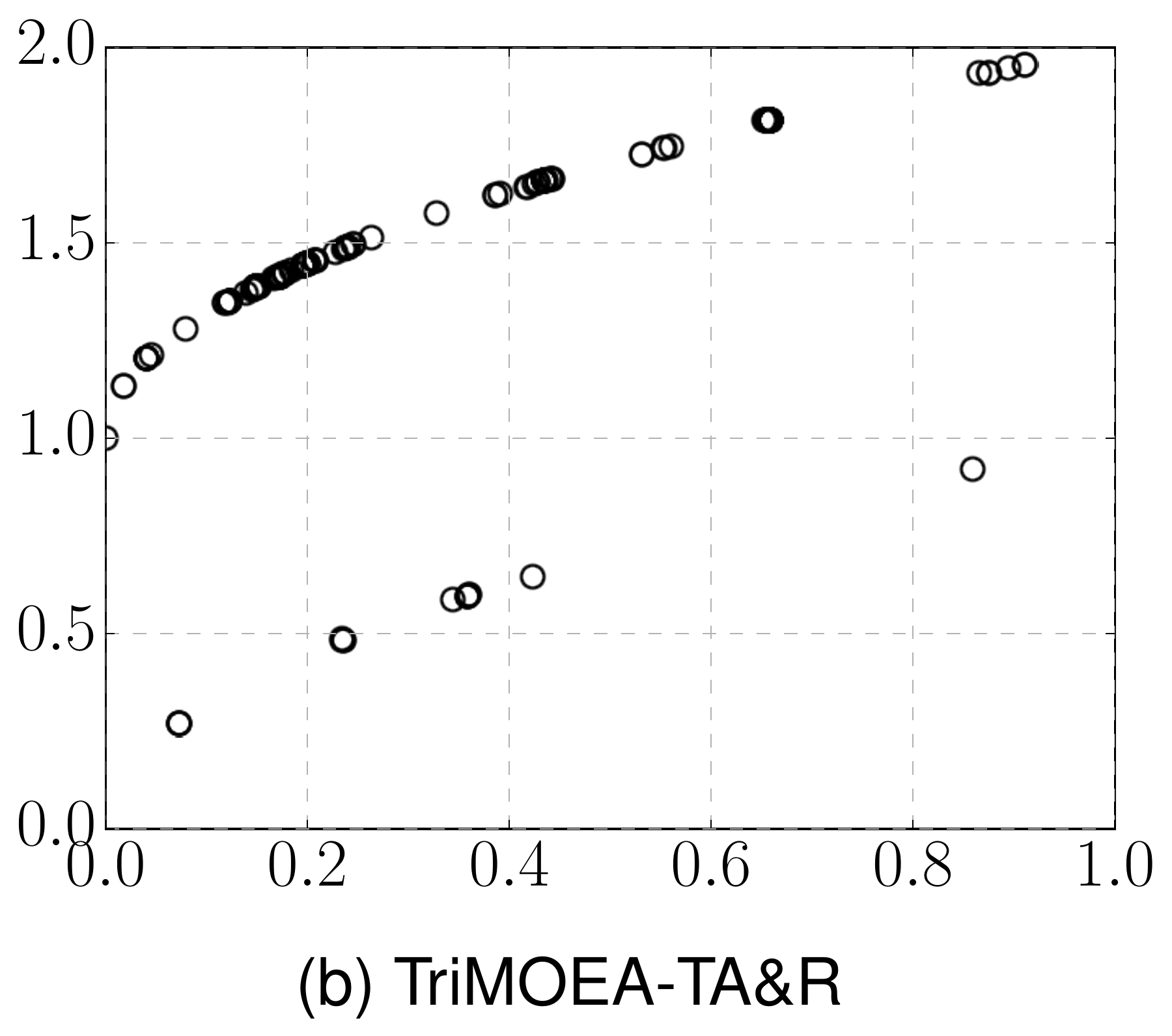}
\includegraphics[width=\widthvar\textwidth]{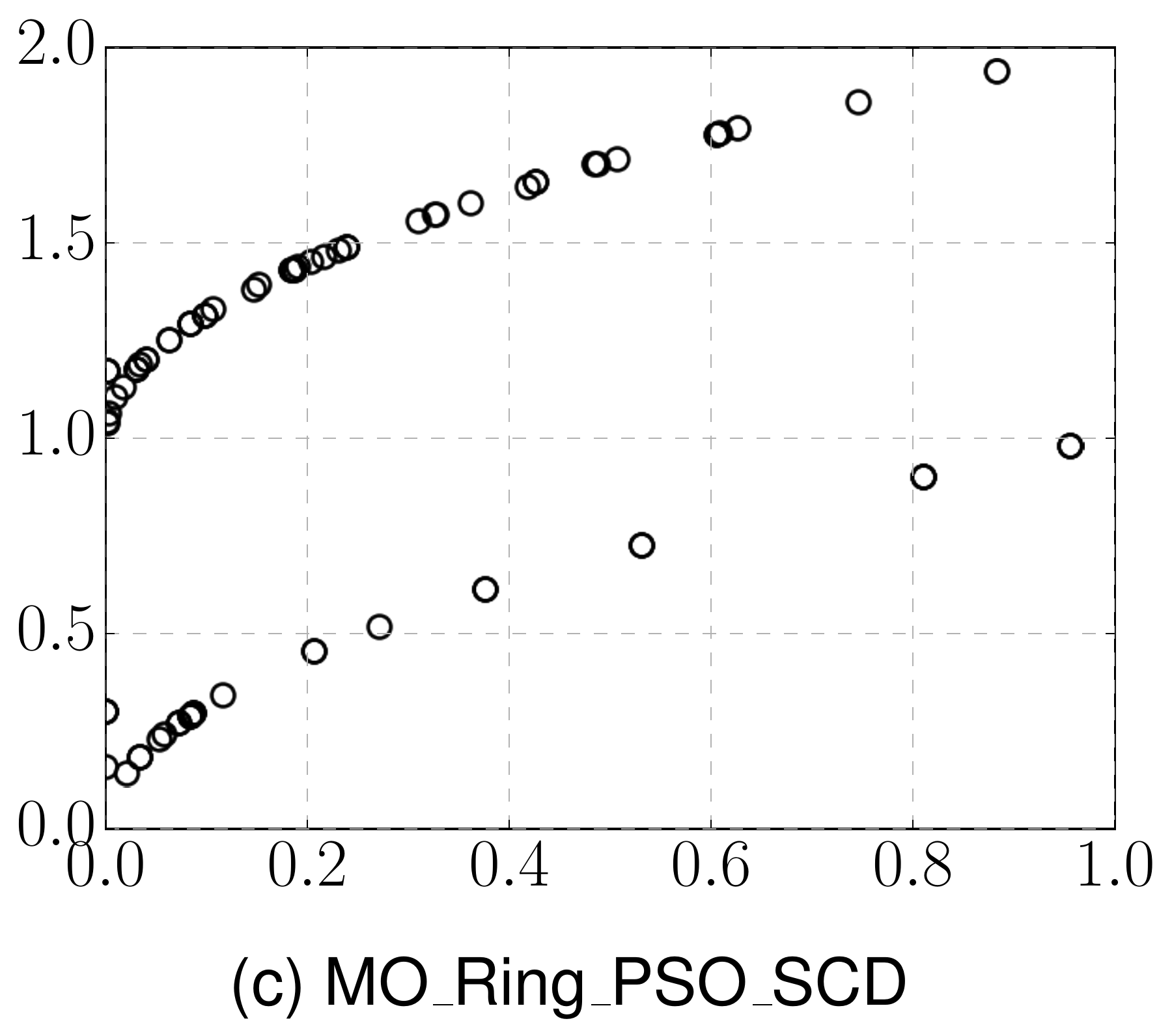}\\
\includegraphics[width=\widthvar\textwidth]{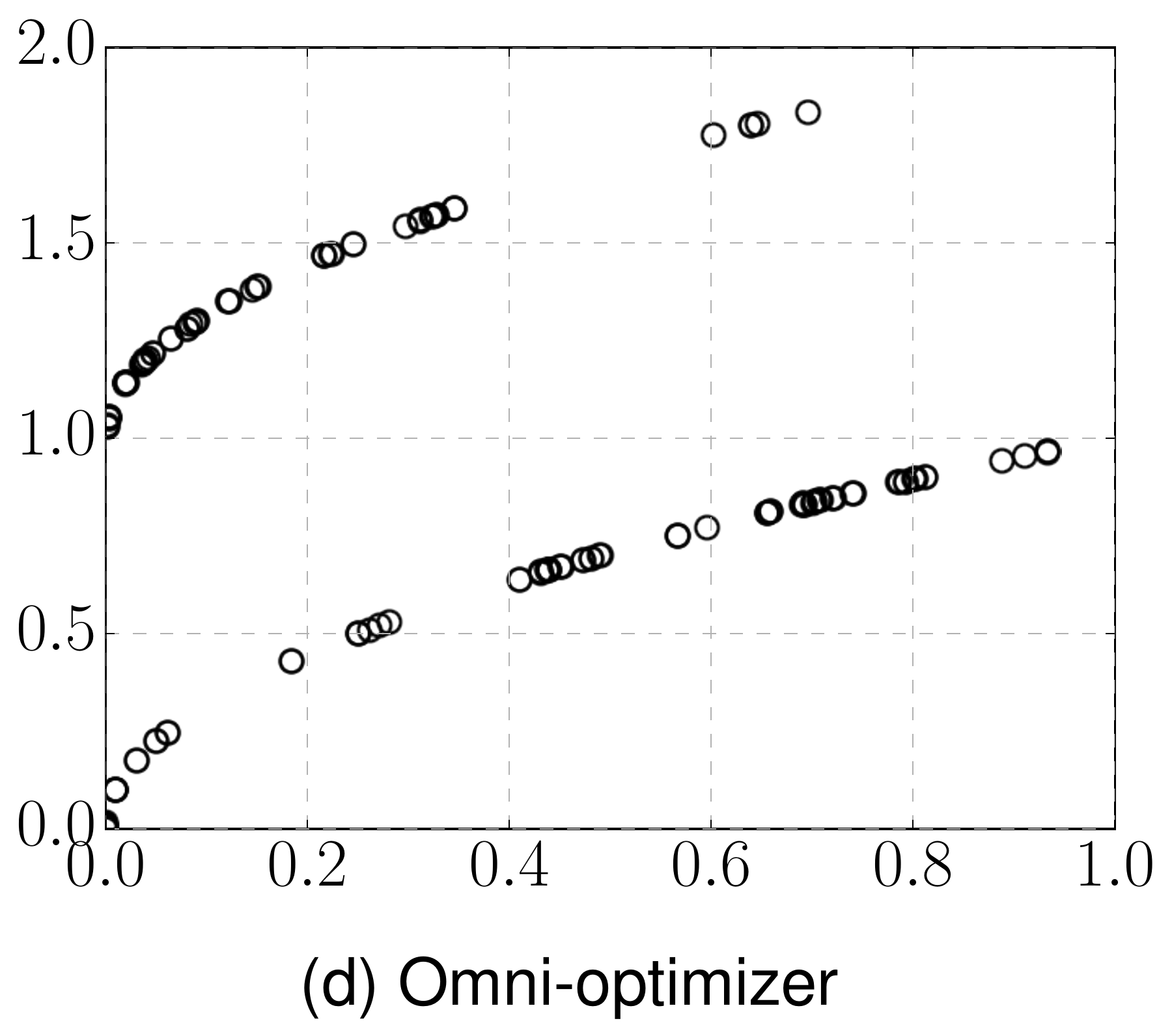}
\includegraphics[width=\widthvar\textwidth]{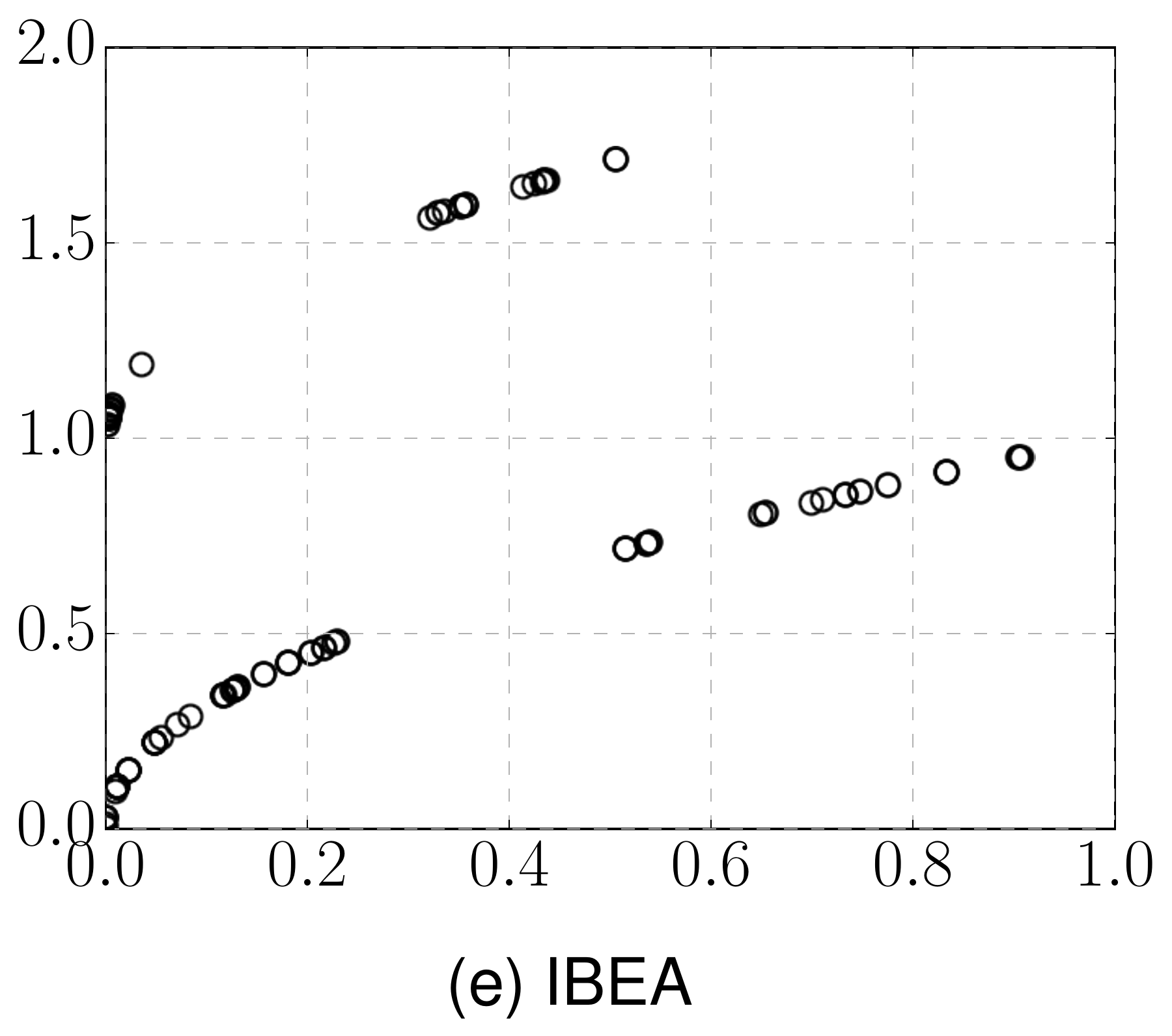}
\includegraphics[width=\widthvar\textwidth]{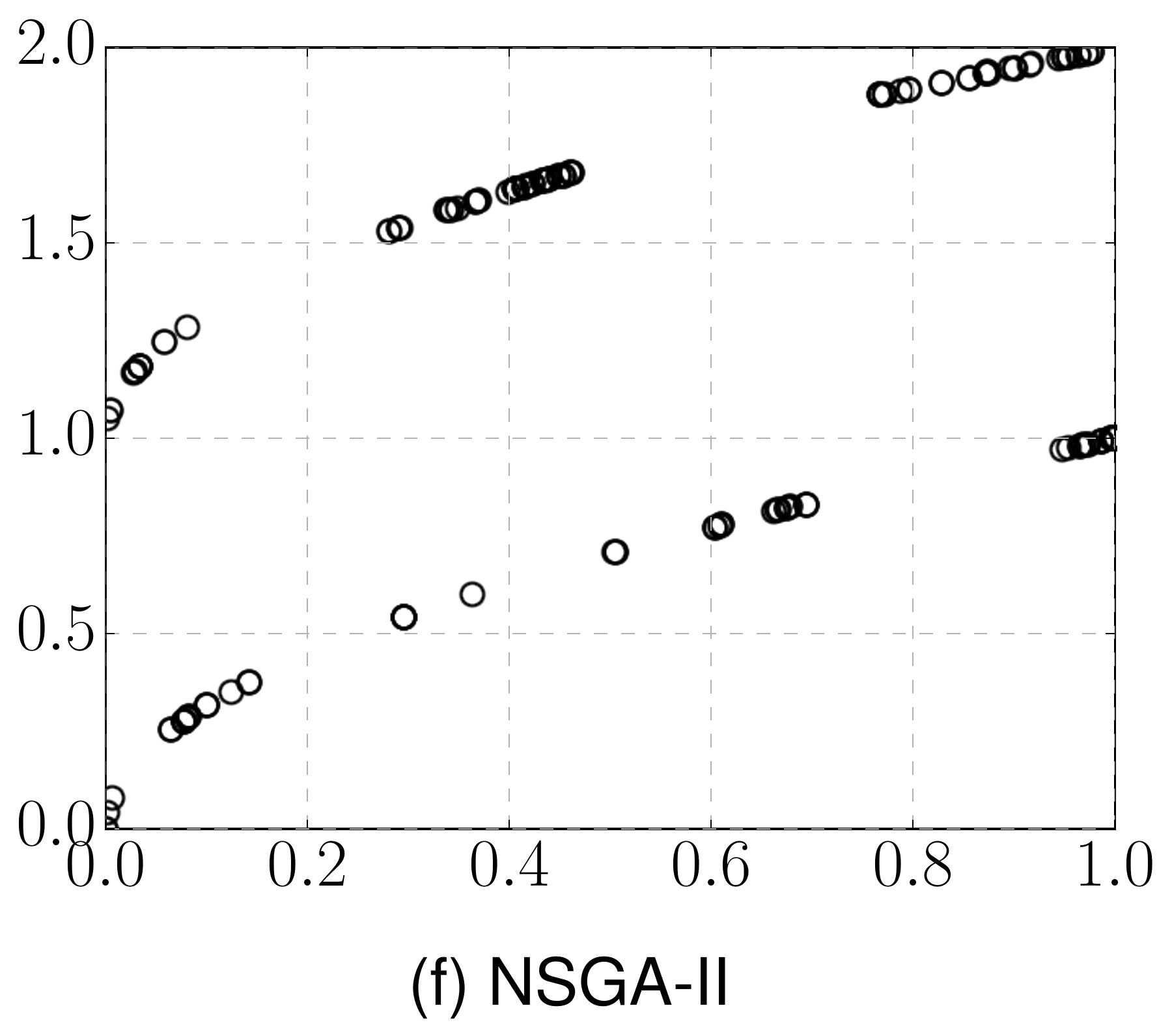}\\
\includegraphics[width=\widthvar\textwidth]{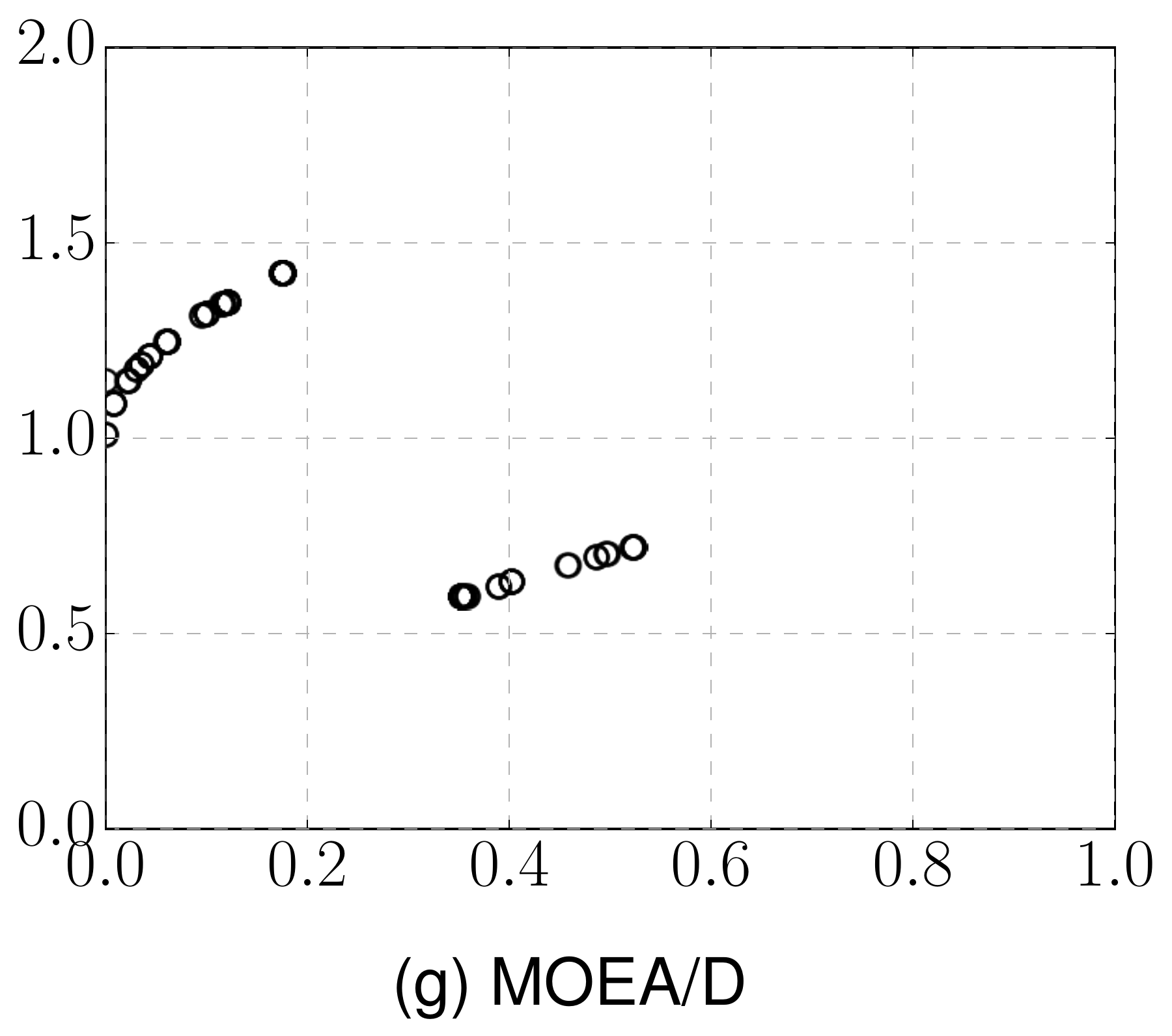}
\includegraphics[width=\widthvar\textwidth]{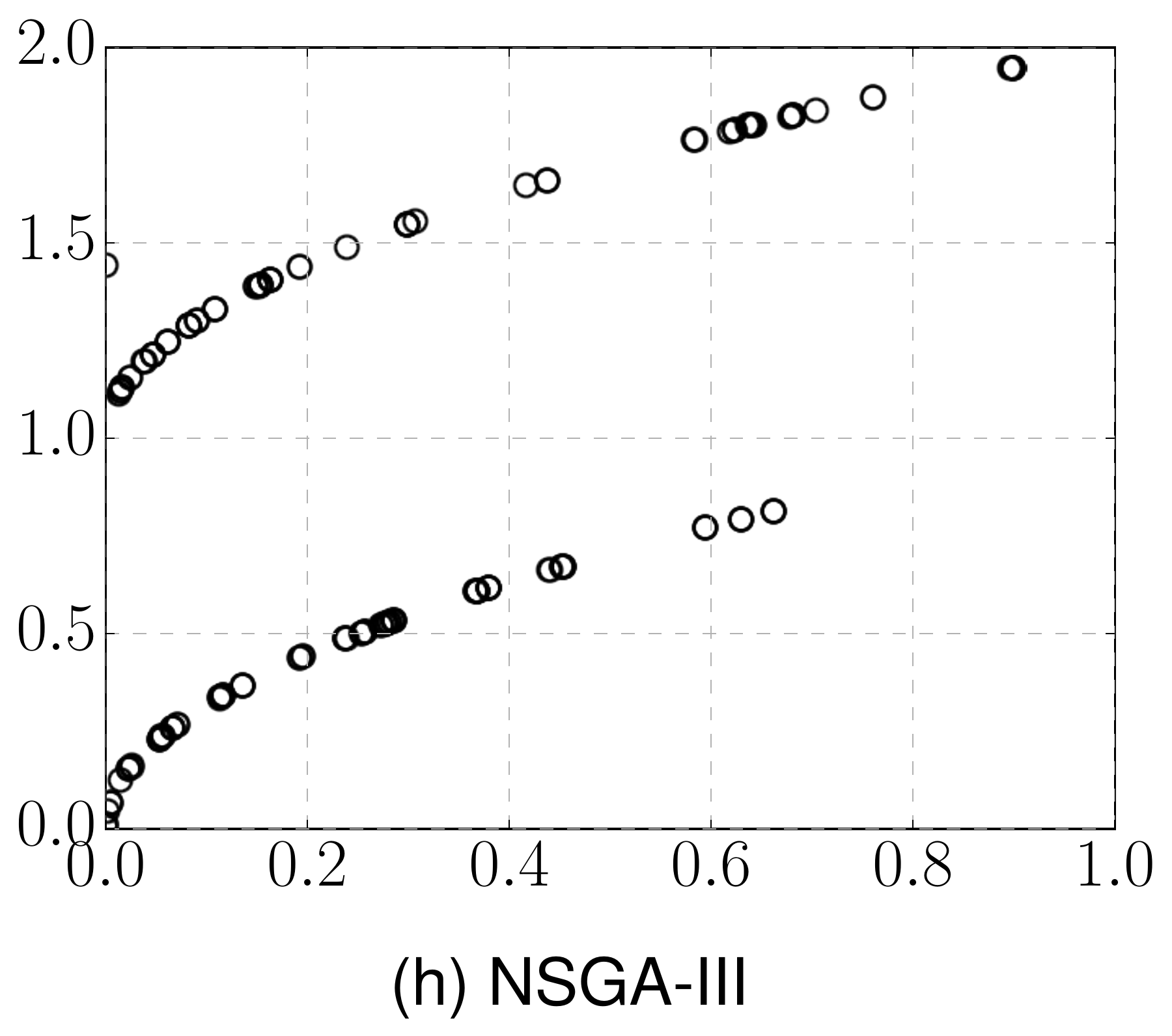}
\includegraphics[width=\widthvar\textwidth]{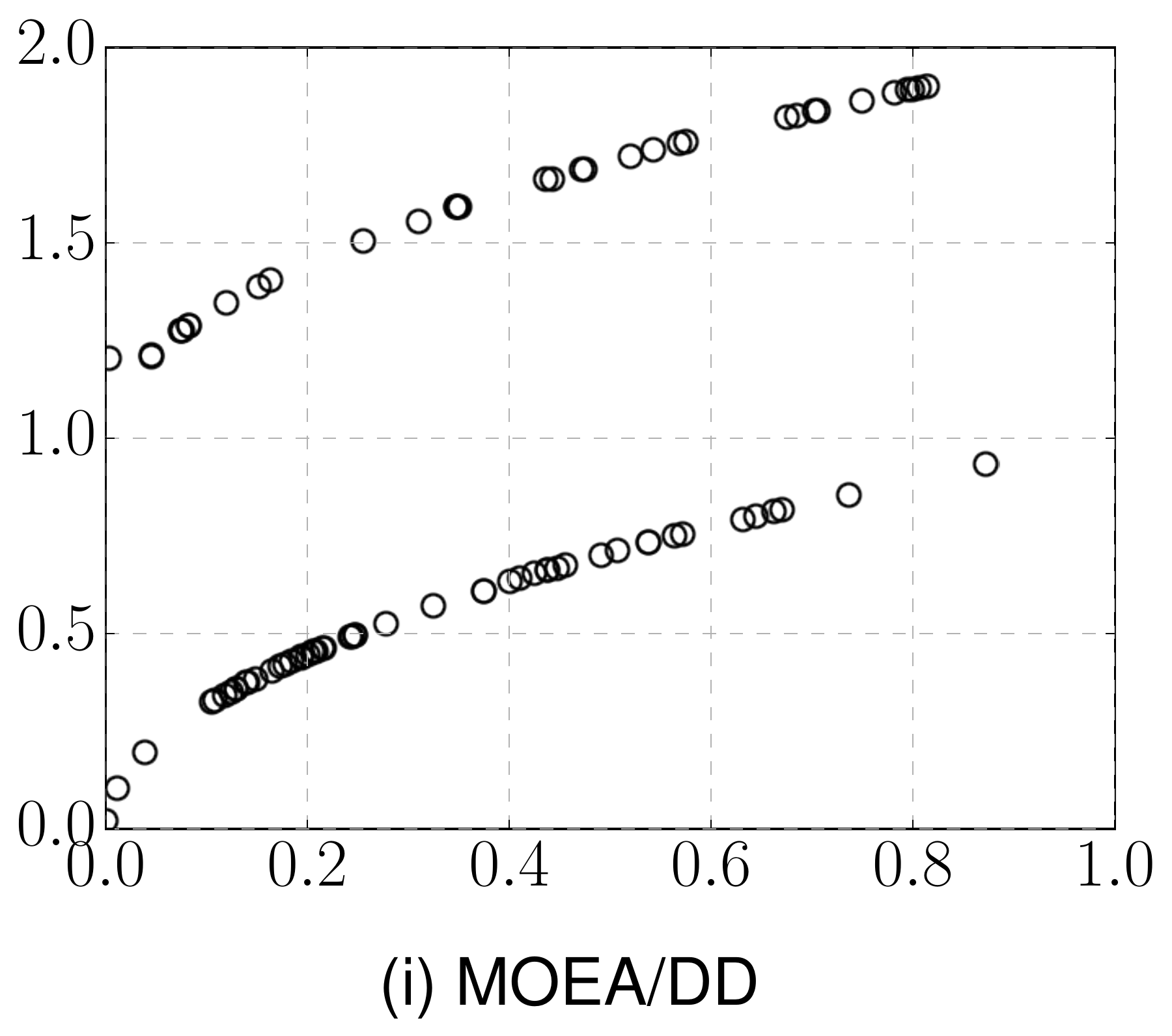}
\caption{
\small
Distribution of non-dominated solutions in the final population of each algorithm in the solution space on MMF2.
The horizontal and vertical axis represent $x_1$ and $x_2$, respectively.
}
\label{fig:mmf2-var}
   \end{center}
%
%
  \begin{center} 
    \includegraphics[width=\widthvar\textwidth]{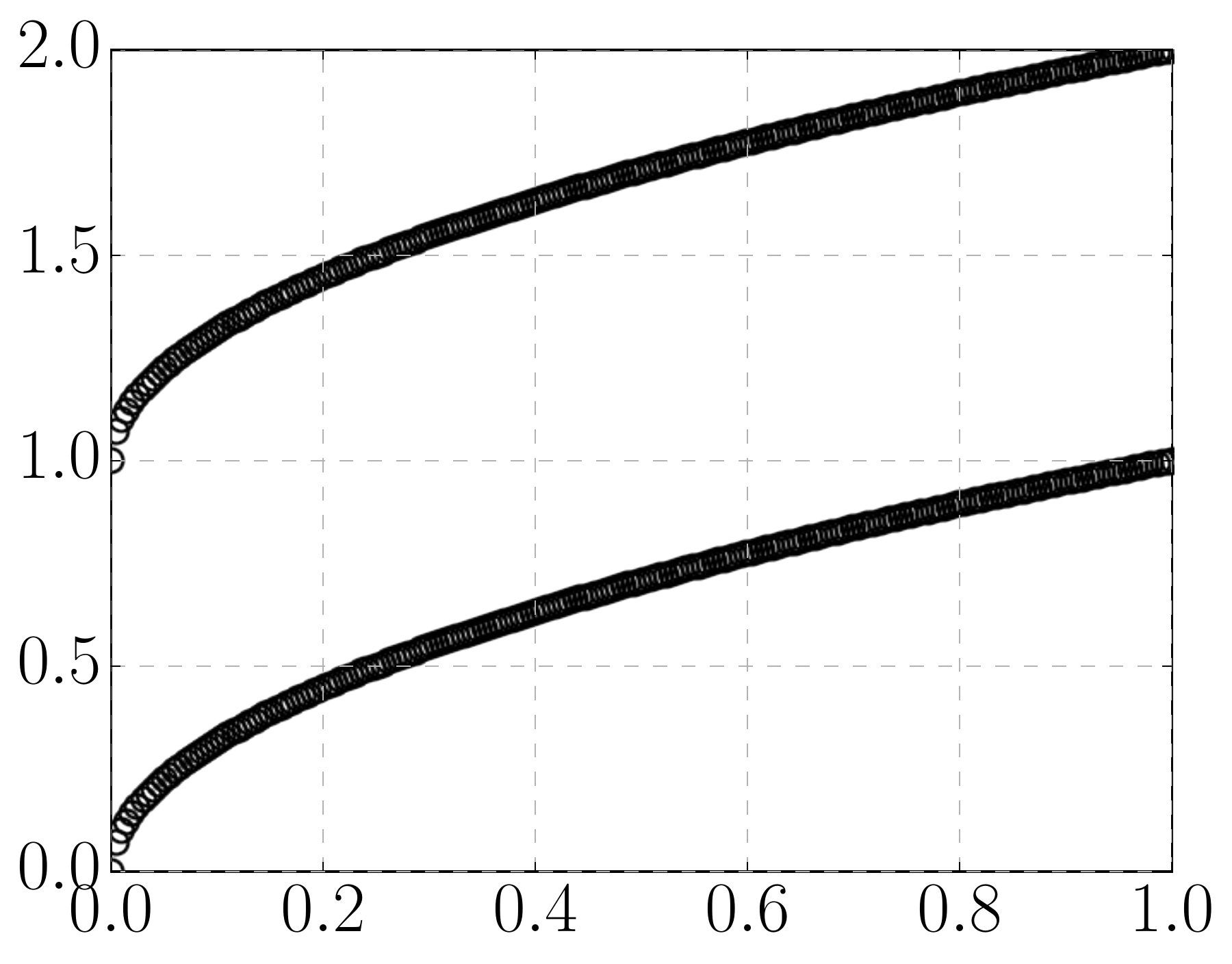}
    \caption{
      \small
      Distribution of reference solutions in the solution space on MMF2.
    }
    \label{fig:ps_mmf2}
  \end{center}
\end{figure}

\subsection{Performance of NIMMO}
\label{sec:vs_existing_methods}

Tables \ref{tab:comparison_8methods_igd}, \ref{tab:comparison_8methods_igdx}, and \ref{tab:comparison_8methods_psp} show the comparison between the nine algorithms. 
Tables \ref{tab:comparison_8methods_igd}, \ref{tab:comparison_8methods_igdx}, and \ref{tab:comparison_8methods_psp} show the IGD, IGDX, and PSP values, respectively.
In Tables \ref{tab:comparison_8methods_igd}, \ref{tab:comparison_8methods_igdx}, and \ref{tab:comparison_8methods_psp}, $M$-(R)Polygon denotes the $M$-objective (R)Polygon problem. 
The symbols $+$, $-$, and $\approx$ indicate that a given algorithm performs significantly better ($+$), significantly worse ($-$), and not significantly different better or worse ($\approx$) compared to NIMMO according to the Wilcoxon rank-sum test with $p < 0.05$.



\noindent {\bf $\bullet$ Results on the two-objective MMOPs}:
Table \ref{tab:comparison_8methods_igd} shows that NSGA-II achieves good IGD values on the two-objective MMOPs.
In particular, NSGA-II performs the best on six problems regarding the IGD metric.
MOEA/DD has the best IGD value on six MMF problems.

Table \ref{tab:comparison_8methods_igdx} shows that NIMMO obtains the best IGDX values on the seven two-objective MMOPs.
According to the IGDX metric, IBEA is not capable of finding multiple equivalent Pareto optimal solutions.
Therefore, the niching mechanism of NIMMO mainly contributes to its good ability to locate multiple Pareto optimal solutions with equivalent quality.
Table \ref{tab:comparison_8methods_psp} shows that NIMMO performs the best regarding PSP on the seven two-objective MMOPs.
The results of PSP are consistent with those of IGDX in almost all cases.

Although MOEA/DD does not have any explicit diversity maintenance mechanism in the solution space, MOEA/DD performs the best on MMF2 regarding IGDX and PSP.
Fig. \ref{fig:mmf2-var} shows the distribution of non-dominated solutions in the final population of each algorithm in the solution space on MMF2.
Results of a single run with a median IGDX value among 31 runs are shown.
Fig. \ref{fig:ps_mmf2} shows the reference solutions used for the IGDX calculation for MMF2.
As seen from Fig. \ref{fig:mmf2-var}, MOEA/DD finds well-distributed non-dominated solutions in the solution space.
The distribution of solutions found by the four MMEAs is worse than that by MOEA/DD.
The reason for these results is not obvious.
The good performance of MOEA/DD may be due to some unintended properties of MMF2.
For example, diverse solutions can be found on MMF2 without handling the solution space diversity.
Further analysis of multi-modal multi-objective test problems including MMF2 is needed as future research.

It seems that NIMMO cannot find better approximations of the PF (not PS) than NSGA-II.
It should be noted that such an observation is not unique to NIMMO and can be found in other MMEAs \cite{ShirPNE09,YueQL17}.
This is because it is difficult (or almost impossible) to obtain a set of non-dominated solutions that minimizes both the IGD and IGDX metrics as shown in Fig. \ref{fig:example_two-on-one}.
As explained in Subsection \ref{sec:performance_indicators} using the illustrative example, a good solution set regarding IGD is not always good regarding IGDX, and vice versa (for details, see Subsection \ref{sec:performance_indicators}).
Thus, there is no single solution set that simultaneously minimizes IGD and IGDX in most multi-modal multi-objective problems.
In other words, the improvements in IGD and IGDX have a trade-off relationship in most cases.

\begin{figure}[t]
  \newcommand{\widthvar}{0.4}
  \begin{center}     
    \includegraphics[width=\widthvar\textwidth]{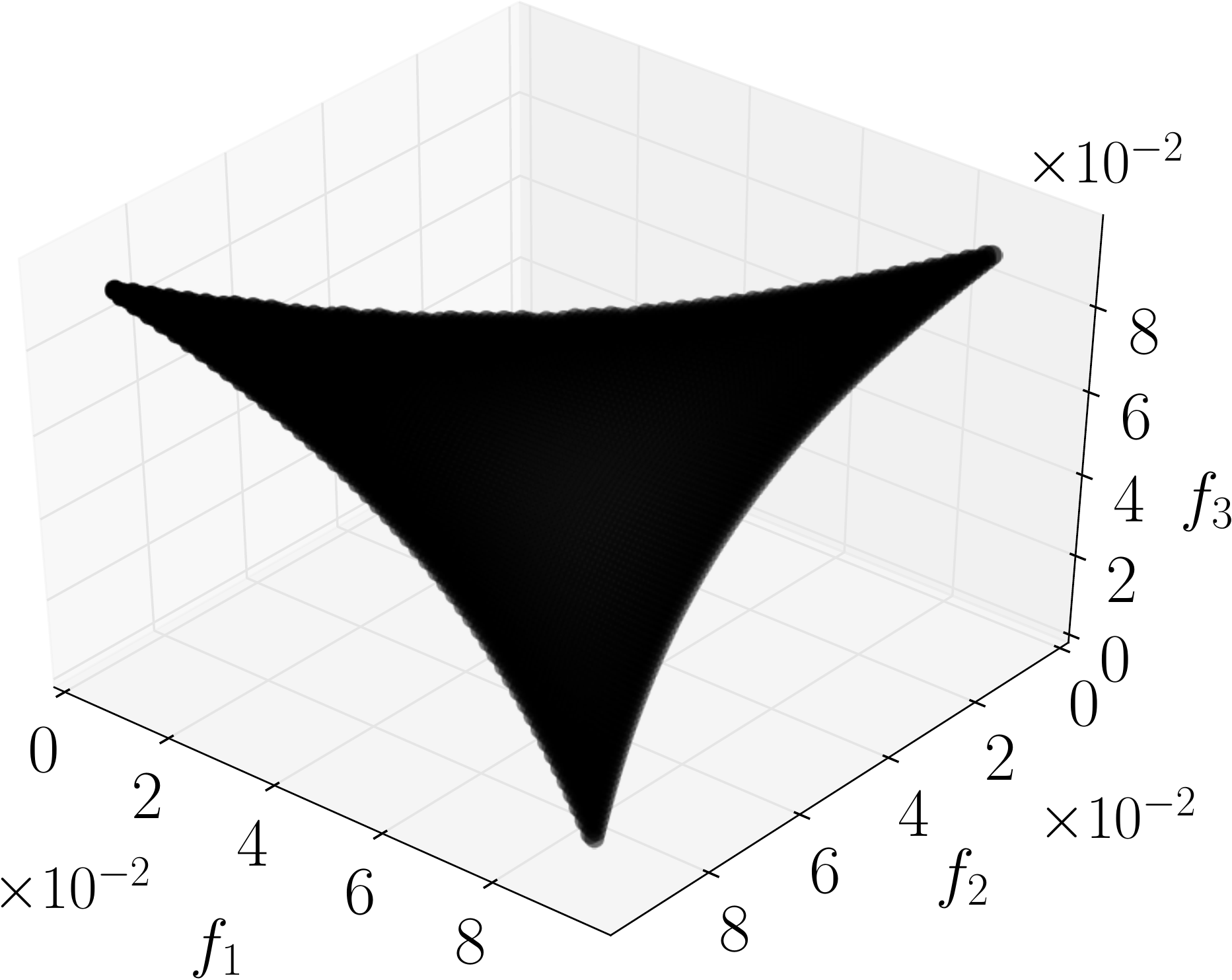}
    \caption{
      \small
      Pareto front of the three-objective Polygon problem.
    }
    \label{fig:ps_3polygon}
  \end{center}
\end{figure}



\noindent {\bf $\bullet$ Results on the MMaOPs with up to 15 objectives}:
Table \ref{tab:comparison_8methods_igd} shows that IBEA achieves the best IGD value on all the MMaOP instances, followed by NIMMO.
%
One may be surprised that NSGA-II outperforms NSGA-III on most MMaOP instances (in terms of the IGD metric).
MOEA/DD also shows the poor IGD values on MMaOPs.
However, such a non-intuitive result has already been reported in \cite{IshibuchiISN16,TanabeIO17}.
Since NSGA-III was specially designed for many-objective optimization, NSGA-III can be outperformed by NSGA-II on two- and three- objective problems.
The poor performance of a many-objective optimizer is also reported in \cite{WagnerN13}.
The poor performance of NSGA-III (and MOEA/DD) on the Polygon problems with more than three objectives is mainly due to their Pareto front shapes.
Fig. \ref{fig:ps_3polygon} shows the Pareto front of the three-objective Polygon problem.
While most multi-objective test problems (e.g., DTLZ \cite{DebTLZ05} and WFG \cite{HubandHBW06}) have triangular Pareto fronts, Polygon has an inverted triangular Pareto front, as shown in Fig. \ref{fig:ps_3polygon}.
The analysis presented in \cite{IshibuchiSMN16} shows that some decomposition-based MOEAs (including NSGA-III and MOEA/DD) perform poorly on problems with inverted triangular Pareto fronts.
This is because the shape of the distribution of weight vectors is different from the shape of the Pareto front.
For the same reason, our results show that NSGA-III performs poorly on the Polygon problems in terms of IGD.

\begin{figure*}[t]
\newcommand{\widthvar}{0.243}
  \begin{center} 
    \includegraphics[width=\widthvar\textwidth]{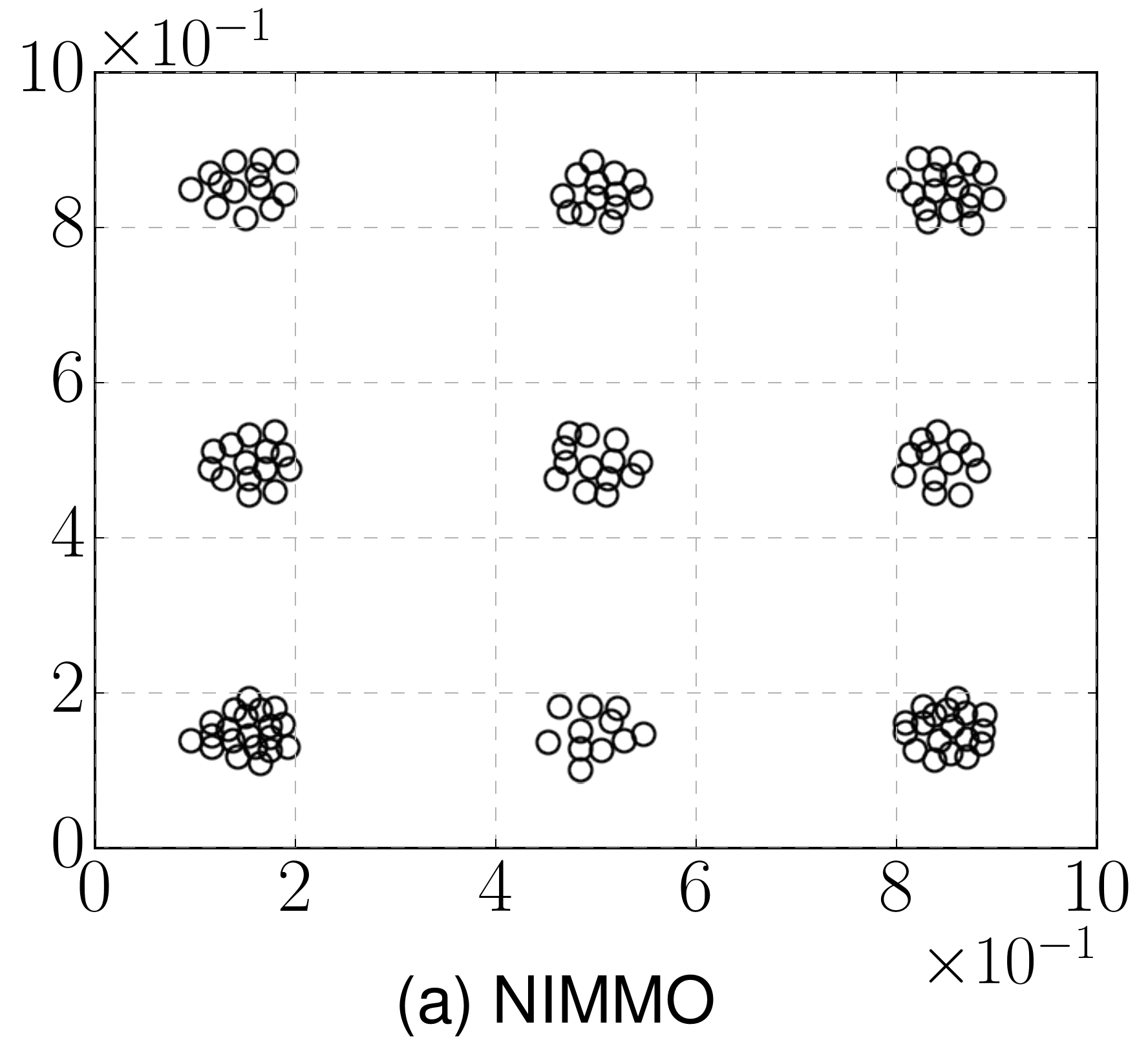}
\includegraphics[width=\widthvar\textwidth]{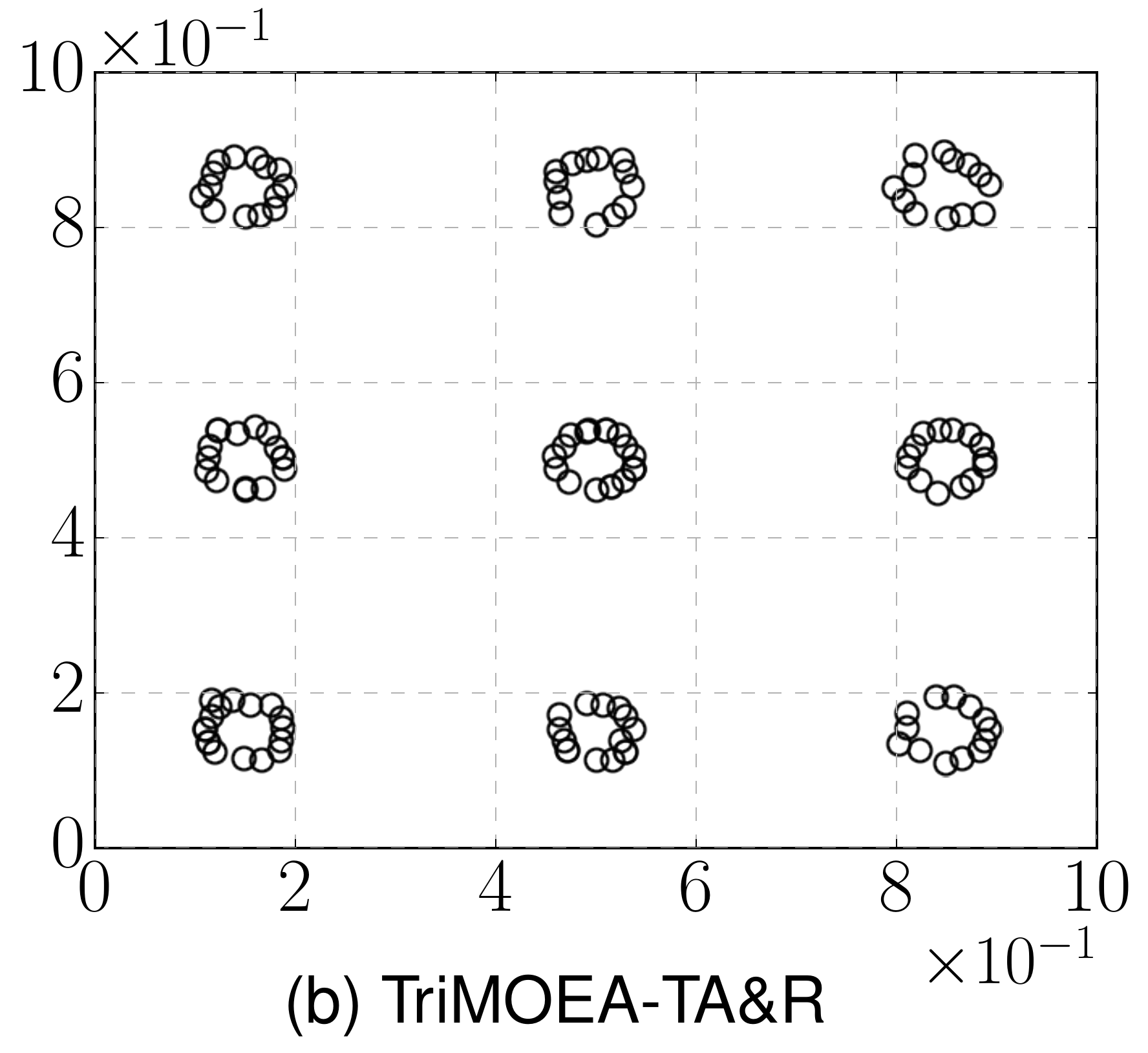}
\includegraphics[width=\widthvar\textwidth]{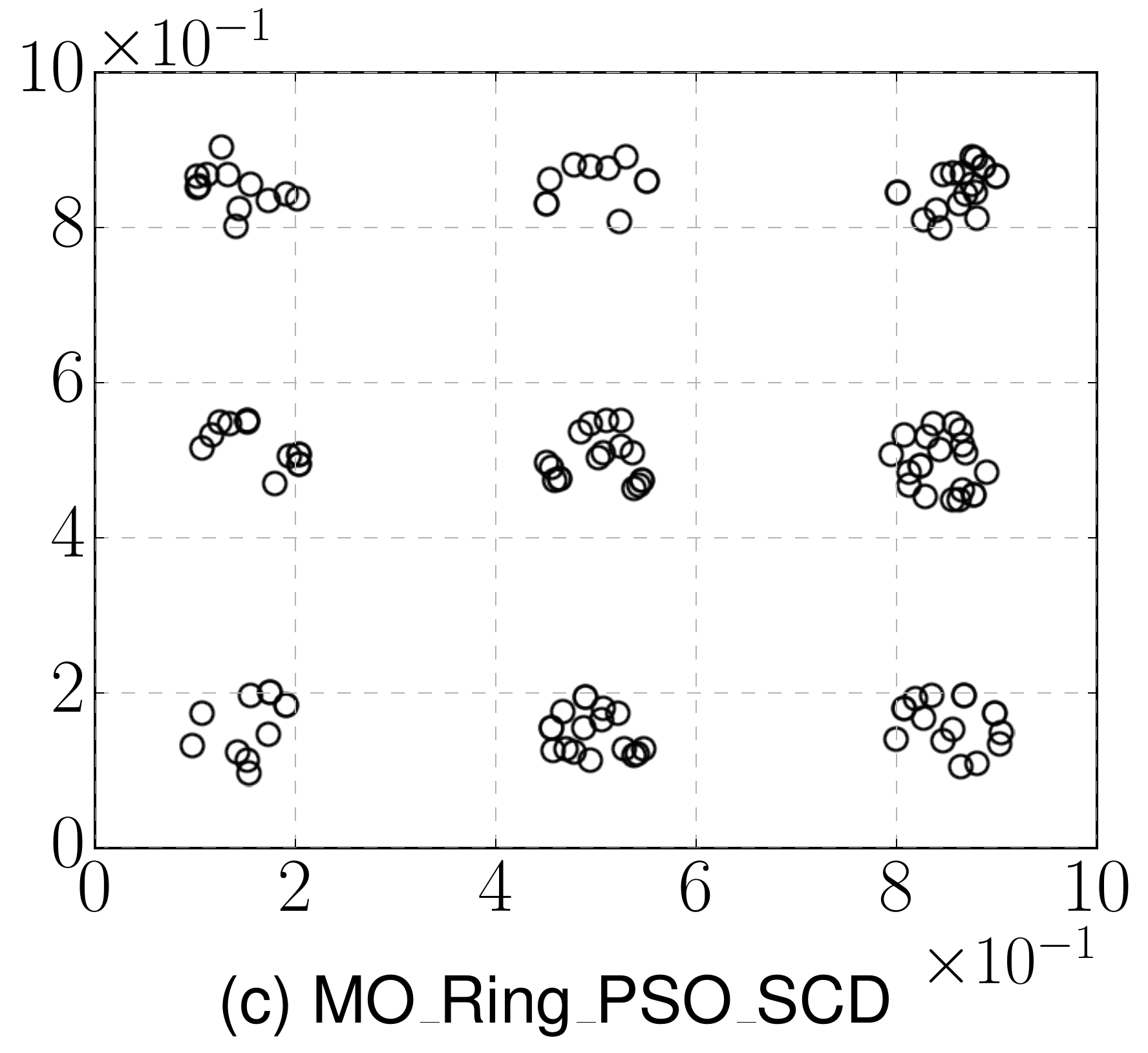}\\
\includegraphics[width=\widthvar\textwidth]{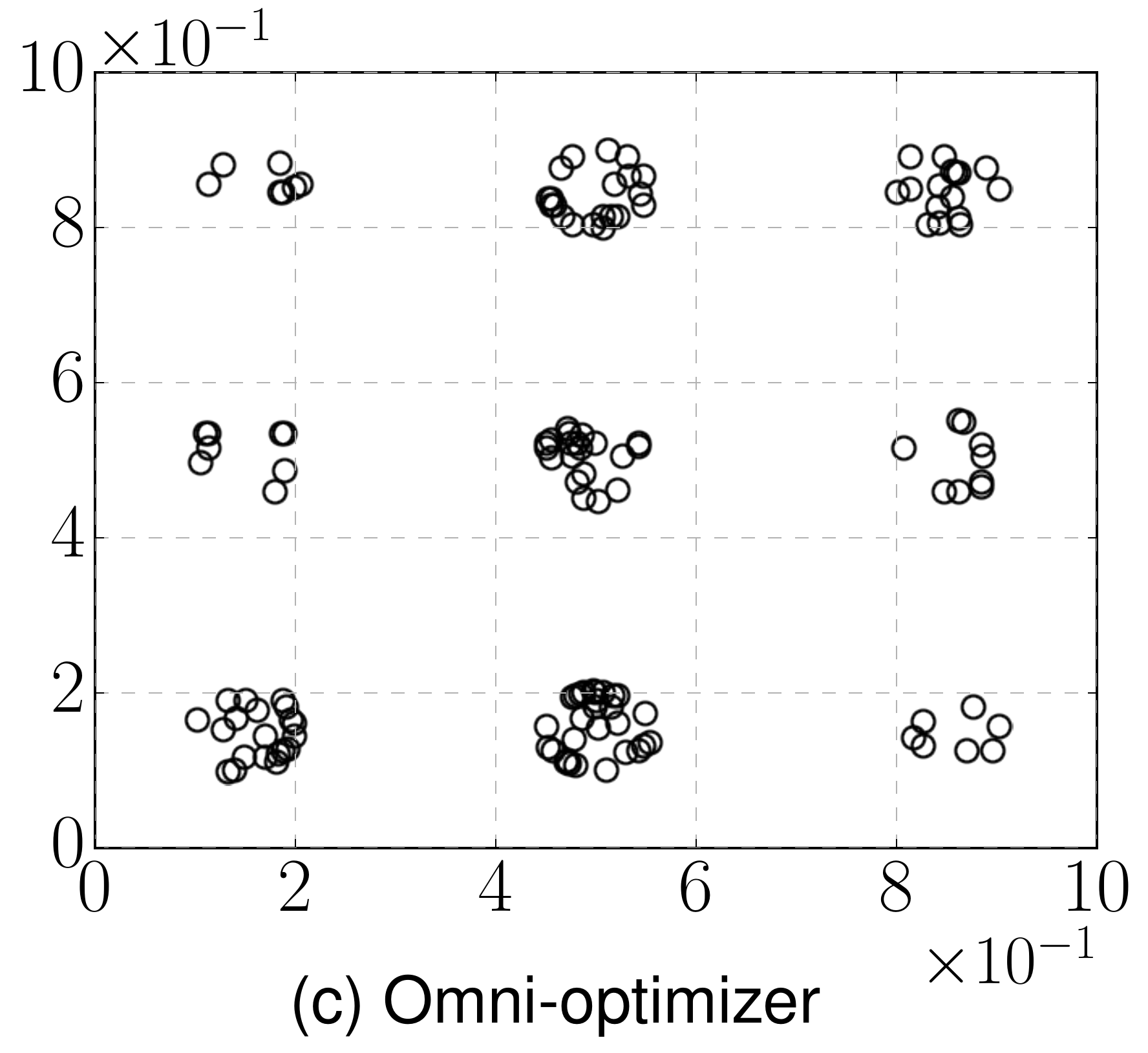}
\includegraphics[width=\widthvar\textwidth]{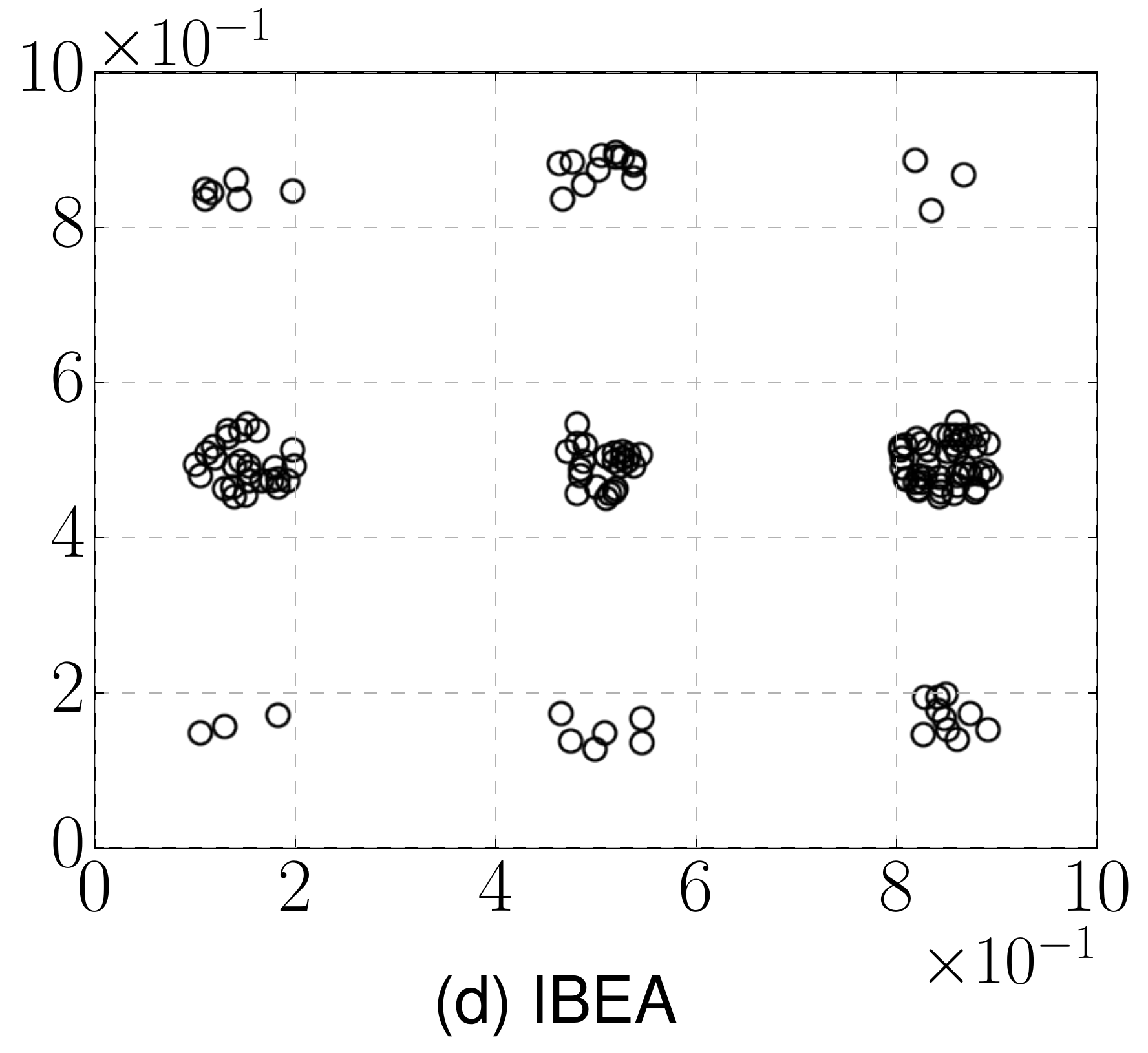}
\includegraphics[width=\widthvar\textwidth]{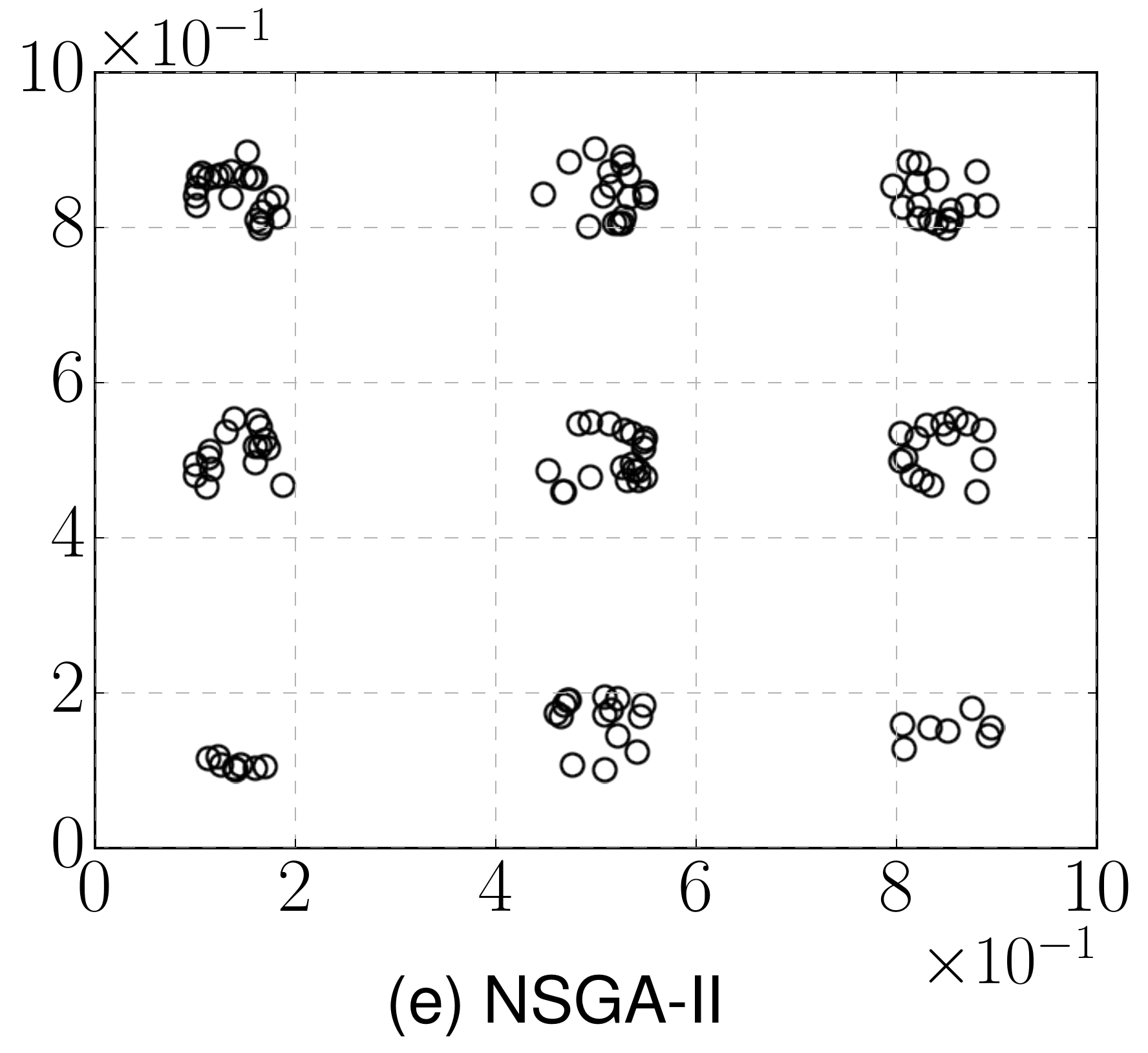}\\
\includegraphics[width=\widthvar\textwidth]{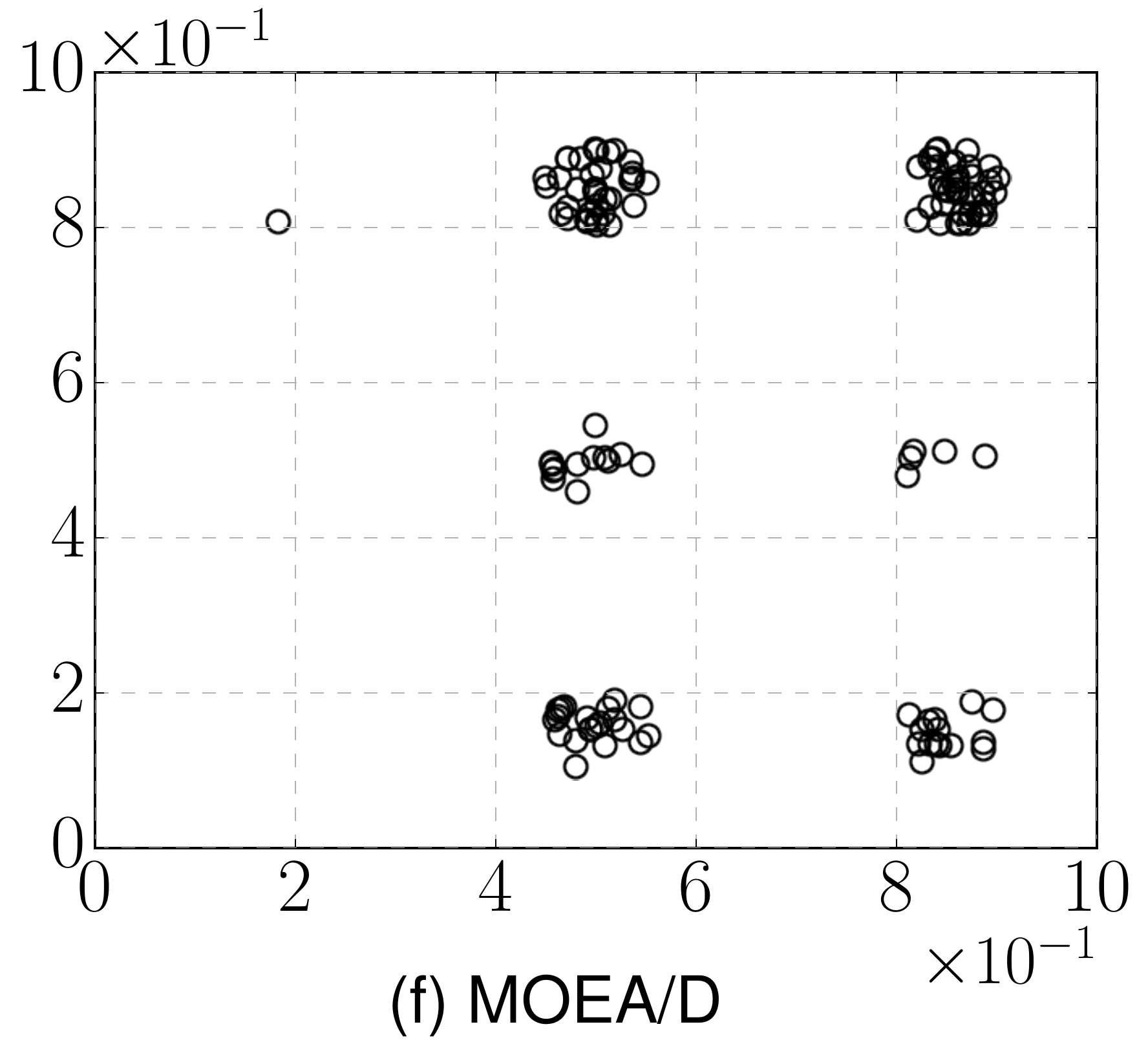}
\includegraphics[width=\widthvar\textwidth]{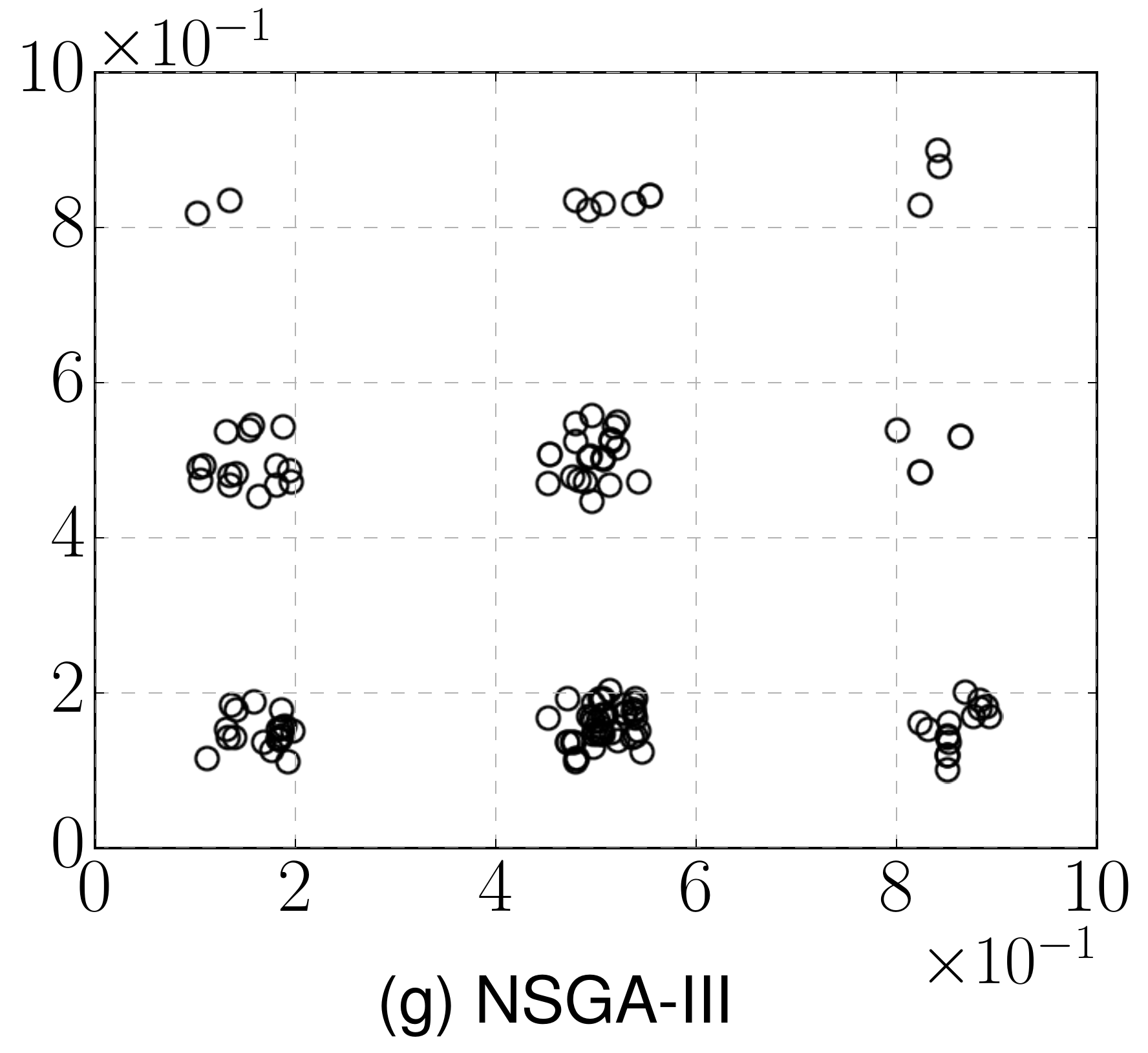}
\includegraphics[width=\widthvar\textwidth]{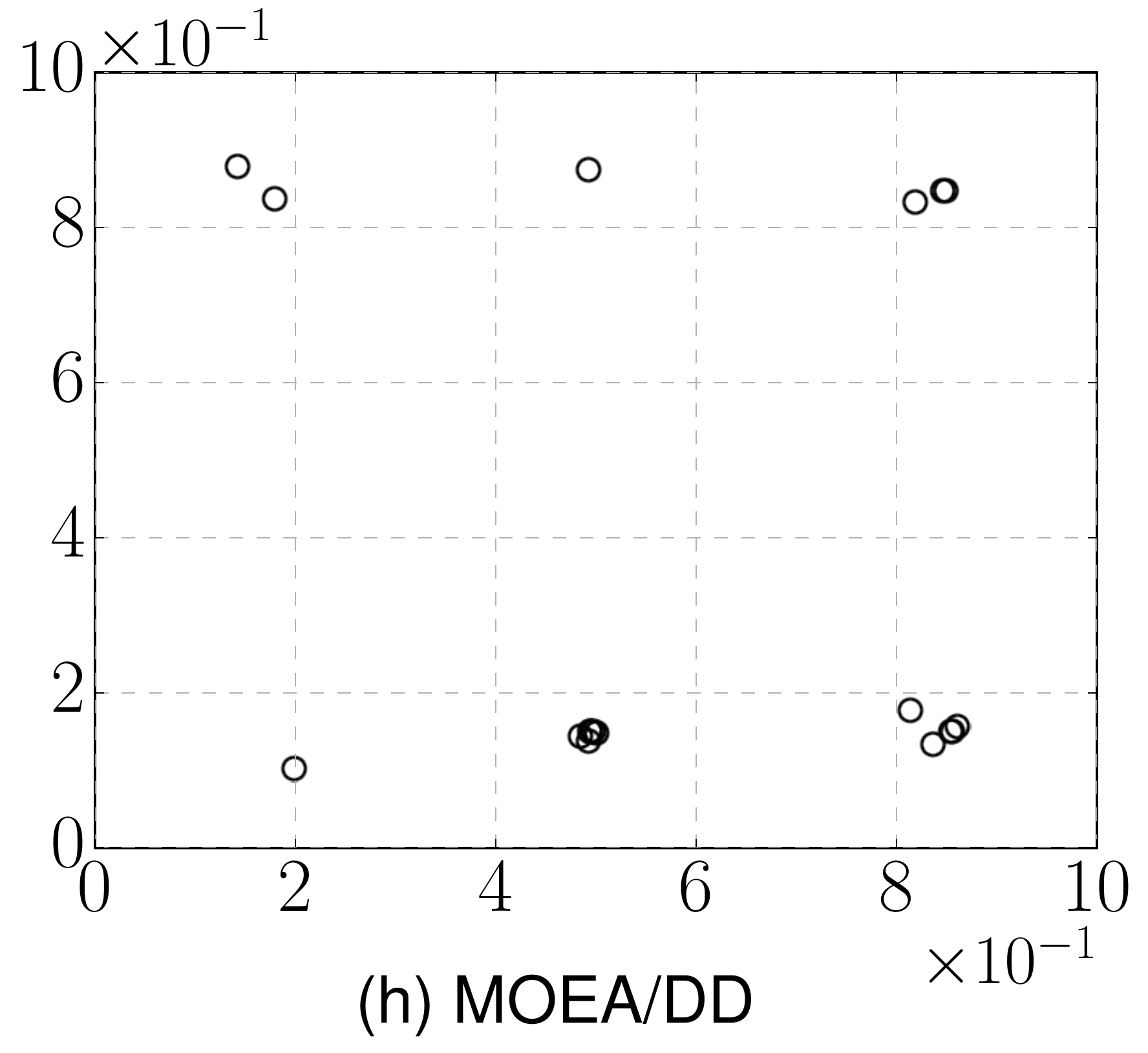}
\caption{
\small
Distribution of non-dominated solutions in the final population of each algorithm in the solution space on the $15$-Polygon problem.
The horizontal and vertical axis represent $x_1$ and $x_2$, respectively.
}
\label{fig:15-9-polygon-var}
  \end{center}
%
\end{figure*}


Tables \ref{tab:comparison_8methods_igdx} and \ref{tab:comparison_8methods_psp} indicate that NIMMO achieves the best IGDX and PSP values on all the MMaOPs, respectively.
Thus, NIMMO is capable of locating multiple equivalent Pareto optimal solutions on MMaOPs with a large number of objectives.
According to the IGDX and PSP indicators, Omni-optimizer performs the second best as a multi-modal multi-objective optimization algorithm. 



Fig. \ref{fig:15-9-polygon-var} shows the distribution of non-dominated solutions in the final population of each algorithm in the solution space on the $15$-Polygon problems, whose Pareto optimal solutions are on the inside of the nine regular $15$-sided polygons.
Results of a single run with a median IGDX value among 31 runs are shown.
Fig. \ref{fig:15-9-polygon-var} shows all algorithms (except for MOEA/D and MOEA/DD) locate the nine Pareto optimal solution subsets.
While the solutions by NIMMO are almost evenly distributed, those by the other algorithms are biased.
Fig. \ref{fig:15-9-polygon-var} (b) shows that the solutions in the decomposition-based TriMOEA-TA\&R are distributed on the edges of the polygons.
As explained above, this result is due to the undesirable property of decomposition-based algorithms on problems with irregular Pareto fronts.

\noindent {\bf $\bullet$ Summary}:
Table \ref{tab:Friedman_ranking_ninealgs_tradtional} summarizes the average rankings of the nine algorithms on the 23 problem instances by the Friedman test \cite{GarciaFLH10,DerracGMH11}.
We used the CONTROLTEST package (\url{https://sci2s.ugr.es/sicidm}) to calculate the rankings.
Table \ref{tab:Friedman_ranking_ninealgs_tradtional} (a) shows that NIMMO is outperformed by Omni-optimizer, IBEA, and NSGA-II regarding IGD.
Tables \ref{tab:Friedman_ranking_ninealgs_tradtional} (b) and (c) show that NIMMO performs the best among the nine algorithms regarding IGDX and PSP.
Although TriMOEA-TA\&R is an MMEA, it is outperformed by some MOEAs (e.g., NSGA-II and NSGA-III regarding PSP).
As explained in Subsection \ref{sec:emoas_for_mmops}, TriMOEA-TA\&R explicitly exploits the properties of position-related and distance-related variables.
However, the 23 problem instances used in our benchmarking study do not have position-related and distance-related variables.
It is also questionable whether a real-world problem has such a special design variable.
In summary, NIMMO is the best multi-modal multi- and many-objective optimizer among the competitors.



\begin{table}[t]
\centering
\caption{\small Average rankings of the nine algorithms by the Friedman test. ``MO\_Ring'' stands for ``MO\_Ring\_PSO\_SCD''.
}
\label{tab:Friedman_ranking_ninealgs_tradtional}
\centering
{\scriptsize
  \subfloat[IGD]{    
\begin{tabular}{|c|c|}
\hline
  Algorithm&Rank\\
  \hline
NIMMO&4.22\\
TriMOEA-TA\&R&8.22\\
MO\_Ring&6.87\\
Omni-optimizer&3.70\\
IBEA&3.17\\
NSGA-II&2.83\\
MOEA/D&5.91\\
NSGA-III&4.30\\
MOEA/DD&5.78\\
\hline
\end{tabular}
}
  \subfloat[IGDX]{    
\begin{tabular}{|c|c|}
\hline
  Algorithm&Rank\\
\hline
NIMMO&1.52\\
TriMOEA-TA\&R&6.43\\
MO\_Ring&2.96\\
Omni-optimizer&2.61\\
IBEA&6.17\\
NSGA-II&5.00\\
MOEA/D&8.65\\
NSGA-III&5.70\\
MOEA/DD&5.96\\
\hline
\end{tabular}
}
  \subfloat[PSP]{    
\begin{tabular}{|c|c|}
\hline
  Algorithm&Rank\\
\hline
NIMMO&1.72\\
TriMOEA-TA\&R&6.04\\
MO\_Ring&3.16\\
Omni-optimizer&2.64\\
IBEA&6.32\\
NSGA-II&4.76\\
MOEA/D&8.56\\
NSGA-III&5.72\\
MOEA/DD&6.08\\
\hline
\end{tabular}
  }
  }
\end{table}

\begin{table}[t]
\centering
\caption{\small Average rankings of NIMMO with the 11 $T$ values by the Friedman test.
}
\label{tab:Friedman_ranking_variousT_tradtional}
\centering
{\scriptsize
  \subfloat[IGD]{    
\begin{tabular}{|c|c|}
\hline
  Algorithm&Rank\\
  \hline
$\lfloor 0.05\mu \rfloor$ &8.12\\
$\lfloor 0.1\mu \rfloor$ &7.76\\
$\lfloor 0.2\mu \rfloor$ &7.88\\
$\lfloor 0.3\mu \rfloor$ &7.64\\
$\lfloor 0.4\mu \rfloor$ &7.24\\
$\lfloor 0.5\mu \rfloor$ &6.84\\
$\lfloor 0.6\mu \rfloor$ &5.80\\
$\lfloor 0.7\mu \rfloor$ &4.76\\
$\lfloor 0.8\mu \rfloor$ &4.04\\
$\lfloor 0.9\mu \rfloor$ &3.32\\
$\lfloor 1\mu \rfloor$ &2.60\\
\hline
\end{tabular}
}
  \subfloat[IGDX]{    
\begin{tabular}{|c|c|}
\hline
  Algorithm&Rank\\
\hline
$\lfloor 0.05\mu \rfloor$ &1.76\\
$\lfloor 0.1\mu \rfloor$ &1.52\\
$\lfloor 0.2\mu \rfloor$ &5.08\\
$\lfloor 0.3\mu \rfloor$ &6.80\\
$\lfloor 0.4\mu \rfloor$ &6.28\\
$\lfloor 0.5\mu \rfloor$ &7.20\\
$\lfloor 0.6\mu \rfloor$ &6.88\\
$\lfloor 0.7\mu \rfloor$ &6.72\\
$\lfloor 0.8\mu \rfloor$ &7.24\\
$\lfloor 0.9\mu \rfloor$ &8.00\\
$\lfloor 1\mu \rfloor$ &8.52\\
\hline
\end{tabular}
}
  \subfloat[PSP]{    
\begin{tabular}{|c|c|}
\hline
  Algorithm&Rank\\
  \hline
$\lfloor 0.05\mu \rfloor$ &1.80\\
$\lfloor 0.1\mu \rfloor$ &1.52\\
$\lfloor 0.2\mu \rfloor$ &4.48\\
$\lfloor 0.3\mu \rfloor$ &6.72\\
$\lfloor 0.4\mu \rfloor$ &6.68\\
$\lfloor 0.5\mu \rfloor$ &6.92\\
$\lfloor 0.6\mu \rfloor$ &7.00\\
$\lfloor 0.7\mu \rfloor$ &7.04\\
$\lfloor 0.8\mu \rfloor$ &7.52\\
$\lfloor 0.9\mu \rfloor$ &8.04\\
$\lfloor 1\mu \rfloor$ &8.28\\
\hline
\end{tabular}
  }
  }
\end{table}

\begin{figure}[t]
\newcommand{\widthvar}{0.33}
\centering
\subfloat[Two-On-One]{\includegraphics[width=0.34\textwidth]{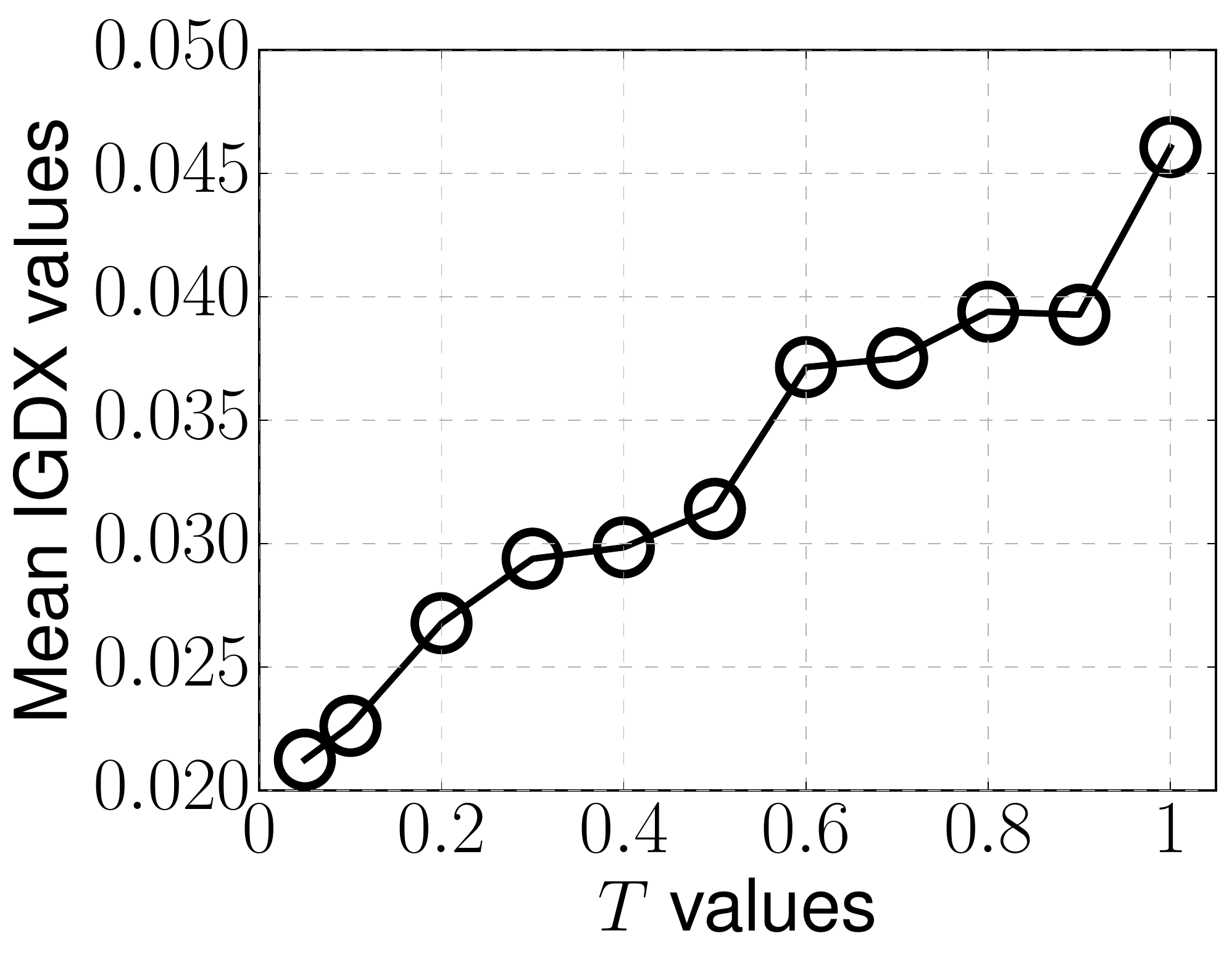}}
\subfloat[Omni-test]{\includegraphics[width=0.32\textwidth]{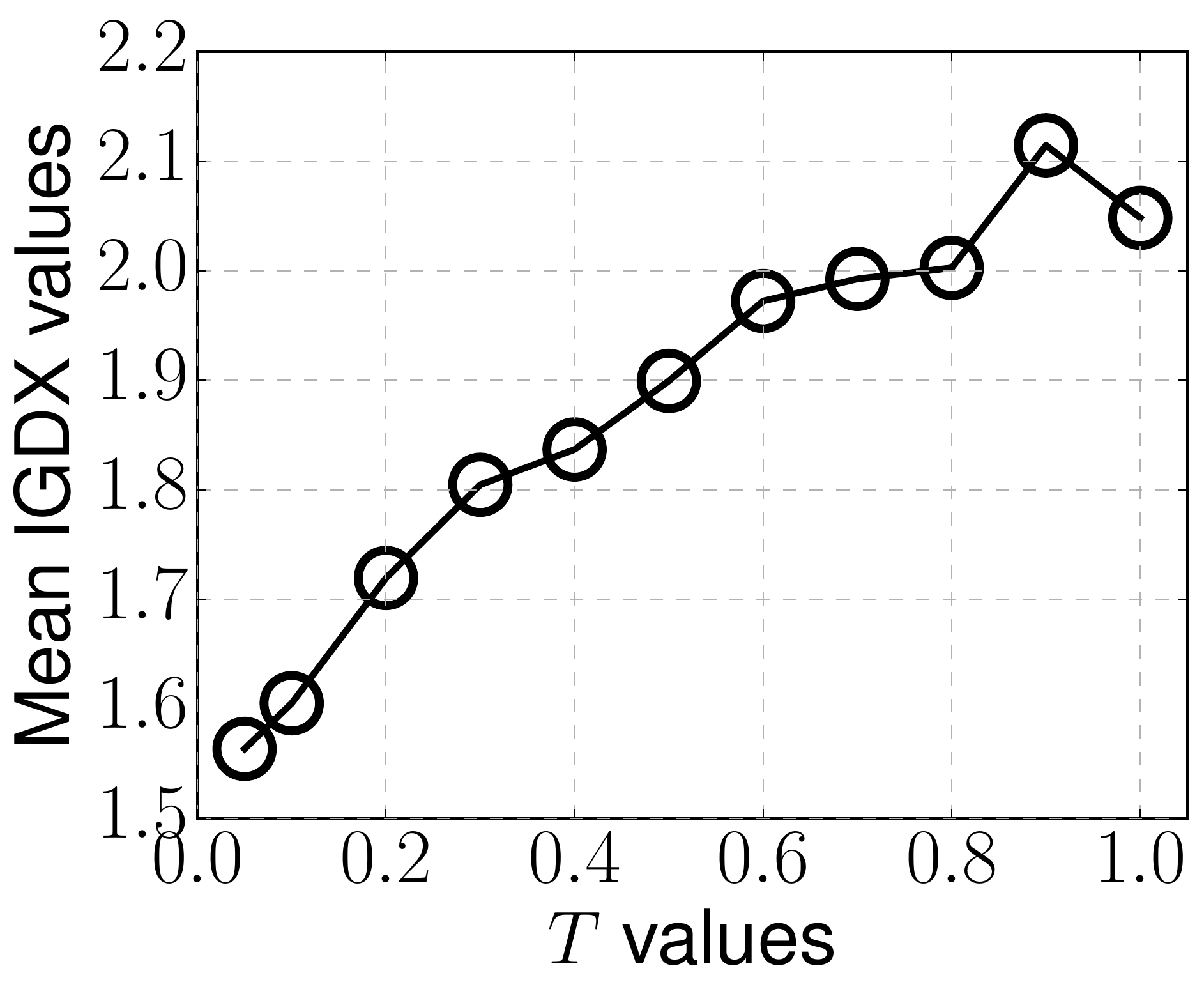}}
\subfloat[SYM-PART1]{\includegraphics[width=0.32\textwidth]{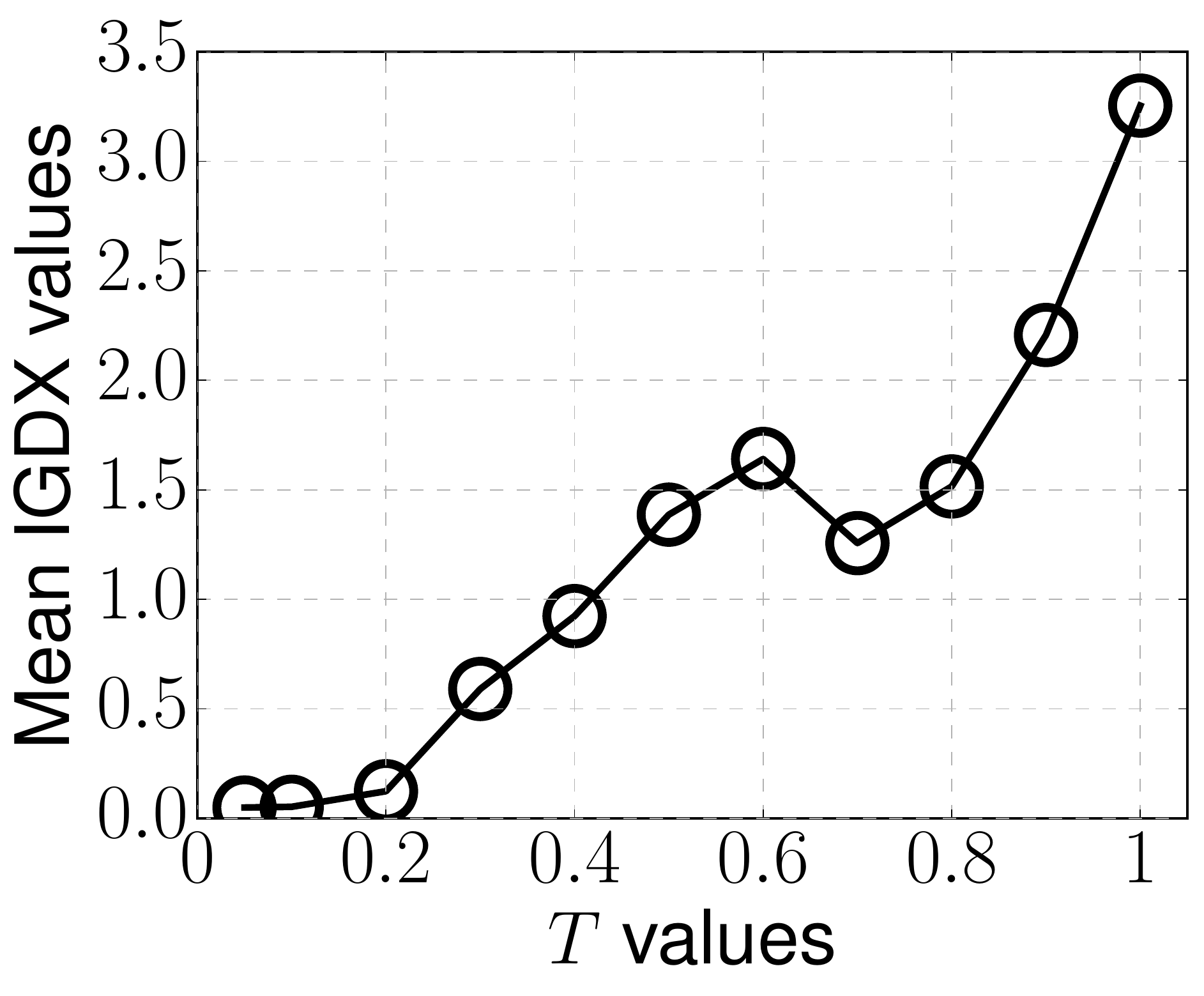}}
\\
\subfloat[$3$-Polygon]{\includegraphics[width=\widthvar\textwidth]{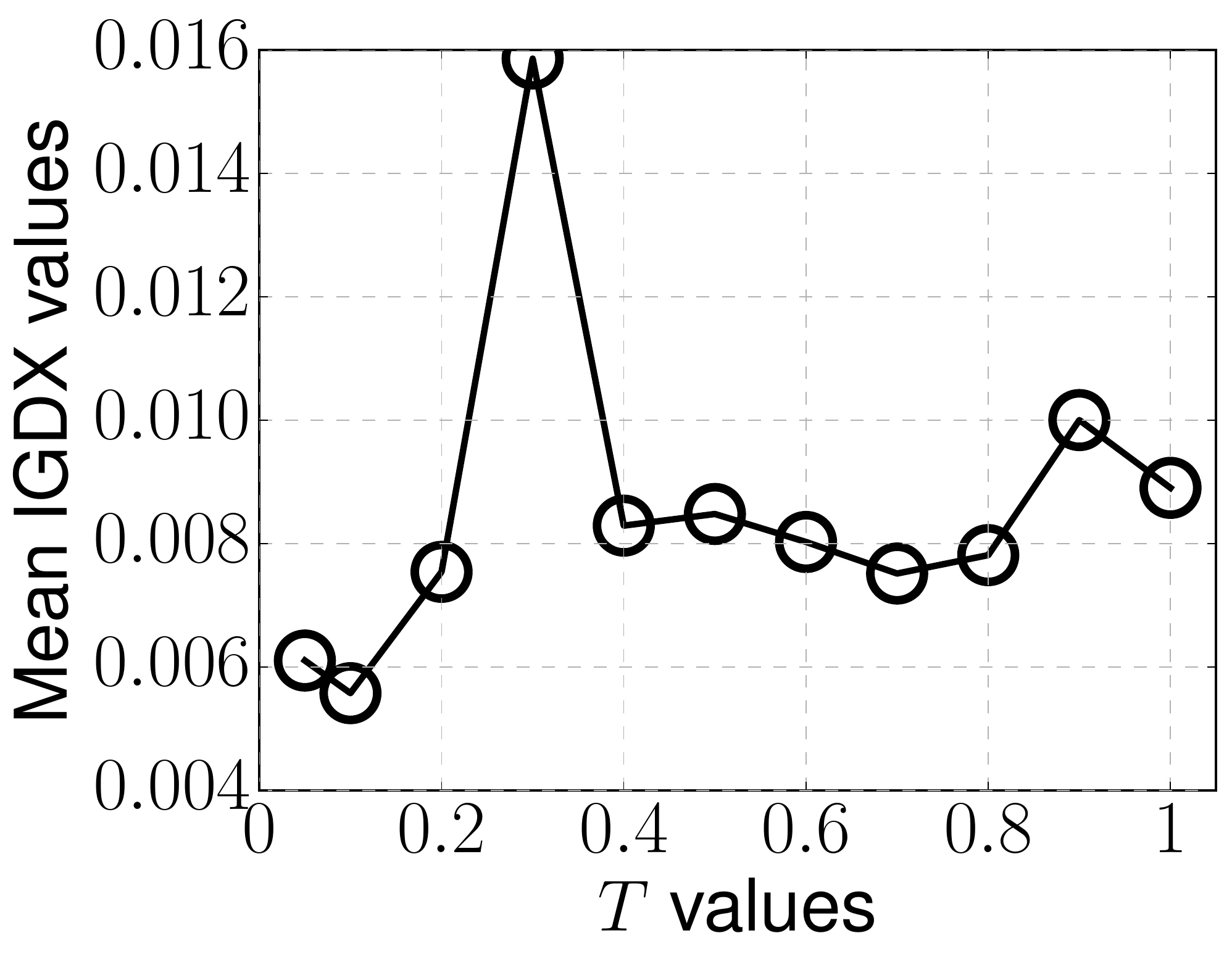}}
\subfloat[$8$-Polygon]{\includegraphics[width=\widthvar\textwidth]{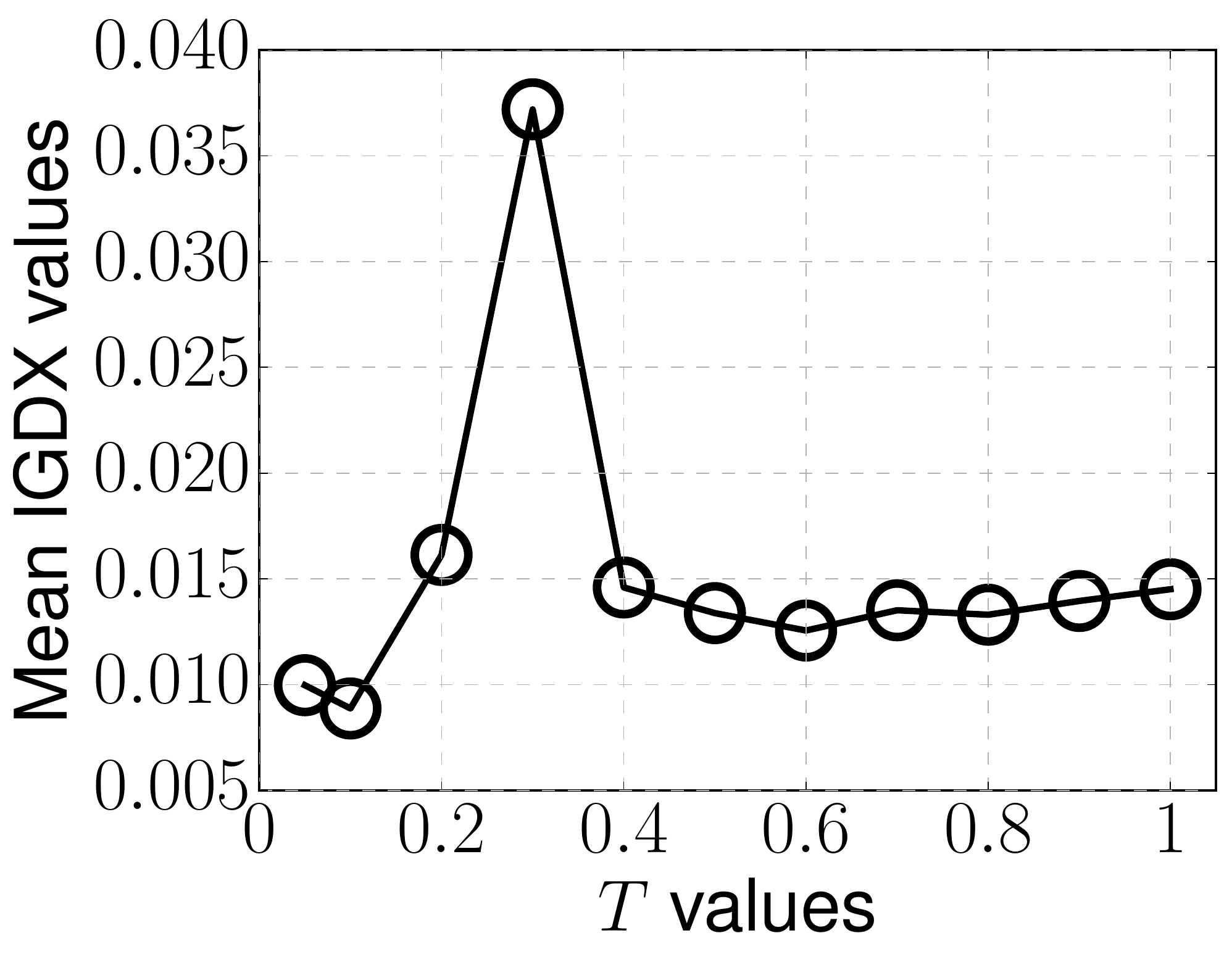}}
\subfloat[$15$-Polygon]{\includegraphics[width=\widthvar\textwidth]{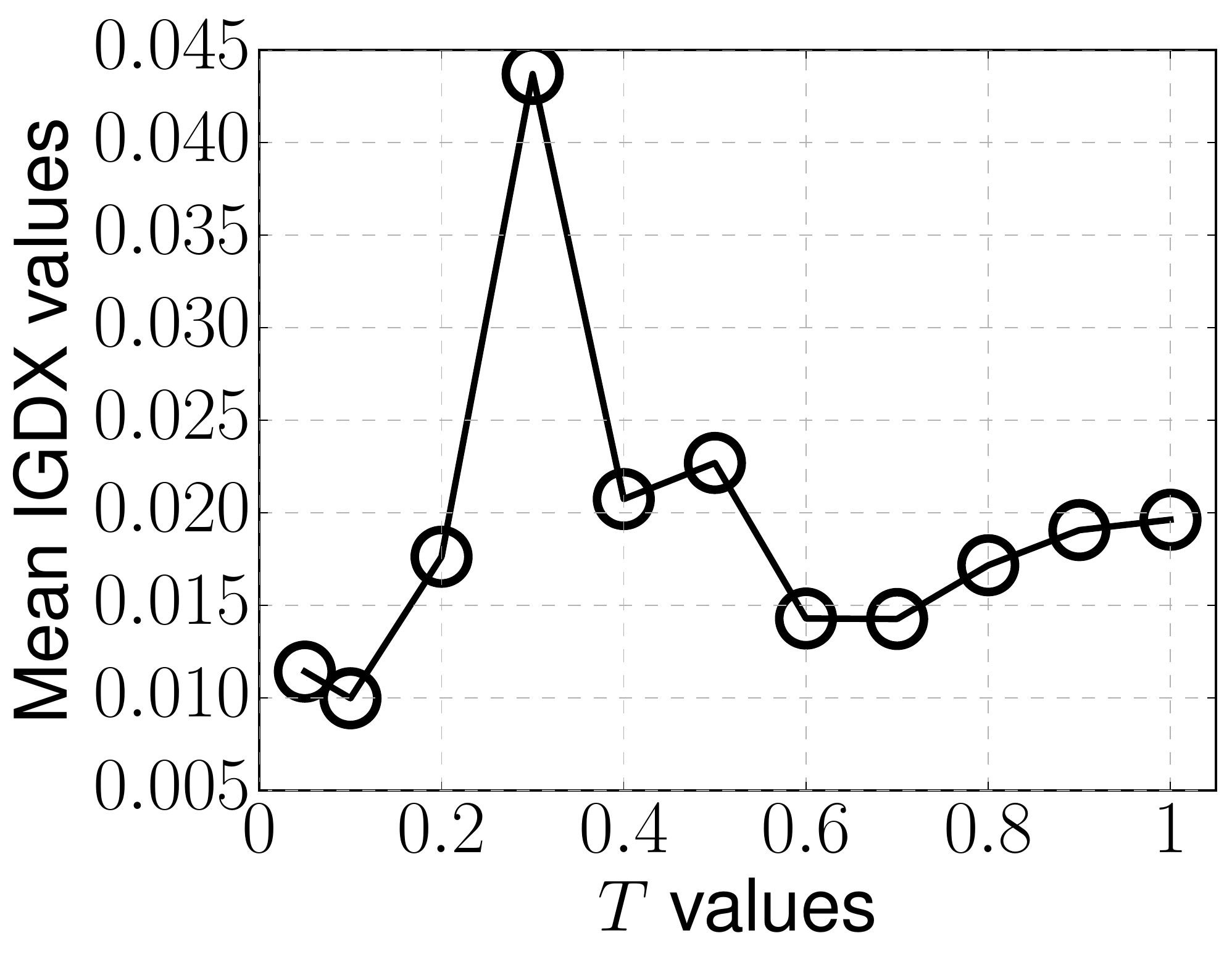}}
\caption{
\small
Comparison of NIMMO with 11 values of $T$.
The mean IGDX value at the final iteration among 31 runs is shown.
}
\label{fig:igdx_curves}
\end{figure}

\subsection{Influence of the neighborhood size $T$ on NIMMO}
\label{sec:invesigation_T}

%
%


Since it is important to find an appropriate control parameter setting of a novel evolutionary algorithm, we investigate how the performance of NIMMO is influenced by the value of $T$.
Table \ref{tab:Friedman_ranking_variousT_tradtional} shows the average ranks of NIMMO with $T \in \{\lfloor 0.05 \mu \rfloor, \lfloor 0.1 \mu \rfloor, \lfloor 0.2 \mu \rfloor, ..., \lfloor 0.9 \mu \rfloor, \lfloor 1.0 \mu \rfloor\}$. 
Tables S.2--S.3 in the supplementary file show detailed results of IGD, IGDX, and PSP, respectively.
The supplementary file is also available at the supplementary website (\url{https://sites.google.com/view/nimmopt/}).

As seen from Table \ref{tab:Friedman_ranking_variousT_tradtional} (a), NIMMO with $T = \lfloor 1.0 \mu \rfloor$ performs the best in terms of IGD.
A large $T$ value is generally beneficial for NIMMO on most problems.
This is because a large $T$ value relaxes the restriction of the niching behavior and makes NIMMO more efficient for multi-objective optimization.
In fact, NIMMO with $T = \lfloor 1.0 \mu \rfloor$ (i.e., $T=\mu$) does not consider the solution space diversity.
NIMMO with $T = \mu$ is identical with IBEA used in this study (see Subsection \ref{sec:settings_EMOAs}).
Tables \ref{tab:Friedman_ranking_variousT_tradtional} (b) and (c) show that NIMMO with $T = \lfloor 0.1 \mu \rfloor$ performs the best in terms of IGDX and PSP, respectively.
As seen from Tables S.2 and S.3 in the supplementary file, NIMMO with $T = \lfloor 0.1 \mu \rfloor$ performs significantly better than NIMMO with $T=\lfloor 0.05 \mu \rfloor$ on all the MMaOPs regarding IGDX and PSP.


Fig. \ref{fig:igdx_curves} shows the comparison of NIMMO with 11 values of $T$ on six representative test problems regarding IGDX.
Figs. \ref{fig:igdx_curves} (a)--(c) show that a small $T$ value is beneficial for NIMMO.
We do not claim that NIMMO with smaller $T$ values always performs better than NIMMO with larger $T$ values.
For example, Fig. \ref{fig:igdx_curves} (c) shows that NIMMO with $T=\lfloor 0.6 \mu \rfloor$ performs worse than NIMMO with $T=\lfloor 0.7 \mu \rfloor$ on SYM-PART1).
Interestingly, Figs. \ref{fig:igdx_curves} (d)--(e) show that NIMMO with $T = \lfloor 0.3 \mu \rfloor$ has the worst performance regarding IGDX on the Polygon problems with 3, 8, and 15 objectives.
These results suggest that an appropriate specification of $T$ is problem-dependent.
$T = \lfloor 0.1 \mu \rfloor$ is the best value for the three Polygon problems whereas $T = \lfloor 0.05 \mu \rfloor$ is the best for Two-On-One and Omni-test in Fig. \ref{fig:igdx_curves}.
NIMMO on the Polygon problems with 3, 8, and 15 objectives.
Although the best $T$ value in NIMMO is different for each problem, $T=\lfloor 0.1 \mu \rfloor$ is used for all experiments in this paper.

\subsection{The influence of the population size $\mu$ on the performance of NIMMO}
\label{sec:infuluence_mu}



Here, we investigate the influence of $\mu$ on the performance of NIMMO.
As discussed in \cite{IshibuchiMN15}, it is difficult to compare MOEAs with different settings of $\mu$ in a straightforward manner.
Since a large number of solutions are beneficial for some indicators (e.g., IGD and IGDX), a large $\mu$ value may be evaluated as being better than a small $\mu$ value, regardless of their actual effectiveness.
To address this issue, we compare NIMMO with different $\mu$ values based only on a pre-defined number of solutions $\mu^{\rm selected}$.
In this manner, we can fairly compare algorithms under the same solution set size.
For each solution set found by an algorithm, we selected $\mu^{\rm selected}$ solutions from the $\mu$ solutions in the final population $\vector{P}$ using the distance-based selection method presented in \cite{TanabeI18}.

At the beginning of the procedure, $\vector{A}$ is an empty set.
First, a solution is randomly selected from the final population $\vector{P}$ and stored into $\vector{A}$.
Then, the solution with the maximum $D(\vector{x}, \vector{A})$ value is repeatedly added to $\vector{A}$ until $|\vector{A}| = \mu^{\rm selected}$.
Here, $D(\vector{x}, \vector{A})$ is the distance between a solution $\vector{x}$ in $\vector{P}$ and its nearest solution in $\vector{A}$ in the normalized solution space.
Finally, we calculate the IGD, IGDX, and PSP indicator values of $\vector{A}$, not $\vector{P}$.

Table \ref{tab:Friedman_ranking_variousmu_tradtional} shows the comparison of NIMMO with $\mu \in \{100, 200, ..., 900, 1\,000\}$.
It should be noted that we can arbitrarily specify the population size $\mu$ in NIMMO.
We set  $\mu^{\rm selected}$ to $100$.
Tables Tables S.4--S.6 in the supplementary file show detailed results of IGD, IGDX, and PSP.
%
Table \ref{tab:Friedman_ranking_variousmu_tradtional} (a) shows that NIMMO with $\mu=200$ performs the best regarding IGD.
Tables \ref{tab:Friedman_ranking_variousmu_tradtional} (b) and (c) show that NIMMO with $\mu=400$ has the best performance regarding IGDX and PSP, respectively.
These results suggest that a little larger $\mu$ value is suitable for NIMMO for multi-modal multi-objective optimization.
However, Table \ref{tab:Friedman_ranking_variousmu_tradtional} shows that NIMMO with $\mu \geq 700$ performs poorly regarding all three indicators.
In summary, any $\mu$ value in $200 \leq \mu \leq 600$  is suitable for NIMMO.


\begin{table}[t]
\centering
\caption{\small Average rankings of NIMMO with 10 values of $\mu$ by the Friedman test.
}
\label{tab:Friedman_ranking_variousmu_tradtional}
\centering
{\scriptsize
  \subfloat[IGD]{    
\begin{tabular}{|c|c|}
\hline
  Algorithm&Rank\\
  \hline
$\mu=100$&6.04\\
$\mu=200$&3.76\\
$\mu=300$&4.24\\
$\mu=400$&4.60\\
$\mu=500$&4.28\\
$\mu=600$&4.92\\
$\mu=700$&5.72\\
$\mu=800$&6.60\\
$\mu=900$&6.96\\
$\mu=1000$&7.88\\
\hline
\end{tabular}
}
  \subfloat[IGDX]{    
\begin{tabular}{|c|c|}
\hline
  Algorithm&Rank\\
  \hline
$\mu=100$&7.64\\
$\mu=200$&4.32\\
$\mu=300$&4.48\\
$\mu=400$&4.20\\
$\mu=500$&4.60\\
$\mu=600$&4.96\\
$\mu=700$&5.20\\
$\mu=800$&5.88\\
$\mu=900$&6.60\\
$\mu=1000$&7.12\\
\hline
\end{tabular}
}
  \subfloat[PSP]{    
\begin{tabular}{|c|c|}
\hline
  Algorithm&Rank\\
  \hline
$\mu=100$&8.64\\
$\mu=200$&4.84\\
$\mu=300$&4.68\\
$\mu=400$&4.08\\
$\mu=500$&4.44\\
$\mu=600$&4.68\\
$\mu=700$&4.92\\
$\mu=800$&5.68\\
$\mu=900$&6.12\\
$\mu=1000$&6.92\\
\hline
\end{tabular}
  }
  }
\end{table}

\subsection{The performance of NIMMO with other fitness assignment schemes}
\label{sec:nimmo_other_indicators}

Although NIMMO uses the $I_{\epsilon+}$ indicator to assign the fitness values to individuals in the population, NIMMO can use any indicator-based fitness assignment scheme.
Here, we investigate the performance of NIMMO with the following three fitness assignment schemes: the hypervolume difference indicator ($I_{HD}$) \cite{ZitzlerK04}, the binary R2 indicator ($I_{R2}$) \cite{PhanS13}, and the ranking method in SRA (SRA) \cite{LiTLY16}.
For $I_{HD}$, $I_{R2}$, and SRA, see Subsection \ref{sec:indicator_based_emoas}.

We denote NIMMO with the three fitness assignment schemes as NIMMO-$I_{HD}$, NIMMO-$I_{R2}$, and NIMMO-SRA, respectively.
In NIMMO-$I_{HD}$ and NIMMO-$I_{R2}$, $I_{\epsilon+}$ in \eqref{eqn:additive_epsilon_indicator} is replaced with $I_{HD}$ and $I_{R2}$, respectively.
In NIMMO-SRA, the fitness assignment in \eqref{eqn:fitness_ibea} is replaced with the ranking method in SRA.
Thus, the ranks of individuals are their fitness values in NIMMO-SRA.
As in \cite{LiTLY16}, the control parameter $p_c$ in SRA is randomly generated in the range $[0.4, 0.6]$ for each iteration.

Table \ref{tab:ranks_nimmo_other_indicators} summarizes the comparison results of the four variants of NIMMO.
NIMMO-$I_{\epsilon+}$ is the original version of NIMMO described in Section \ref{sec:proposed_method}.
Tables S.7--S.9 in the supplementary file show detailed results of IGD, IGDX, and PSP, respectively.
Note that we do not intend to generalize the results of NIMMO with the four fitness assignment schemes.
Even though NIMMO-$I_{\epsilon+}$ performs better than NIMMO-$I_{R2}$, it does not mean that $I_{\epsilon+}$ performs better than $I_{R2}$.
It only means that $I_{\epsilon+}$ is more suitable in NIMMO than $I_{R2}$ for multi-modal many-objective optimization.

Table \ref{tab:ranks_nimmo_other_indicators} (a) shows that NIMMO-SRA performs the best regarding IGD.
Thus, NIMMO-SRA is the best configuration for multi-objective optimization.
The good performance of NIMMO-SRA regarding IGD is consistent with the results reported in \cite{LiTLY16}.
In contrast, Tables \ref{tab:ranks_nimmo_other_indicators} (b) and (c) show that NIMMO-$I_{\epsilon+}$ performs the best regarding IGDX and PSP, respectively.
Although NIMMO-$I_{HD}$ and NIMMO-SRA outperforms NIMMO-$I_{\epsilon+}$ on some test problems (e.g., some MMF problems) regarding IGDX and PSP, NIMMO-$I_{\epsilon+}$ has the best performance on the Polygon problems with up to 15 objectives.
$I_{\epsilon+}$ and the solution space diversity maintenance mechanism in NIMMO may be compatible.
In summary, our results indicate that $I_{\epsilon+}$ is most suitable in NIMMO for multi-modal multi-objective optimization.

\begin{table}[t]
\centering
\caption{\small Average rankings of the four variants of NIMMO by the Friedman test.
}
\label{tab:ranks_nimmo_other_indicators}
\centering
{\scriptsize
  \subfloat[IGD]{    
\begin{tabular}{|c|c|}
\hline
  Algorithm&Rank\\
  \hline
NIMMO-$I_{\epsilon}$&1.83\\
NIMMO-$I_{HD}$&3.26\\
NIMMO-$I_{R2}$&3.35\\
NIMMO-SRA&1.57\\
\hline
\end{tabular}
}
  \subfloat[IGDX]{    
\begin{tabular}{|c|c|}
\hline
  Algorithm&Rank\\
  \hline
NIMMO-$I_{\epsilon}$&1.61\\
NIMMO-$I_{HD}$&3.22\\
NIMMO-$I_{R2}$&3.35\\
NIMMO-SRA&1.83\\
\hline
\end{tabular}
}
  \subfloat[PSP]{    
\begin{tabular}{|c|c|}
\hline
  Algorithm&Rank\\
  \hline
NIMMO-$I_{\epsilon}$&1.65\\
NIMMO-$I_{HD}$&3.26\\
NIMMO-$I_{R2}$&3.35\\
NIMMO-SRA&1.74\\
\hline
\end{tabular}
  }
  }
\end{table}

\section{Conclusion}
\label{sec:conclusion}



We proposed NIMMO for multi-modal many-objective optimization.
The performance of NIMMO was investigated on various problems.
Our results showed that NIMMO is capable of finding multiple equivalent Pareto optimal solutions on MMaOPs with up to $15$ objectives.
%

We uploaded source code of NIMMO at the supplementary website (\url{https://sites.google.com/view/nimmopt/}).
Source code of the 23 test problem instances used in our benchmarking study is also available at the supplementary website.
Although MMOPs have not been well studied \cite{LiEDE17,TanabeI19emmo}, we hope that this paper and its supplement encourage further development of efficient multi-modal multi- and many-objective optimizers.

It was shown in \cite{NojimaNKI05} that there are multiple equivalent Pareto optimal solutions in multi-objective knapsack problems.
An analysis of the performance of NIMMO on combinatorial MMOPs and MMaOPs is an avenue for future work.

 \section*{Acknowledgement}

 This work was supported by National Natural Science Foundation of China (Grant No. 61876075), the Program for Guangdong Introducing Innovative and Enterpreneurial Teams (Grant No. 2017ZT07X386), Shenzhen Peacock Plan (Grant No. KQTD2016112514355531), the Science and Technology Innovation Committee Foundation of Shenzhen (Grant No. ZDSYS201703031748284), and the Program for University Key Laboratory of Guangdong Province (Grant No. 2017KSYS008).

 

\section*{References}

\bibliography{reference}

\end{document}